\documentclass[preprint,11pt,a4paper]{elsarticle}

\usepackage[T1]{fontenc}
\usepackage[english]{babel}

\usepackage[scaled=.98]{newtxtext}

\usepackage{geometry}
\usepackage{amsmath,amssymb,amsthm,mathtools,array,bm,dsfont,mathrsfs,nccmath}
\usepackage{graphicx}
\usepackage{subcaption}
\usepackage{float}
\usepackage{algorithm}
\usepackage{algpseudocode}
\usepackage{enumitem}
\usepackage{comment}
\usepackage{todonotes}
\usepackage{xcolor}
\usepackage[colorlinks=true,linkcolor=cyan,urlcolor=red,citecolor=black]{hyperref}

\biboptions{authoryear}

\numberwithin{equation}{section}
\numberwithin{figure}{section}
\numberwithin{table}{section}

\newcommand{\R}{\mathbb{R}}
\newcommand{\N}{\mathbb{N}}

\newcommand{\E}{\mathbb{E}}

\newcommand{\cA}{\mathcal{A}}

\DeclareMathOperator*{\argmin}{arg\,min}

\newcommand{\edg}[1]{\left[#1\right]}
\newcommand{\brak}[1]{\left(#1\right)}

\newtheorem{Theorem}{Theorem}[section]
\newtheorem{Lemma}[Theorem]{Lemma}
\newtheorem{Assumption}[Theorem]{Assumption}
\newtheorem{Proposition}[Theorem]{Proposition}

\theoremstyle{definition}

\theoremstyle{remark}

\usepackage{hyperref}
\usepackage[hyperref]{xcolor}

\hypersetup{
	colorlinks=true,
}
\usepackage{etoolbox}
\DeclareCaptionSubType*{figure} 
\makeatletter
\patchcmd{\chapter}{\if@openright\cleardoublepage\else\clearpage\fi}{}{}{}
\def\ps@pprintTitle{%
	\let\@oddhead\@empty
	\let\@evenhead\@empty
	\def\@oddfoot{}%
	\let\@evenfoot\@oddfoot}
\makeatother
\begin{document}
		\hypersetup{
		linkcolor=cyan,
		urlcolor=cyan,
		citecolor=cyan,
	}
	\setcounter{page}{0}
	\begin{frontmatter}
		
		
		
		\title{Deep Pontryagin Learning for Continuous-Time Stochastic Control with Jumps}
		
		\author[inst1]{Jean-Loup Dupret\corref{cor1}}
		\author[inst1]{Patrick Cheridito}
		\affiliation[inst1]{organization={Department of Mathematics and RiskLab, ETH Zurich},
			addressline={R\"amistrasse 101},
			city={Zurich},
			postcode={8092},
			country={Switzerland}}

		\cortext[cor1]{Corresponding author, Email address: \url{jeanloup.dupret@math.ethz.ch}}
		

		\begin{abstract}
			We develop a deep Pontryagin learning framework for high-dimensional continuous-time stochastic control problems that naturally accommodates jump-diffusions. In contrast to direct policy optimization, which repeatedly differentiates the objective through the entire time-discretized dynamics, our method separates adjoint learning from policy improvement by explicitly incorporating state-sensitivity information from the Pontryagin stochastic maximum principle. For a fixed Markov control, we first compute pathwise adjoint labels  by backpropagating the reward through time along simulated state trajectories, and project these labels onto the Markov state through a supervised regression step. The control is then updated by enforcing the pointwise Hamiltonian stationarity condition. We prove that the resulting  projected adjoint process satisfies a discrete-time Pontryagin principle and converges, as the time step tends to zero, to the solution of the corresponding continuous-time adjoint backward stochastic differential equation. Numerical experiments illustrate the accuracy, stability, and scalability of the approach.

		\end{abstract}
		
		
	\end{frontmatter}
	
	\section{Introduction}
	
	Stochastic control problems are central in many areas of applied mathematics,
	finance, insurance, engineering and operations research. In continuous time,
	they are classically characterized either by the dynamic programming principle \citep{fleming2006controlled},
	which leads to a Hamilton--Jacobi--Bellman (HJB) partial differential equation (PDE), or by the Pontryagin stochastic
	maximum principle (PMP) \citep{peng1990general}, which leads to a system of forward--backward stochastic
	differential equations (FBSDE). While these characterizations are theoretically
	powerful, their numerical implementation remains challenging in high dimension.
	Grid-based methods  for HJB equations \citep{bonnans2003consistency, oberman2006convergent,beard1997galerkin,boulbrachene2001finite} suffer from the curse of dimensionality, and classical
	tree  or regression-based  BSDE  methods \citep{ma2002numberical,gobet2005regression} often become prohibitively
	expensive in the PMP framework when the state dimension or the number of time steps is large.
	\\[3mm]
	Recent progress in deep learning has opened new possibilities for solving
	high-dimensional stochastic control problems \citep{hu2023recent}. A first class of methods is based on the HJB--PDE and
	approximates the value function globally, for instance through  physics-informed neural networks  \citep{raissi2017physics,sirignano2018dgm,duarte2024machine, dupret2026deep}, continuous-time Bellman updating \citep{cheridito2025deep, dupret2026ineus} or constrained Gaussian process regressions \citep{chen2021solving,hainaut2025optimal}. These methods are attractive because
	they can provide a global approximation of the value function and the optimal
	feedback control over the entire space-time domain. This is useful when one
	needs a reusable solution over many initial time-space conditions. However, when the goal
	is to solve the control problem for a fixed initial state or along the optimal state trajectory, computing a global
	space-time solution may be unnecessarily expensive.
	\\[3mm]
	In contrast to global PDE-based approaches, another stream of literature
	adopts a simulation-based viewpoint. The controlled dynamics are discretized in
	time, and neural networks are trained on simulated state trajectories to approximate
	the feedback control, the value function, or related adjoint quantities
	\citep{han2016deep,han2017deep,han2018solving,han2021recurrent,hure2021deep,beck2021deep,bachouch2022deep,ji2022solving,
		van2024global, alasseur2024deep}. Particularly popular is the direct policy optimization (DPO) of \cite{han2016deep},  which extends ideas from the earlier neural-network control literature for deterministic nonlinear systems
	\citep{psaltis1988multilayered,hunt1992neural}. In DPO, the control is parameterized directly by a neural network and trained by differentiating an empirical approximation of the objective functional. This avoids the explicit solution of the HJB
	equation and is therefore simple to implement. However, DPO has important
	limitations. Since the loss is defined through long simulated trajectories,
	gradients have to be propagated through the full time-discretized dynamics.
	This can lead to memory-intensive training, accumulation of discretization and
	approximation errors over time and vanishing/exploding gradients on long
	time horizons \citep{bachouch2022deep}. As a consequence, DPO and related simulation-based time-discretized methods may be unstable and may
	converge to poor local minima, especially in high-dimensional or strongly
	nonlinear problems.
	\\[3mm]
	To avoid the aforementioned difficulty  that a local change in the control affects the entire   future state trajectory,  adjoint sensitivity methods \citep{pontryagin2018mathematical, peng1990general} instead introduce  auxiliary adjoint state that propagates the state sensitivities of the objective backwards in time. Once the adjoint is known, optimality reduces to a pointwise
	maximization of the Hamiltonian with respect to the control, along the state
	trajectory; see Theorem~\ref{th:PMP}. 
	In the PDE-constrained optimization literature \citep{plessix2006review, hinze2009optimization}, adjoint equations are widely used in a deterministic setting to compute gradients efficiently, since their computational cost scales  linearly with the problem dimension and have low memory cost. Closely related ideas have also been used in scientific machine learning to improve the training of neural ODEs and neural PDE solvers \citep{chen2018neural, mowlavi2023optimal, nzoyem2023comparison,yin2024aonn, hintermuller2026hybrid}. Recent work has extended these adjoint-based techniques to continuous-time stochastic portfolio optimization via backpropagation through time \citep{huh2025breaking, huh2025pontryagin}. However, the resulting \textit{Pontryagin-Guided Direct Policy Optimization}  method remains computationally expensive because they rely on a preliminary DPO stage and require nested Monte Carlo simulations to estimate state-dependent adjoint quantities. Nevertheless, adjoint sensitivity methods demonstrate that incorporating state sensitivity information can substantially improve learning by providing first-order information about how future
	rewards depend on current states.
	\\[3mm]
	In this paper, we propose a deep stochastic Pontryagin maximum principle (deep PMP) framework  that avoids
	the nested simulation bottleneck of  \cite{huh2025breaking,huh2025pontryagin} and which contrasts with existing deterministic Pontryagin-based methods \citep{jin2020pontryagin,bottcher2022ai}. The  idea is to compute pathwise
	adjoint labels by backpropagation through time (BPTT) of the objective along simulated state trajectories.
	For each simulated trajectory, differentiating the realized future payoff with
	respect to intermediate states yields noisy but cheap estimates of the
	discrete-time adjoint variables. These labels are then projected onto the
	Markov state through a supervised regression step, producing a neural
	approximation of the discrete-time adjoint process. The control network is subsequently trained
	by enforcing the pointwise Hamiltonian stationarity condition from the
	PMP. In this way, the method separates the learning of
	adjoint sensitivities from the learning of the control and avoids repeated
	backpropagation through the full empirical objective as in \cite{han2016deep, ji2022solving, alasseur2024deep}.
	\\[3mm]
	The proposed approach therefore combines several advantages. First, the control is updated through a pointwise Hamiltonian optimization, thereby avoiding the stability issues associated with end-to-end policy training by backpropagation through the full time-discretized dynamics. Second, it incorporates the adjoint/sensitivity structure of the
	continuous-time stochastic control problem through PMP, improving convergence of the scheme. Third, the BPTT adjoint labels are obtained from a single set of
	simulated trajectories and do not require nested Monte Carlo simulations; only a supervised regression step is needed instead. Fourth, the proposed framework naturally accommodates controlled
	jump-diffusions, where the adjoint BSDE contains both Brownian and
	jump martingale components and for which  deep-learning methods remain comparatively scarce. Finally, we prove that the projected  BPTT discrete adjoint process
	satisfy a discrete-time analogue of the PMP and
	converge, as time step tends to zero, to the continuous-time adjoint BSDE associated with the limiting controlled jump-diffusion.

	\section{Preliminaries}
	Let \(T>0\) be a fixed finite time horizon and let \( (\Omega,\mathcal F,\mathbb F=(\mathcal F_t)_{0\leq t\leq T},\mathbb P) \) be a filtered probability space satisfying the usual conditions. We assume that \(\mathbb F\) is the usual augmentation of the filtration jointly generated by a \(d\)-dimensional Brownian motion \(W\) and a Poisson random measure \(N\) on \([0,T]\times E\), where \( E:=\mathbb R^\ell\setminus\{0\} \)  is equipped with its Borel \(\sigma\)-algebra $\mathcal{B}(E)$. The random measure \(N\) is assumed to be independent of \(W\) and to have deterministic compensator \( \lambda(t)\nu(dz)\,dt, \) where \(\nu\) is a Lévy measure on \(E\), that is, \( \int_E (1\wedge |z|^2)\,\nu(dz)<\infty, \) 
	and \(\lambda:[0,T]\to(0,\infty)\) is deterministic and continuous. We then denote the associated compensated Poisson random measure by \( \widetilde N(dt,dz) := N(dt,dz)-\lambda(t)\nu(dz)dt \, .\) 
	In this paper, we consider control problems of the form 
	\begin{equation}
		\sup_{\alpha\in\cA} J(\alpha),
		\qquad
		J(\alpha):=
		\E\edg{\int_0^T f(t,X_t^\alpha,\alpha_t)\,dt+F(X_T^\alpha)} \, , 
		\label{pr}
	\end{equation}
	such that $\alpha = (\alpha_t)_{0 \le t \le T} \in \mathcal{A}$ for the set of admissible controls $\mathcal{A}$  given by 
	\begin{equation}
		\mathcal{A} := \Big\{\alpha : [0,T] \times \Omega \to A \subseteq \R^m \, \big| \, \alpha  \text{ is $\mathbb{F}$-predictable and } \E\Bigl[\int_0^T |\alpha_t|^2\,dt\Bigr]<\infty \Big\} .
	\end{equation}  The controlled process $X^\alpha$ then evolves in  $\R^n$ according to 
	\begin{align}
		dX^\alpha_t &= b(t, X^\alpha_t, \alpha_t) dt + \sigma(t, X^\alpha_t, \alpha_t) dW_t  
		+ \int_E \gamma(t, X^\alpha_{t-}, z, \alpha_t) \widetilde{N}(dt, dz), \quad X^\alpha_0 = x_0 \in \R^n, 	\label{sde}
	\end{align}
	for  suitable functions $b \colon [0,T] \times \R^n \times A \to \R^n$, 
	$\sigma \colon [0,T] \times \R^n \times A \to \R^{n \times d}$,
	$\gamma \colon [0,T] \times \R^n \times E \times A \to \R^n$.  The functions $f \colon [0,T] \times \R^n \times A \to \R$ 
	and  $F \colon \R^n \to \R$ then  model the running reward and the final reward, respectively.
	For $k\in\mathbb N$, we use the following spaces {\small
		\begin{align*}
			&\mathcal S^2_{\mathbb F}(0,T;\mathbb R^k)
			:=
			\left\{
			Y:[0,T]\times \Omega \to\mathbb R^k
			\,\middle|\,
			Y \text{ is c\`adl\`ag and } \mathbb F\text{-adapted, }
			\mathbb E\big[\sup_{0\le t\le T}|Y_t|^2\big]<\infty
			\right\},
			\\
			&L^2_\mathbb{F}(0,T; \R^{k}) := \Big\{Z :[0,T] \times \Omega \to \R^{k} \, \big| \, Z \text{ is $\mathbb{F}$-predictable and } \E\Bigl[\int_0^T | Z_t |^2 \, dt\Bigr] < \infty \Big\} \, ,
			\\
			&L^2_{\nu,\mathbb{F}}(0,T; \R^k) \hspace{-1mm} := \hspace{-1mm}\Big\{ R: [0,T]  \times \Omega \times E \to \R^k \, \big| \, R \text{ is } \mathcal P\otimes\mathcal B(E)\text{-measurable} 
			\text{, }\,\E \Bigl[\int_0^T \int_E  |R_t(z)|^2 \nu(dz) \lambda(t) dt \Bigr]< \infty \Big\} .
	\end{align*}}Here \(\mathcal P\) denotes the predictable \(\sigma\)-algebra on \([0,T]\times\Omega\), and we use the standard spaces
$
L^2(\Omega,\mathcal{F};\mathbb R^k)
:=
L^2(\Omega,\mathcal F,\mathbb P;\mathbb R^k)
$
and
$
L^2(\Omega\times[0,T];\mathbb R^k)
:=
L^2(\Omega\times[0,T],
\mathcal F\otimes\mathcal B([0,T]),
\mathbb P\otimes dt;\mathbb R^k)$, which do not impose any adaptedness or predictability. Throughout, \(|\cdot|\) denotes the Euclidean norm for vectors and the Frobenius norm for matrices. For a measurable function \(g:E\to\mathbb R^k\), we also  write \[ \|g\|_{L^2(E,\nu)} := \left( \int_E|g(z)|^2\,\nu(dz) \right)^{1/2}. \]  
	
	If $\alpha\in \mathcal{A}$ and $X^\alpha \in \mathcal{S}^2_{\mathbb{F}}(0,T;\R^n)$ is the corresponding càdlàg solution of \eqref{sde}, we call $(X^\alpha,\alpha)$ an \textit{admissible pair}. Our stochastic control problem \eqref{pr} can be stated as maximizing $J(\alpha)$ over $\mathcal{A}$. The goal is then to find a control $\alpha^* \in \mathcal{A}$ (if it exists) such that 
	\begin{equation}
		J(\alpha^*) = \sup_{\alpha \in \mathcal{A}} J(\alpha)\, .
	\end{equation}
	We impose the following regularity and growth assumptions on the
	coefficients.
	
	\begin{Assumption}
		\label{Ass1}
		The following conditions hold.
		
		\begin{enumerate}[label=\rm(\roman*)]
			
			\item
			The maps $b,\sigma,\gamma$ are measurable and $C^1$ in $x$. Moreover, there exist constants $L,L_x>0$ such that, for every
			$(t,x,a)\in[0,T]\times\mathbb R^n\times A$,
			\[
			|b(t,0,a)|
			+
			|\sigma(t,0,a)|
			+
			\|\gamma(t,0,\cdot,a)\|_{L^2(E,\nu)}
			\le
			L(1+|a|),
			\]
			and
			\[
			|\nabla_x b(t,x,a)|
			+
			|\nabla_x\sigma(t,x,a)|
			+
			\|\nabla_x\gamma(t,x,\cdot,a)\|_{L^2(E,\nu)}
			\le L_x.
			\]
			
			\item
			The reward functions $f,F$
			are measurable and $C^1$ in $x$. Moreover, there exists a constant $C>0$ such that,
			for every $(t,x,a)\in[0,T]\times\mathbb R^n\times A$,
			\[
			|f(t,0,a)|\le C(1+|a|^2),
			\qquad
			|F(0)|\le C,
			\]
			and
			\[
			|\nabla_x f(t,x,a)|
			\le C(1+|x|+|a|),
			\qquad
			|\nabla_xF(x)|
			\le C(1+|x|).
			\]
			
		\end{enumerate}
	\end{Assumption}
	Under Assumption~\ref{Ass1}(i), the coefficients of the state equation
	are globally Lipschitz continuous in $x$, uniformly in $(t,a)$, and they satisfy the growth estimate
	\[
	\begin{aligned}
		|b(t,x,a)|
		+|\sigma(t,x,a)|
		+\|\gamma(t,x,\cdot,a)\|_{L^2(E,\nu)}
		\le
		C\bigl(1+|x|+|a|\bigr) \, .
	\end{aligned}
	\]
	Consequently, for every control
	$\alpha\in\mathcal A$ and every deterministic initial condition
	$x_0\in\mathbb R^n$, equation \eqref{sde}
	admits a unique càdlàg strong solution $
	X^\alpha\in\mathcal S^2_{\mathbb F}(0,T;\mathbb R^n)$. Moreover,
	\[
	\mathbb E\left[
	\sup_{0\le t\le T}|X^\alpha_t|^2
	\right]
	\le
	C\left(
	1+|x_0|^2
	+
	\mathbb E\left[\int_0^T|\alpha_t|^2\,dt\right]
	\right),
	\]
	and hence $(X^\alpha,\alpha)$ is an admissible pair, 
	see also  \cite{bouchard2007introduction, applebaum2009levy}$^{\ref{test}}$. 
	Assumption~\ref{Ass1}(ii) further implies that
	\(
	|f(t,x,a)|
	\le C\bigl(1+|x|^2+|a|^2\bigr)\) and \( 
	|F(x)|
	\le C\bigl(1+|x|^2\bigr).
	\) It then follows  that
	\[
	\mathbb E\left[
	\int_0^T
	|f(t,X^\alpha_{t-},\alpha_t)|\,dt
	+
	|F(X^\alpha_T)|
	\right]<\infty.
	\]
	Thus, $J(\alpha)$ is well defined and finite for every
	$\alpha\in\mathcal A$.
	\\[2mm]
	We then define the Hamiltonian $H:[0,T]\times \R^n \times A \times \R^n \times \R^{n\times d} \times L^2(E,\nu;\R^n) \to \R$ by, 
	\begin{align*}
		H(t, x, a, p, q, r) \hspace{-0.3mm}:=&  \, \hspace{-0.5mm}f(t,x, a) \hspace{-0.3mm}+ \hspace{-0.3mm}
		b(t,x, a)^\top \, p\hspace{-0.4mm} + \hspace{-0.4mm}  \text{Tr} \left[ \sigma(t,x,a)^\top q \right]  + \hspace{-0.3mm}\lambda(t) \hspace{0.3mm} \int_E \hspace{-0.3mm} \gamma \hspace{-0.1mm}(t,x,z, a)^\top r(z) \, \nu(dz) .
	\end{align*} 
	The last integral is finite by Cauchy-Schwarz inequality since $\gamma(t,x,\cdot,a),r \in L^2(E,\nu;\R^n)$. Given an admissible pair \((X^\alpha,\alpha)\), the  adjoint equation is the BSDE
	\begin{equation}
		\begin{cases}
			dp_t
			=
			-\nabla_x H(t,X^\alpha_{t-},\alpha_t,p_t,q_t,r_t(\cdot))\,dt
			+
			q_t\,dW_t
			+
			\displaystyle\int_E r_{t}(z)\,\widetilde N(dt,dz), \\[1.5ex]
			p_T=\nabla_x F(X^\alpha_T).
		\end{cases}
		\label{adjoint}
	\end{equation}
	We call an \emph{adapted solution} of \eqref{adjoint} a triplet \[ (p,q,r) \in \mathcal S^2_{\mathbb F}(0,T;\mathbb R^n) \times L^2_{\mathbb F}(0,T;\mathbb R^{n\times d}) \times L^2_{\nu,\mathbb F}(0,T;\mathbb R^n) \]
	satisfying \eqref{adjoint}.\ Assumption \ref{Ass1}, together with the boundedness\footnote{\label{test}The deterministic time-dependent intensity \(\lambda(t)\) does not alter the usual existence and unicity arguments. Because $\lambda$ is continuous on the compact interval $[0,T]$, it is automatically bounded above and away from zero, $0 <\underset{\bar{}}{\lambda} \leq \lambda(t) \leq \bar{\lambda} < \infty$.} of $\lambda$ and square-integrability of $X^\alpha$ and $ \alpha$,  implies that the driver $
	(p,q,r)\mapsto
	\nabla_x H(t,X^\alpha_{t-},\alpha_t,p,q,r)$
	is uniformly Lipschitz in $(p,q,r)$ and satisfies 
	$$\mathbb E\left[
	\int_0^T
	\left|
	\nabla_x H(t,X^\alpha_{t-},\alpha_t,0,0,0)
	\right|^2 dt
	+
	|\nabla_xF(X^\alpha_T)|^2
	\right]<\infty .$$
	Therefore, the standard well-posedness theory for BSDEs driven by Brownian motion and compensated Poisson measure yields a unique adapted solution of \eqref{adjoint} for every admissible pair $(X^\alpha, \alpha) \in \mathcal S^2_{\mathbb F}(0,T;\mathbb R^n) \times \mathcal{A}$, see e.g.\ Theorem 2 of  \cite{Rong1997}.
	\\[2mm]
	The following result  extends  Theorem 2.1 of \cite{framstad2004sufficient} and gives sufficient conditions for admissible pairs to be optimal in our setting with time-dependent jump intensity $\lambda(t)$.
	\begin{Theorem}[Pontryagin's stochastic Maximum Principle] \label{th:PMP}
		Let Assumption \ref{Ass1}   hold and suppose that $F$ is concave. Let $(\hat{X}, \hat{\alpha})$ be an admissible pair  to the state equation \eqref{sde} and let $(\widehat{p}, \hat{q}, \hat{r})$ be the unique adapted solution of the associated adjoint BSDE  \eqref{adjoint}.
		Assume that the following conditions hold:
		\begin{enumerate}[leftmargin=*,  label=(\roman*)]
			\item \label{prop:contrac:item1} 
			(Integrability condition) The following conditions hold for all $\alpha \in \mathcal{A}$:
			\begin{equation} \label{eq:integrability conditions}
				\begin{aligned}
					&\E\left[
					\int_0^T
					\left|(\hat X_t-X_t^\alpha)^\top \hat q_t\right|^2dt
					+
					\int_0^T\int_E
					\left|(\hat X_{t-}-X_{t-}^\alpha)^\top \hat r_t(z)\right|^2
					\lambda(t)\nu(dz)dt
					\right]
					<\infty,
					\\
					&\E\left[
					\int_0^T
					\left|
					\hat p_{t-}^\top
					\bigl(\sigma(t,X_{t-}^\alpha,\alpha_t)
					-\sigma(t,\hat X_{t-},\hat\alpha_t)\bigr)
					\right|^2dt
					\right.
					\\
					&\hspace{1cm}\left.
					+
					\int_0^T\int_E
					\left|
					\hat p_{t-}^\top
					\bigl(\gamma(t,X_{t-}^\alpha,z,\alpha_t)
					-\gamma(t,\hat X_{t-},z,\hat\alpha_t)\bigr)
					\right|^2
					\lambda(t)\nu(dz)dt
					\right]
					<\infty .
				\end{aligned}
			\end{equation}
			
			\item (Maximum principle) For \(dt\otimes d\mathbb P\)-a.e.\ \((t,\omega)\in[0,T]\times\Omega\),
			\begin{equation}
				H(t,\hat{X}_{t-}, \hat{\alpha}_t, \widehat{p}_t, \hat{q}_t, \hat{r}_t(\cdot)) = \sup_{a \in A} 	H(t,\hat{X}_{t-},  a, \widehat{p}_t, \hat{q}_t, \hat{r}_t(\cdot)) \label{max1}\,  ,
			\end{equation}
			\item (Arrow condition) For \(dt\otimes d\mathbb P\)-a.e.\ \((t,\omega)\in[0,T]\times\Omega\), the function
			\begin{equation}
				\hat{H} : \R^n \to \R; \, x \mapsto \hat{H}(x) := \sup_{a \in A} 	H(t,x,  a, \widehat{p}_t(\omega), \hat{q}_t(\omega), \hat{r}_t(\cdot)(\omega)) \label{max2} \, ,
			\end{equation}
			is finite-valued and concave on $\R^n$. 
		\end{enumerate}
		Then, $( \hat{X}, \hat{\alpha})$ is an optimal pair, that is, \( J(\hat\alpha) = \sup_{\alpha\in\mathcal A}J(\alpha). \)
	\end{Theorem}
	The detailed proof is deferred in Appendix \ref{proof:theorem}. We next describe the connection between dynamic programming and the stochastic maximum principle. We define the state-dependent reward functional 
	\begin{equation}
		J(t,x;\alpha)
		:=
		\E\edg{
			\int_t^T f(s,X^\alpha_s,\alpha_s)\,ds+F(X^\alpha_T)
			\;\Big|\;
			X_t=x
		}.
		\label{eq:performance-functional}
	\end{equation}
	The corresponding value function is then
	\begin{equation}
		V(t,x)
		:=
		\sup_{\alpha\in\cA} J(t,x;\alpha).
		\label{value}
	\end{equation}
	In view of the Markovian nature of the dynamics \eqref{sde} and of the reward functions $f$ and $F$, the control problem is Markovian. 
	Under mild assumptions, see e.g.\ Theorem 11.2.3 of \cite{oksendal2013stochastic}, restricting the optimization to admissible Markov feedback controls does not change the value function.  Accordingly,  we restrict attention in the remainder of the paper to controls of the form \(\alpha_s=\alpha(s,X^\alpha_{s-}), \) where \( \alpha :[0,T]\times\mathbb R^n\to A \) is measurable and the resulting control admissible. With slight abuse of notation, we retain the notation \(\mathcal A\) for the admissible control class and define the value function by \eqref{value}. The following result identifies the adjoint processes in terms of spatial derivatives of this value function $V(t,x)$.
	\begin{Theorem}[Relation with dynamic programming principle] \label{The2} Suppose that the following conditions hold.
		\begin{enumerate}
			\item The value function $V \in C^{1,3}([0,T)\times \R^n) \cap C^{0,1}([0,T] \times \R^n)$ is a classical solution of the HJB equation
				\begin{align} \displaystyle \partial_t&V(t,x) + \sup_{a\in A} \bigg\{ f(t,x,a) + b(t,x,a)^\top\nabla_xV(t,x)  + \frac12 \operatorname{Tr} \left[ \sigma(t,x,a)\sigma(t,x,a)^\top \nabla_x^2V(t,x) \right] \nonumber \\[-0.5mm] \displaystyle &\hspace{-2mm}+ \lambda(t) \int_E \Bigl[ V\bigl(t,x+\gamma(t,x,z,a)\bigr) - V(t,x) - \nabla_xV(t,x)^\top\gamma(t,x,z,a) \Bigr]\nu(dz) \bigg\} =0, \hspace{-1mm} \label{eq:HJB}  \end{align}
  for every \((t,x)\in[0,T)\times\mathbb R^n\), with terminal condition
  \(
  V(T,x)=F(x) .
  \)
  Moreover, the nonlocal integral in \eqref{eq:HJB} is finite for every
  \((t,x,a)\in[0,T)\times\mathbb R^n\times A\).
			\item  There exists an optimal admissible Markov feedback control
			\(
			\alpha_t^*
			=
			\alpha^*(t,X^*_{t-})
			\)
			such that $\alpha^*(t,x)$ attains the supremum in
			\eqref{eq:HJB} for all $(t,x) \in [0,T] \times \R^n$,  and such that the feedback control \( \alpha_t^* = \alpha^*(t,X_{t-}^*) \) is admissible. 
			\item The processes defined by \eqref{ppp}--\eqref{rrr}
			belong to
			\(
			\mathcal S^2_{\mathbb F}(0,T;\mathbb R^n)
			\times
			L^2_{\mathbb F}(0,T;\mathbb R^{n\times d})
			\times
			L^2_{\nu,\mathbb F}(0,T;\mathbb R^n).
			\)
		\end{enumerate}
		Then, the adjoint processes 
		\begin{align}
			p^*_t &=  \nabla_x V(t,X^*_{t}) \label{ppp}
			\\
			q_t^*
			&=
			\nabla_x^2 V(t,X_{t-}^*)
			\,\sigma(t,X_{t-}^*,\alpha^*(t,X_{t-}^*)),\label{qqq}
			\\
			r^*_t(z) &=  \nabla_x V(t, X^*_{t-} + \gamma(t,X^*_{t-},z, \alpha^*(t,X^*_{t-}))) - \nabla_x V(t,X^*_{t-}) \label{rrr}
		\end{align}
		solve the adjoint BSDE \eqref{adjoint} corresponding to the optimal pair $( X^*, \alpha^*)$.
	\end{Theorem}
	This theorem is stated without proof as it is an immediate$^{\text{\ref{test}}}$ consequence of Theorem 3.1.\ of \cite{framstad2004sufficient}.  It follows from Theorem~\ref{The2} that, upon defining the decoupling field
	\(
	p:[0,T]\times \R^n\to\mathbb{R}^n,
	\,
	p(t,x):=\nabla_xV(t,x),
	\)
	the adjoint processes can be represented as
	\begin{align*}
		q^*_t &= \nabla_x p(t,X^*_{t-}) \, \sigma(t,X^*_{t-}, \alpha^*(t,X^*_{t-}))\\
		r^*_t(z) & = p(t, X^*_{t-} + \gamma(t,X^*_{t-}, z,\alpha^*(t,X^*_{t-}))) - p(t, X^*_{t-}) \, .
	\end{align*}
	These identities will play a key role in the construction of our numerical algorithm.   Although Theorem~\ref{The2} identifies the adjoint processes through spatial
	derivatives of the value function, computing these quantities directly through automatic differentiation is
	typically expensive. Evaluating
	$V(t,x)$
	at a given state $(t,x)$ indeed requires estimating a conditional expectation by
	simulating many trajectories starting from $(t,x)$. Therefore, if the states
	$X^\alpha_t$ themselves are generated by an outer simulation, then
	approximating $\nabla_x V(t,X^\alpha_{t})$ leads to a nested Monte Carlo
	procedure: for each outer sample path and each time step, a new
	collection of inner trajectories must be simulated from the current state. This computational bottleneck motivates the pathwise BPTT construction developed
	below. Instead of estimating $\nabla_x V(t,x)$ separately at every state, we
	differentiate the realized (pathwise) reward along each simulated state trajectory. We show in the following section that these
	pathwise state derivatives provide cheap noisy targets, whose conditional expectation recovers a discrete-time adjoint process  that converges  in the limit $\Delta t \to 0$ to the true adapted adjoint  satisfying the BSDE \eqref{adjoint} and its Markov representation \eqref{ppp}.

	\section{Methodology} \vspace{-1mm}
	
	Fix a Markov feedback control $\alpha : [0,T] \times \R^n \to A$ and a time grid $t_i=i\Delta t$, $i=0,\dots,N$ with $t_N = T$, and define
	\[
	\Delta W_i := W_{t_{i+1}}-W_{t_i},\qquad
	\tilde N_i(dz):=\tilde N((t_i,t_{i+1}],dz).
	\]
	Let \((X_i)_{i=0}^N\) denote the Euler--Maruyama\footnote{For a general L\'evy measure satisfying
		\(\nu(E)=\infty\), the jump integral has to be suitably truncated  in the numerical scheme.} discretization of the state equation \eqref{sde}, \begin{align*} \small
		X_{i+1} 
		={}&
		X_i
		+
		b(t_i,X_i,a_i )\Delta t
		+
		\sigma(t_i, X_i,a_i )\Delta W_i
		+
		\int_E
		\gamma(t_i,X_i,z,a_i )
		\,\widetilde N_i(dz), \quad \, i=0,\ldots,N-1,  \hspace{-1mm}
		\label{eq:jump-scheme1}
	\end{align*}  with $a_i:=\alpha(t_i,X_i)$. To compute state sensitivities while keeping the realized controls fixed, we introduce the stop-gradient control
	\begin{equation}
		a_i^{\mathrm{sg}}
		:=
		\alpha(t_i,\operatorname{sg}[X_i]) .
		\label{eq:stopgrad-control}
	\end{equation}
	The stop-gradient operator $\operatorname{sg}$ acts as the identity in the forward pass but has zero derivative in the backward pass. Consequently, $a_i^{\mathrm{sg}}$ and $a_i$ have the same numerical value, but no derivative is propagated through the map $x\mapsto\alpha(t_i,x)$. For \(i\in\{0,\ldots,N\}\) and \(x\in\mathbb R^n\), let
	\((X_j^{i,x})_{j=i,\ldots,N}\) be the \textit{frozen-control} continuation starting
	from \(x\) at time \(t_i\). It is defined for $j=i,\ldots,N-1$ by
	\begin{align}
		X_{j+1}^{i,x}
		={}&
		X_j^{i,x}
		+
		b(t_j,X_j^{i,x},a_j^{\mathrm{sg}})\Delta t
		+
		\sigma(t_j,X_j^{i,x},a_j^{\mathrm{sg}})\Delta W_j
		+
		\int_E
		\gamma(t_j,X_j^{i,x},z,a_j^{\mathrm{sg}})
		\,\widetilde N_j(dz)
		\label{eq:jump-scheme}
	\end{align}
	with $X_i^{i,x}=x$. Hence, the continuation \((X_j^{i,x})_{j=i}^N\) uses the same future
	noise increments \((\Delta W_j,\widetilde N_j)_{j=i}^{N-1}\) and the same
	realized action sequence \((a^{\mathrm{sg}}_j)_{j=i}^{N-1}\) as the reference trajectory, but has different gradients $\nabla_x$.
	In particular, since the stop-gradient operation does not change the forward
	values,  it follows pathwise that
	\(
	X_j^{i,X_i}=X_j
	\),  $j=i,\ldots,N.$
	We then define the random future payoff at $(t_i,x)$,
	\begin{equation}
		\label{eq:future-payoff}    
		\widehat{J}_i^\Delta(x) 
		:=
		\sum_{j=i}^{N-1}
		f\!\left(
		t_j,
		X^{i,x}_j,
		a_j^{\operatorname{sg}}
		\right)\Delta t
		+
		F(X^{i,x}_N) \, .
	\end{equation}
	For every fixed $x \in \R^n$,   $\widehat{J}_i^\Delta(x)$ is an $\mathcal{F}_T$-measurable random variable. The associated pathwise \emph{BPTT labels} are defined  by
	\begin{equation}
		\label{eq:lambda-def}
		\widehat{p}_i   =  \nabla_x \widehat{J}_i^\Delta(x) \Big|_{x=X_i} , \, \quad i=0,\ldots,N.
	\end{equation}
	Thus, each $\widehat{p}_i$ is the  $\mathcal F_T$-measurable pathwise sensitivity of the frozen future payoff with respect to the state $X_i$ at time \(t_i\), with all subsequent realized controls and noise increments held fixed.
	By  the pathwise chain rule, the BPTT labels satisfy the backward recursion
	\begin{equation}
		\label{eq:bptt-recursion}
		\widehat p_i
		=
		\nabla_x f(t_i, X_i, a_i^{\operatorname{sg}}) \, \Delta t
		+
		\left(\nabla_xX^{i,X_i}_{i+1}\right)^\top
		\widehat p_{i+1},
		\qquad i=N-1,\dots,0,
	\end{equation}
	with terminal condition $
	\widehat p_N = \nabla_x F(X_N)$, and where we use
	$$\nabla_xX^{i,X_i}_{i+1} = I
	+
	\nabla_x b(t_i,X_i,a^{\operatorname{sg}}_i)\Delta t
	+
	\nabla_x\sigma(t_i,X_i,a^{\operatorname{sg}}_i)\Delta W_i
	+
	\int_E
	\nabla_x\gamma(t_i,X_i,z,a^{\operatorname{sg}}_i)\,
	\widetilde N_i(dz).$$
	The next result shows that projecting the pathwise BPTT labels onto the current state produces an \(\mathcal F_{t_i}\)-measurable Markovian quantity.
	
	\begin{Proposition}\label{prop:noisy}
		Suppose that \(\widehat p_i \in L^2(\Omega,\mathcal F_T;\mathbb R^n)\) and define
		\[
		p_i
		:=
		\mathbb E\!\left[
		\widehat p_i
		\,\middle|\,
		\sigma(X_i)
		\right].
		\]
		Then there exists a Borel function
		\(p^\Delta_i:\R^n\to\mathbb R^n\), unique \(\mathbb P_{X_i}\)-a.e., such that 
		\begin{equation} \label{def_pi}
			p_i = p^\Delta_i(X_i), \qquad \mathbb{P}\text{-a.s.}
		\end{equation}
		In particular,   $p_i \in L^2(\Omega,\mathcal{F}_{t_i};\R^n)$.  Moreover, \(p^\Delta_i\) is the unique minimizer $g^*:\R^n \to \R^n$ of the regression
		\begin{equation}
			\label{eq:l2-projection-pi}
			\inf_{g}
			\mathbb E\!\left[
			\left|g(X_i)-\widehat p_i\right|_2^2
			\right],
		\end{equation}
		where the infimum is taken over all Borel functions
		\(g:\R^n\to\mathbb R^n\). Equivalently, $ p_i = g^*(X_i)$, $\mathbb{P}\text{-a.s.}$.
	\end{Proposition}
	See Appendix \ref{proof:noisy} for the proof. We emphasize that the interpretation of \(p_i^\Delta\) as the state sensitivity
	of a discrete value function is only valid under strong additional smoothness and integrability assumptions.  Such an identification is however not required in our construction since the adjoint field $p_i^\Delta$ is defined directly through the conditional $L^2$-projection \eqref{def_pi}.
	In the remainder, we use the notation $\E_i[\cdot] := \E[\, \cdot \, | \, \mathcal{F}_{t_i}]$ and focus on  the discretized  adjoint process $(p_i)_{i\in \N}$, which admits the following predictable representation.
	\begin{Proposition}
		\label{prop:step-averaged-martingale-coefficients}
		Let $i\in\{0,\ldots,N-1\}$ and $p_{i+1}\in L^2(\Omega,\mathcal F_{t_{i+1}};\mathbb R^n)$.
		\begin{enumerate}[label=\textup{(\roman*)}]
			\item Then there exist unique predictable processes
			\begin{equation}
				\label{eq:marting_coeff}
				q^{i+1}\in L^2_{\mathbb F}(t_i,t_{i+1};\mathbb R^{n\times d}),
				\qquad
				r^{i+1}\in L^2_{\nu,\mathbb F}(t_i,t_{i+1};\mathbb R^n),
			\end{equation}
			such that
			\begin{equation}
				\label{eq:pred_repr_markov}
				p_{i+1}
				=
				\mathbb E_i[p_{i+1}]
				+
				\int_{t_i}^{t_{i+1}} q_s^{i+1}\,dW_s
				+
				\int_{t_i}^{t_{i+1}}\int_E
				r_s^{i+1}(z)\,\widetilde N(ds,dz).
			\end{equation}
			\item Define the step-averaged Brownian coefficient by
			\begin{equation}
				\label{eq:qbar-def-markov}
				\bar q_{i+1}
				:=
				\frac{1}{\Delta t}
				\mathbb E_i\!\left[p_{i+1}\Delta W_i^\top\right] \, .
			\end{equation}
			Then,  $\bar{q}_{i+1} \in L^2(\Omega, \mathcal{F}_{t_i}; \R^{n\times d})$ and
			\begin{equation}
				\label{eq:qbar-time-average}
				\bar q_{i+1}
				=
				\frac{1}{\Delta t}
				\mathbb E_i\!\left[
				\int_{t_i}^{t_{i+1}} q_s^{i+1}\,ds
				\right] .
			\end{equation}
			\item Let \( \Lambda_i := \int_{t_i}^{t_{i+1}}\lambda(s)\,ds>0. \) Then there exists a unique element \( \bar r_{i+1} \in L^2\left( \Omega,\mathcal F_{t_i}; L^2(E,\nu;\mathbb R^n) \right) \) such that, for every \( h \in L^2\left( \Omega,\mathcal F_{t_i}; L^2(E,\nu;\mathbb R^n) \right), \) \begin{equation} \label{eq:rbar-def-markov} \mathbb E_i\left[ \left\langle p_{i+1}, \int_E h(z)\,\widetilde N_i(dz) \right\rangle \right] = \Lambda_i \int_E \left\langle \bar r_{i+1}(z),h(z) \right\rangle \nu(dz), \qquad \mathbb P\text{-a.s.} \end{equation} Moreover, \(\bar r_{i+1}\) is given by \begin{equation} \label{eq:rbar-time-average} \bar r_{i+1} = \frac{1}{\Lambda_i} \mathbb E_i\left[ \int_{t_i}^{t_{i+1}} \lambda(s)r_s^{i+1}(\cdot)\,ds \right] \, . \end{equation}
		\end{enumerate}
	\end{Proposition}
	
	See Appendix \ref{proof:coeff} for the proof. Thus, \(\bar q_{i+1}\) is the conditional time average of the Brownian
	martingale coefficient over \((t_i,t_{i+1}]\), whereas
	\(\bar r_{i+1}\) is its jump counterpart, obtained as a conditional
	intensity-weighted time average. We now show that the projected BPTT adjoint $(p_i)_{i=0}^N$ satisfies a discrete backward recursion with the same structural form as the continuous-time PMP adjoint BSDE \eqref{adjoint} from Theorem \ref{th:PMP}. 
	\begin{Proposition}
		\label{prop:jump-envelope-markov}
		Define the discrete Hamiltonian
		\[
		\begin{aligned}
			H_i^\Delta(x,a,p,q,r)
			&:=
			f(t_i,x,a)
			+
			b(t_i,x,a)^\top p
			+
			\mathrm{Tr}\!\left[\sigma(t_i,x,a)^\top q\right] 
			+
			\frac{\Lambda_i}{\Delta t}
			\int_E
			\gamma(t_i,x,z,a)^\top r(z)\,\nu(dz).
		\end{aligned}
		\]
		Then the discretized adjoint process satisfies 
		\begin{equation}
			\label{eq:lambda-fin1-markov}
			p_i
			=
			\mathbb E_i[p_{i+1}]
			+
			\nabla_x H_i^\Delta
			\bigl(X_i,a^{\operatorname{sg}}_i,\mathbb E_i[p_{i+1}],\bar q_{i+1},\bar r_{i+1}\bigr)
			\Delta t \, .
		\end{equation}
		Consequently, using the predictable representation \eqref{eq:pred_repr_markov},
		\begin{equation}
			\label{eq:lambda-fin2-markov}
			\begin{aligned}
				p_{i+1}
				&=
				p_i
				-
				\nabla_x H_i^\Delta
				\bigl(X_i,a^{\operatorname{sg}}_i,\mathbb E_i[p_{i+1}],\bar q_{i+1},\bar r_{i+1}\bigr)
				\Delta t  
				+
				\int_{t_i}^{t_{i+1}} \hspace{-1.5mm} q_s^{i+1}\,dW_s
				+
				\int_{t_i}^{t_{i+1}}\hspace{-1.5mm}\int_E \hspace{-1mm}
				r_s^{i+1}(z)\,\widetilde N(ds,dz)
				\, .
			\end{aligned}
		\end{equation}
	\end{Proposition}
	We refer to Appendix \ref{proof:envelope} for the proof. We finally show that, as $\Delta t\to 0$, the discrete-time  adjoint process
	$(p_i)_{i=0}^N$ converges weakly to the solution of the continuous-time PMP adjoint equation \eqref{adjoint}.
	\begin{Proposition}
		\label{prop:bptt-pmp-convergence-jumps}
		Let $\Delta t=T/N$, $t_i=i\Delta t$, and  define the piecewise-constant 
		interpolations 
		\[
		\bar X_t^\Delta:=X_i,\qquad
		\bar a_t^\Delta:=a^{\operatorname{sg}}_i,\qquad
		\bar p_t^\Delta:=p_i,  \qquad  t\in[t_i,t_{i+1}),
		\]
		with $\bar p_T^\Delta:=p_N$. Define the  interpolations of the martingale coefficients  \eqref{eq:marting_coeff} by
		\[
		\widehat q_t^\Delta:=q_t^{i+1},
		\qquad
		\widehat r_t^\Delta:=r_t^{i+1},
		\qquad t\in[t_i,t_{i+1}),
		\]
		and the piecewise-constant interpolations of the averaged coefficients \eqref{eq:qbar-time-average}--\eqref{eq:rbar-time-average} by
		\[
		\bar q_t^\Delta:=\bar q_{i+1},
		\qquad
		\bar r_t^\Delta:=\bar r_{i+1},
		\qquad t\in[t_i,t_{i+1}).
		\]
		Assume that the following conditions hold.
		
		\begin{enumerate}[label=\textup{(\roman*)}]
			\item \label{itemize:1}
		Assumption~\ref{Ass1} holds. In addition, the maps
		\(
		(t,x,a)\mapsto
		\nabla_x b(t,x,a),\,
		\nabla_x\sigma(t,x,a),\, \nabla_x \gamma(t,x,\cdot, a),$ $ 
		\nabla_x f(t,x,a)
		\)
		are jointly continuous, and 
		\(
		x\mapsto \nabla_xF(x)
		\)
		is continuous.
			
			\item \label{itemize:2} There exist
			\(
			X\in\mathcal S^2_{\mathbb F}(0,T;\mathbb R^n)$ and $
			a\in L^2_{\mathbb F}(0,T;\mathbb R^m)
			\)
			such that
			\[
			\bar X^\Delta\rightarrow X_{-}
			\quad\text{in }
			L^2(\Omega\times[0,T];\mathbb R^n), \quad 
		X_N \longrightarrow X_T
			\quad\text{in }
			L^2(\Omega,\mathcal F_T;\mathbb R^n),
			\]
			and
			\[
			\bar a^\Delta\rightarrow a
			\quad\text{in }
			L^2(\Omega\times[0,T];\mathbb R^m).
			\]
			Moreover, the limiting control is Markovian: there exists a  measurable feedback map
			$\alpha:[0,T]\times\mathbb R^n\to A$ such that 
			\(
			a_t=\alpha(t,X_{t-})
			\) $dt\otimes d\mathbb P\text{-a.e.}$

		\end{enumerate}

		Then there exist predictable processes
		\(q\in L^2_{\mathbb F}(0,T;\mathbb R^{n\times d})\), 
		\(r\in L^2_{\nu,\mathbb F}(0,T;\mathbb R^n)\), and an adapted process \(p\), such that
		\begin{align*}
			\bar p^\Delta\rightharpoonup p
			\quad&\text{weakly in }
			L^2(\Omega\times[0,T];\mathbb R^n),
			\\
			\widehat q^\Delta, \,
			\bar q^\Delta \rightharpoonup q
			\quad &\text{weakly in }L^2_{\mathbb{F}}(0,T;\R^{n\times d}),
			\\
			\widehat r^\Delta , \,
			\bar r^\Delta \rightharpoonup r
			\quad &\text{weakly in }
			L^2_{\nu, \mathbb{F}}(0,T;\R^n).&
		\end{align*}
		The limiting adjoint process admits a càdlàg adapted modification,  relabeled by $p$, such that $ (p,q,r)
		\in
		\mathcal S^2_{\mathbb F}(0,T;\R^n)
		\times
		L^2_{\mathbb F}(0,T;\R^{n\times d})
		\times
		L^2_{\nu,\mathbb F}(0,T;\R^n)$   is the unique adapted solution of  the PMP adjoint BSDE
		\[
		dp_t
		=
		-\nabla_xH(t,X_{t-},a_t,p_t,q_t,r_t)\,dt
		+
		q_t\,dW_t
		+
		\int_E r_t(z)\,\widetilde N(dt,dz),
		\qquad
		p_T=\nabla_xF(X_T).
		\]
		If, in addition, the limiting Markov adjoint admits a smooth decoupling field
		$p\in C^{1,2}([0,T]\times \R^n)$, so that
		\(
		p_t=p(t,X_t),
		\)
		then for
		$dt\otimes d\mathbb P\otimes\nu(dz)$-a.e.\
		$(t,\omega,z)$, we have
		\begin{equation}
			\label{eq:q-limit-markov-jump}
			q_t
			=
			\nabla_xp(t,X_{t^-})\,
			\sigma(t,X_{t^-},a_t),
		\end{equation}
		\begin{equation}
			\label{eq:r-limit-markov-jump}
			r_t(z)
			=
			p\!\left(t,X_{t-}+\gamma(t,X_{t-},z,a_t)\right)
			-
			p(t,X_{t-}) .
		\end{equation}

	\end{Proposition}
	
	See Appendix \ref{proof: convergence} for the proof. Establishing strong convergence in Proposition~\ref{prop:bptt-pmp-convergence-jumps} would require additional and restrictive grid-consistency assumptions that are generally difficult to verify in practice, see \cite{huh2025pontryagin}. More precisely, this requires that the terminal error converges strongly, \( \mathbb E|p_N-p_T|^2\to0, \) and that the continuous-time adjoint admits a time-discrete recursion consistent with the discrete BPTT--PMP scheme, with accumulated truncation error converging to zero in \(L^2\). Moreover, the convergence result should be understood as a time-discretization result for the ideal projected adjoint process. It does not directly imply convergence of the fully trained neural-network algorithm, since the latter also involves statistical error, neural approximation error, and optimization error. A complete convergence theory for full neural jump-FBSDE solvers still remains largely open \citep{alasseur2024deep}.
	\\[3mm]
From these theoretical results, we now introduce our numerical algorithm which relies on a continuous-time
	policy-improvement principle \citep{jacka2017policy}. Starting from a current Markov feedback control $\alpha^k$, one first evaluates  the corresponding adjoint quantities of the value function induced by $\alpha^k$, and then defines an improved policy $\alpha^{k+1}$ by maximizing the Hamiltonian pointwise with respect to the control. Our method can be viewed as an approximate, data-driven implementation of this principle. In the first step, the current adjoint field $p^\Delta$ is estimated from the $L^2$-projection \eqref{eq:l2-projection-pi}. In the second step, the Hamiltonian maximization condition \eqref{max1} is enforced  through a pointwise stationarity loss.

	\section{Numerical method} \vspace{-2mm}
	\textbf{Step 1: Discrete adjoint approximation through supervised regression.} Our numerical scheme first builds on Proposition~\ref{prop:noisy}, which provides a computationally efficient procedure for approximating the discrete adjoint process $(p_i)_{i=0}^N$ satisfying the discrete-time adjoint recursion  \eqref{eq:lambda-fin2-markov}. 
More precisely, from a single set of \(M\) simulated trajectories with \(N\) time steps, one obtains the supervised samples \[ \left( t_i, X_i^{(m)}, \widehat p_i^{(m)} := \nabla_x \widehat J_i^\Delta(x)\big|_{x=X_i^{(m)}} \right), \qquad m=1,\ldots,M,\quad i=0,\ldots,N . \] Thus, the pathwise BPTT labels provide \(M(N+1)\) adjoint targets $\widehat{p}^{(m)}_i$ from only \(M\) simulated rollouts, without requiring a separate nested MC simulation from each  individual state \((t_i,X_i^{(m)})\). We approximate the discrete Markov adjoint field \(p_i^\Delta(x)\) by a single global-in-time neural network
	\[
	p_\phi^\Delta:
	[0,T]\times\mathbb R^n
	\rightarrow
	\mathbb R^n.
	\]
	In view of the conditional \(L^2\)-projection from Proposition~\ref{prop:noisy}, its parameters are trained by solving
	\begin{equation}
		\label{eq:adjoint-regression-loss}
		\phi^*
		\in
		\argmin_{\phi}
		\mathcal L_{\mathrm{reg}}(\phi),
		\qquad
		\mathcal L_{\mathrm{reg}}(\phi)
		:=
		\frac{1}{M(N+1)}
		\sum_{m=1}^M
		\sum_{i=0}^{N}
		\left|
		p_\phi^\Delta(t_i,X_i^{(m)})
		-
		\widehat p_i^{(m)}
		\right|^2.
	\end{equation}
	
	Therefore, this regression problem together with Proposition \ref{prop:jump-envelope-markov} ensures that the neural network $p_\phi^\Delta$ approximates the solution of a discrete-time version  of the adjoint BSDE.
	\\[3mm]
	\textbf{Step 2: Control update via Hamiltonian stationarity.} The pointwise maximization condition \eqref{max1} implies that, whenever \(A\) is open, the maximizing action lies in the interior of \(A\), and the Hamiltonian is differentiable with respect to the control, the optimal feedback $\alpha^*$ satisfies
	\begin{align*} \small
	0 &= 	\nabla_aH
	\bigl(
	t,
	x,
	\alpha^*(t,x),
	p(t,x),
	q(t,x),
	r(t,x,\cdot)
	\bigr) 
	\\
	&= \nabla_a f(t,x,\alpha^*(t,x)) + \nabla_a b(t,x,\alpha^*(t,x))^\top p(t,x)
	\\
&\hspace{5mm}+ \operatorname{Tr}[\nabla_a \sigma(t,x,\alpha^*(t,x))^\top q(t,x)] + \lambda(t) \int_E \nabla_a \gamma(t,x,z,\alpha^*(t,x))^\top r(t,x,z)  \nu(dz)
	.
	\end{align*}
	For a constrained control set, this condition can instead be replaced by the corresponding  Karush--Kuhn--Tucker condition\footnote{E.g.\ if $A$ is compact and convex and $H$ is concave in $a$, the maximizer satisfies $\langle \nabla_a H(t,x,\alpha,p,q,r), a-\alpha \rangle \leq 0 $ for $a\in A$.}.  We approximate the Markov feedback control by a neural network
	\[
	\alpha_\theta:
	[0,T)\times\mathbb R^n
	\rightarrow
	A.
	\] Once the adjoint network \(p_{\phi^*}^\Delta\) has been trained, its parameters are kept fixed during the control-update step. Using Theorem~\ref{The2} and Proposition~\ref{prop:bptt-pmp-convergence-jumps}, we then reconstruct the Brownian and jump adjoint coefficients from this learned adjoint $p_\phi^\Delta$ by setting
	\begin{equation} \label{eq:algo1}
		q_{\phi,\theta}(t,x)
		:=
		\nabla_x p_\phi^\Delta(t,x)\,\sigma(t,x,\alpha_\theta(t,x)),
	\end{equation}
	\begin{equation} \label{eq:algo2}
		r_{\phi,\theta}(t,x,z)
		:=
		p_\phi^\Delta\!\bigl(t,x+\gamma(t,x,z, \alpha_\theta(t,x))\bigr)
		-
		p_\phi^\Delta(t,x) \, .
	\end{equation}
	
	To enforce this pointwise stationary condition along the simulated state trajectories, we define $X_i^{\operatorname{sg},(m)} := \operatorname{sg}[X_i^{(m)}]$ where $\operatorname{sg}$  again denotes the stop-gradient operator with respect to $x$. This prevents gradients from being propagated through the simulated state trajectories, while preserving differentiation through the control network $\alpha_\theta$.
	We then minimize the empirical stationarity loss 
	\begin{equation} \small
		\begin{aligned}
			&\mathcal L_{\mathrm{stat}}(\theta)
			:=\\
			&\hspace{5mm}
			\frac{1}{MN}\sum_{m=1}^M\sum_{i=0}^{N-1}
			\left|
			\nabla_a H\Bigl(
			t_i,X_i^{\operatorname{sg},(m)},\alpha_\theta(t_i,X_i^{\operatorname{sg},(m)}),
			p_\phi^\Delta(t_i,X_i^{\operatorname{sg},(m)}),
			q_{\phi,\theta}(t_i,X_i^{\operatorname{sg},(m)}),
			r_{\phi,\theta}(t_i,X_i^{\operatorname{sg},(m)},\cdot)
			\Bigr)
			\right|^2, \label{eq:stationary_algo}
		\end{aligned}
	\end{equation}
	by solving \vspace{-3mm}
	\[
	\theta^* \in \arg\min_\theta \mathcal L_{\mathrm{stat}}(\theta).
	\]
	The full Deep PMP scheme is summarized in Algorithm \ref{alg:bptt-pmp}.\vspace{3mm}
	\begin{algorithm} \small
		\caption{Deep PMP with BPTT adjoint regression}
		\label{alg:bptt-pmp}
		\begin{algorithmic}[1]
			\State \textbf{Input:} Number of epochs $k^*$, batch size $M_1$, number of time steps $N$,
			jump quadrature size $M_2$, initial-state distribution $\rho$, initial network parameters $\phi^{(0)}$ and
			$\theta^{(0)}$, and  learning rates
			$\eta_p,\eta_\alpha>0$.
			
			\State Set $\Delta t:= T/N, t_i=i\Delta t$,  $\Lambda_i:=\int_{t_i}^{t_{i+1}}\lambda(s)\,ds ,$   $i=0,\dots,N-1$.
			\State Let \( \bar\nu:=\nu(E), \,  \mu(dz):=\nu(dz) / \bar\nu. \)
			
			\For{epoch $k=0$ to $k^*-1$}
			
			\Statex \textbf{Step 1: Approximate the discrete adjoint through supervised regression.}
			
			\For{$m=1$ to $M_1$}
			\State Sample $X_0^{(m)} \sim \rho$.
			
			\For{$i=0$ to $N-1$}
			\State Evaluate the fixed control at step $k$ with stopped state gradient, \( a_i^{\operatorname{sg},(m)} := \alpha_{\theta^{(k)}}\!\left( t_i, \operatorname{sg}[X_i^{(m)}] \right). \)
			\State Sample \( \Delta W_i^{(m)} \sim \mathcal N(0,\Delta t\,I_d), \, N_i^{(m)} \sim \operatorname{Poisson}(\Lambda_i\bar\nu), \) and \( Z_{i,1}^{(m)},\ldots,Z_{i,N_i^{(m)}}^{(m)} \overset{\mathrm{i.i.d.}}{\sim} \mu. \)
			\State Sample independent quadrature points \( \zeta_{i,1}^{(m)},\ldots,\zeta_{i,M_2}^{(m)} \overset{\mathrm{i.i.d.}}{\sim} \mu \) and approximate the compensator by\vspace{-2.3mm} \[ \widehat\Gamma_i^{(m)} := \frac{\bar\nu}{M_2} \sum_{j=1}^{M_2} \gamma\!\left( t_i, X_i^{(m)}, \zeta_{i,j}^{(m)}, a_i^{\operatorname{sg},(m)} \right).\vspace{-2.3mm} \]
			\State Update the state by the compensated jump \vspace{-3mm}scheme {\footnotesize	\[
			\begin{aligned}
				X_{i+1}^{(m)} \hspace{-0.4mm}
				&= \hspace{-0.4mm}
				X_i^{(m)} \hspace{-1mm}  
				+b(t_i,X_i^{(m)} \hspace{-0.5mm},a_i^{\operatorname{sg},(m)})\Delta t \hspace{-0.4mm}
				+ \hspace{-0.4mm}
				\sigma(t_i,X_i^{(m)} \hspace{-0.46mm},a_i^{\operatorname{sg},(m)})\Delta W_i^{(m)} \hspace{-1mm}                           + \hspace{-0.69mm}  
				\sum_{\ell=1}^{N_i^{(m)}} \hspace{-1mm}  
				\gamma(t_i,X_i^{(m)} \hspace{-0.49mm},Z_{i,\ell}^{(m)},a_i^{\operatorname{sg},(m)}) \hspace{-0.4mm}
				- \hspace{-0.4mm}
				\Lambda_i\widehat\Gamma_i^{(m)} \hspace{-0.4mm}.
			\end{aligned}
			\]}
			\EndFor
			\State Compute the realized payoff: $\widehat{J}^{\Delta,(m)} = \sum_{i=0}^{N-1} f(t_i, X_i^{(m)}, a_i^{\operatorname{sg},(m)}) \Delta t +F(X_N^{(m)})$.\vspace{1mm}
			
			\State Compute the BPTT adjoint targets by reverse-mode AD: $\widehat{p}^{(m)}_i =   \nabla_x \widehat{J}^{\Delta, (m)} \Big|_{x=X^{(m)}_i}$, $i=N,\ldots, 0$.
			\EndFor

			\State Update the \vspace{-2mm}adjoint network for fixed control weights $\theta^{(k)}$,
			\[
			\phi^{(k+1)}
			=
			\phi^{(k)}
			-
			\eta_p
			\nabla_\phi
			\widehat{\mathcal L}_{\text{reg}}(\phi^{(k)}), \qquad        \widehat{\mathcal L}_{\text{reg}}(\phi)
			:=
			\frac{1}{M_1(N+1)}
			\sum_{m=1}^{M_1}
			\sum_{i=0}^{N}
			\left|
			p_\phi^\Delta(t_i,X_i^{(m)})
			-
			\widehat p_i^{(m)}
			\right|^2 .
			\]
			
			\Statex \textbf{Step 2: Train the feedback control by PMP stationarity.}
			
			\For{$m=1$ to $M_1$ and $i=0$ to $N-1$}
			\State  Let $X_i^{\operatorname{sg},(m)} = \operatorname{sg}[X_i^{(m)}]$  and the current control $a_i^{\operatorname{sg},(m)} := \alpha_{\theta^{(k)}}\!\left( t_i, X_i^{\operatorname{sg},(m)} \right). $
			\State Compute the Hamiltonian-stationarity residual \vspace{-1mm}
		{\footnotesize	\begin{align} 
				\nabla_a H_i^{(m)}
				&:=
				\nabla_a f(t_i,X_i^{\operatorname{sg},(m)},a_i^{\operatorname{sg},(m)})
				+
				\nabla_a b(t_i,X_i^{\operatorname{sg},(m)},a_i^{\operatorname{sg},(m)})^\top 	p_{\phi^{(k+1)}}^\Delta(t_i,X_i^{\operatorname{sg},(m)}) \nonumber
				\\
				&\hspace{-15mm}+
				\operatorname{Tr}
				\!\left[
				\nabla_a\sigma(t_i,X_i^{\operatorname{sg},(m)},a_i^{\operatorname{sg},(m)})^\top 	\nabla_x p_{\phi^{(k+1)}}^\Delta(t_i,X_i^{\operatorname{sg},(m)})
				\,         \sigma(t_i,X_i^{\operatorname{sg},(m)},a_i^{\operatorname{sg},(m)})
				\right] \label{eq:grad}
				\\
				&\hspace{-15mm}+
				\frac{	\lambda(t_i)\bar{\nu}}{M_2}
				\sum_{j=1}^{M_2}
				\nabla _a \gamma(t_i,X_i^{\operatorname{sg},(m)},\zeta_{i,j}^{(m)},a_i^{\operatorname{sg},(m)})^\top \hspace{-0.45mm}
			\Big( p_{\phi^{(k+1)}}^\Delta  \hspace{-0.25mm}
			\!\left(
			t_i,
			X_i^{\operatorname{sg},(m)}  \hspace{-0.85mm}
			+  \hspace{-0.25mm}
			\gamma(t_i,X_i^{\operatorname{sg},(m)} \hspace{-0.85mm},\zeta_{i,j}^{(m)},a_i^{\operatorname{sg},(m)}\hspace{-0.35mm})
			\right)  \hspace{-0.35mm}
			-  \hspace{-0.25mm}
			p_{\phi^{(k+1)}}^\Delta(t_i,X_i^{\operatorname{sg},(m)} \hspace{-0.25mm}) \Big) . \nonumber
			\end{align}}
			
			\EndFor
			
			\State Update the control network  for fixed adjoint weights $\phi^{(k+1)}$, 
			\[
			\theta^{(k+1)}
			=
			\theta^{(k)}
			-
			\eta_\alpha
			\nabla_\theta
			\widehat{\mathcal L}_{\mathrm{stat}}(\theta^{(k)}), \qquad         \widehat{\mathcal L}_{\mathrm{stat}}(\theta)
			:=
			\frac{1}{M_1N}
			\sum_{m=1}^{M_1}
			\sum_{i=0}^{N-1}
			\left|\nabla_aH_i^{(m)}\right|^2 .
			\]
			\EndFor
			
			\State \Return Trained networks $p_{\phi^{(k^*)}}^\Delta$ and
			$\alpha_{\theta^{(k^*)}}$.
			
		\end{algorithmic}
	\end{algorithm} 
	\newpage
	\textbf{Remark 1.} Because the discretized dynamics are causal\vspace{-1mm}, the reward terms accumulated before time \(t_i\) do not depend on the intermediate state \(X_i^{(m)}\). Consequently, a single reverse-mode automatic differentiation (AD) pass of the realized objective
	$\widehat{J}^{\Delta,(m)}$ along the computational graph of the trajectory yields directly all
	labels
	$
	\widehat p_i^{(m)}
	=  
	\nabla_x \widehat{J}_i^\Delta(x)
	\big|_{x=X_i^{(m)}},$ $ i=0,\ldots,N.$
	Thus, BPTT does not require a separate backward pass for each time index, but only one backward pass per simulated trajectory $m$. \\[2mm]
	\textbf{Remark 2.}
	When updating the control network $\alpha_\theta$, the sampled states $X_i^{\operatorname{sg},(m)} = \operatorname{sg}[X_i^{(m)}]$ are treated
	as fixed/frozen collocation points using the stop-gradient operator $\operatorname{sg}$. Hence, we do not backpropagate through the
	simulated state dynamics. Instead, the $\theta$-gradients are then only propagated through the current
action $
	a_i^{\operatorname{sg}, (m)}=\alpha_\theta(t_i,X_i^{\operatorname{sg},(m)})$
	and through the quantities that depend on this action. In particular, these
	gradients flow directly through the coefficients $f$, $b$, $\sigma$, $\gamma$, and
	also indirectly  through the induced coefficients $
	q_{\phi,\theta}$ and $
	r_{\phi, \theta}
	$
	via their dependence on $a_i^{\operatorname{sg},(m)}$, see Eq.\ \eqref{eq:grad}. This indeed corresponds to the pointwise enforcement of the Pontryagin stationarity
	condition \eqref{max1} through the loss \eqref{eq:stationary_algo} and avoids the unstable BPTT
	through the full rollout objective and the state dynamics.
	\\[2mm]
	\textbf{Remark 3.}
	The proposed Deep PMP algorithm differs from DPO
	methods \citep{han2016deep} in the way gradient information is generated and used.
	In DPO, the control network is trained by minimizing the
	empirical rollout cost $\widehat{J}^{\Delta,(m)} $ directly. Hence, every gradient update requires differentiating
	through the full time-discretized dynamics
	\(
	X_0\to X_1\to\cdots\to X_N.
	\)
	With $M$ simulated trajectories and $N$ time steps, each update has a computational graph of depth $N$ and memory requirements proportional to
	the full collection of simulated states and network activations.
	By contrast, Deep PMP separates adjoint estimation from control optimization, which has several practical advantages:
	\begin{enumerate}[label=\textup{(\roman*)}]
		\item The pathwise adjoint targets \( \big( \widehat p_i^{(m)} \big)_{i=0}^N \) are obtained from a single reverse-mode BPTT pass along each simulated trajectory. The Markov adjoint field is then learned through a standard supervised regression problem based on these targets, which is more stable than DPO on long-horizon objectives.

		\item  The repeated control updates do not require backpropagation
		through the full time-discretized state dynamics. This reduces memory requirements,
		shortens the depth of the computational graph, and improves optimization
		stability when the number of time steps $N$ is large, mitigating error
		accumulation as well as vanishing or exploding gradients over long horizons.
		\item  Our method explicitly incorporates the PMP into the learning procedure through the conditional $L^2$-projection in Proposition~\ref{prop:noisy} and the discrete adjoint recursion in Proposition~\ref{prop:jump-envelope-markov}. This allows the neural networks to learn adjoint information that is consistent with the physical and variational structure of the controlled process, and  to improve convergence and stability of the numerical scheme.

	\end{enumerate}

	\textbf{Remark 4.} Step 1 of Algorithm~\ref{alg:bptt-pmp} relies on MC estimates to approximate the compensator term associated with the jumps. When the jump-mark distribution has additional structure, alternative quadrature rules may be used, such as Gauss--Hermite quadrature for Gaussian marks or suitable transformations of lognormal marks. Thus, Deep PMP does not eliminate the numerical approximation of the jump integral. However, the Markovian representation \eqref{eq:r-limit-markov-jump} makes their incorporation particularly natural: the jump contribution is evaluated locally through the learned adjoint $p^\Delta_\phi$ within the pointwise Pontryagin stationarity condition. This contrasts with DPO, where each control update requires backpropagation through the full time-discretized jump SDE, including all subsequent jump realizations and compensator terms. 
	\section{Numerical experiments} \label{numerics}
	\subsection{Mean-variance portfolio optimization}
	\label{mean-var}
	We first consider a one-dimensional optimal control problem from \cite{framstad2004sufficient} as a simple benchmark for the proposed deep PMP method \ref{alg:bptt-pmp}. Let $ A, E = \R$, and consider the wealth process of a portfolio with one risk-free asset and one risky asset, 
	$$
	\begin{aligned}
		d X^\alpha_t & =\left(\rho X^\alpha_t+(\mu-\rho) \alpha_t\right) d t+\sigma \alpha_t d W_t+\int_{\mathbb{R}} z \alpha_t \tilde{N}(d z, d t) , \qquad t\in[0,T],  \\
	\end{aligned}
	$$
	with $X_0 \sim \mathcal{U}([0.5,2])$, $T=1,  \rho=0.05, \mu=0.1, \sigma=0.25$.  Here, $\tilde{N}$ denotes the corresponding compensated Poisson random measure  with compensator $\lambda \nu(dz)dt$, where $\lambda=2$ and  $\nu$ is the jump-size law of $Z = \exp (Y)-1$ with $Y \sim \mathcal{N}\left(0.15,0.15^2\right)$. The agent aims at minimizing the quadratic terminal cost
	$$ 
	J(\alpha) = \mathbb{E}\left[\left(X^\alpha_T-c\right)^2 / 2\right]
	$$
	with target value $c=1.5$. In this case, the Hamiltonian is given by
	$$
	H(t, x, a, p, q, r)=(\rho x+(\mu-\rho) a) p+\sigma a q+\lambda \mathbb{E}^{\nu} \left[a Z r(Z)  \right] .
	$$
	Under this minimization framework, the corresponding adjoint BSDE is
	\begin{equation}
		\begin{cases} 
			dp_t  &=-\rho p_t d t+ q_t d W_t+\int_{\mathbb{R}} r_t(z) \tilde{N}(d t, d z) \, , \\
			p_T &= -(X^\alpha_T - c)\, .
		\end{cases}
	\end{equation}
	
\begin{figure}[H]	
		\centering
		\includegraphics[scale=0.5]{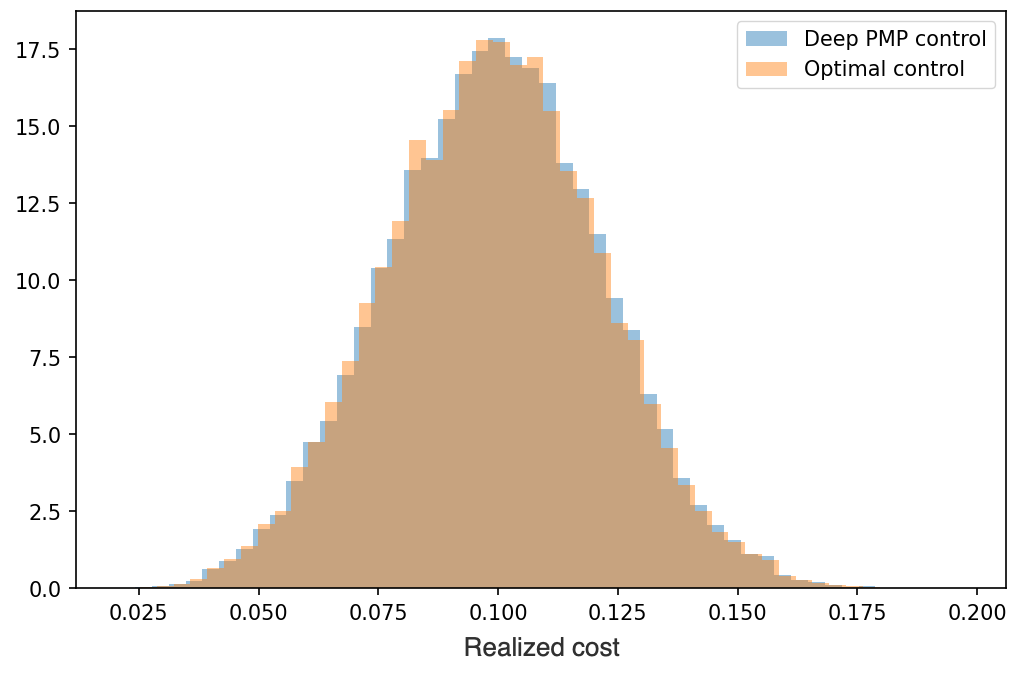}
		\label{meanvardist}
		\vspace{-2mm}
		\caption{Distribution of the terminal cost $0.5(X^\alpha_T -c)^2$ under the learned control via Algorithm \ref{alg:bptt-pmp} (blue) and under the optimal control \eqref{opti_cont} (orange).}
		\label{meanvar}
	\end{figure}

\begin{figure}[H]
		\hspace{-3mm}	\begin{subfigure}{.5\textwidth}
			\includegraphics[width=0.99\linewidth, height=5.21cm]{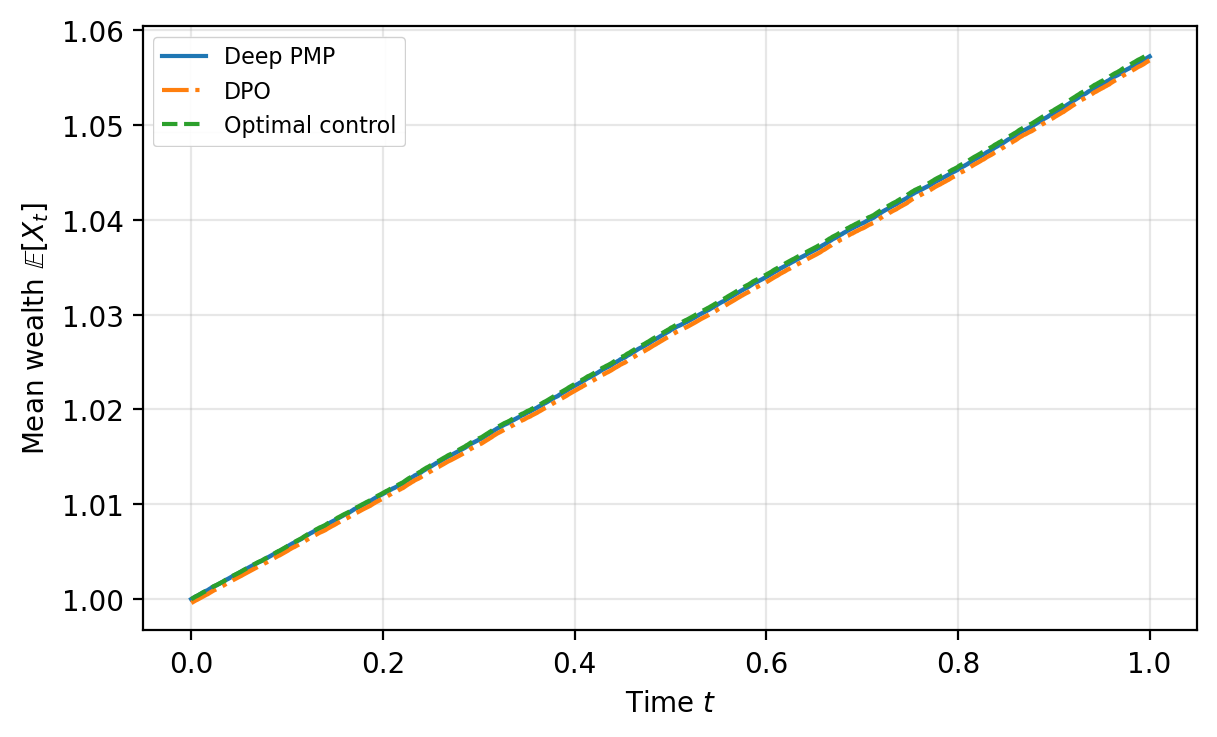}
			\label{ratejentzen}
		\end{subfigure}%
		\hspace{-1.2mm}	\begin{subfigure}{.5\textwidth}
			\includegraphics[width=0.99\linewidth, height=5.2cm]{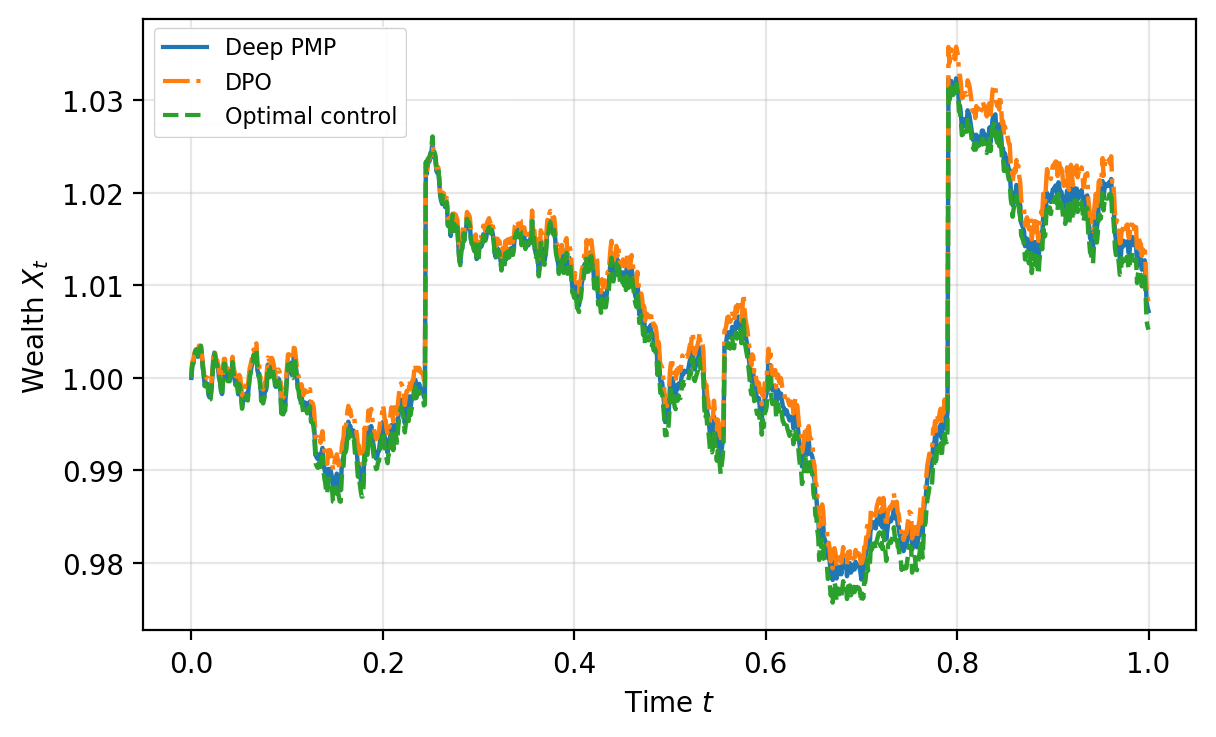}
			\label{invjentzen}
		\end{subfigure}
		\vspace{-3mm}
		\caption{Comparison of the mean state trajectory $(\mathbb{E}[X^\alpha_t])_{0\leq t\leq T}$ (left) and of a  given state trajectory $(X^\alpha_t)_{0\leq t \leq T}$ (right) under the learned control via Algorithm \ref{alg:bptt-pmp} (blue), via the DPO method (orange) and under the optimal control \eqref{opti_cont} (dashed green).}
		\label{meanvar2}
	\end{figure}

	By Theorem 4.1 in \cite{framstad2004sufficient}, the pair ($\alpha^*, X^*$) satisfying
	\begin{equation}
		\alpha_t^*=  \alpha^*(t, X^*_t) = \frac{\rho-\mu}{\Lambda}\left(X_t^* - ce^{-\rho(T-t)}\right) \label{opti_cont}
	\end{equation}
	is optimal for the mean-variance portfolio selection problem, where
	$	\Lambda  =\sigma^2+\lambda \mathbb{E}^{\nu} [Z^2] $.
	We implement the deep PMP algorithm with  $N=25$ time steps and train for $k^*=5000$ epochs. Figure \ref{meanvar} compares the distribution of the terminal cost $0.5(X^\alpha_T-c)^2$ under the learned PMP control $\alpha_{\theta^{k*}}$ and the true optimal control $\alpha^*$, for the fixed initial condition $X_0=1$. Figure \ref{meanvar2} provides the corresponding comparison for the mean state trajectory $(\mathbb{E}[X^\alpha_t])_{0\leq t\leq T}$ and for one representative sample trajectory $(X^\alpha_t(\omega))_{0\leq t\leq T}$, also including the DPO method of \citet{han2016deep}. These results show that the proposed Deep PMP algorithm more accurately recovers the optimal control and the associated state dynamics.
	
	\vspace{-2mm}
\begin{figure}[H]
		\hspace{-3mm}	\begin{subfigure}{.5\textwidth}
			\includegraphics[width=0.99\linewidth]{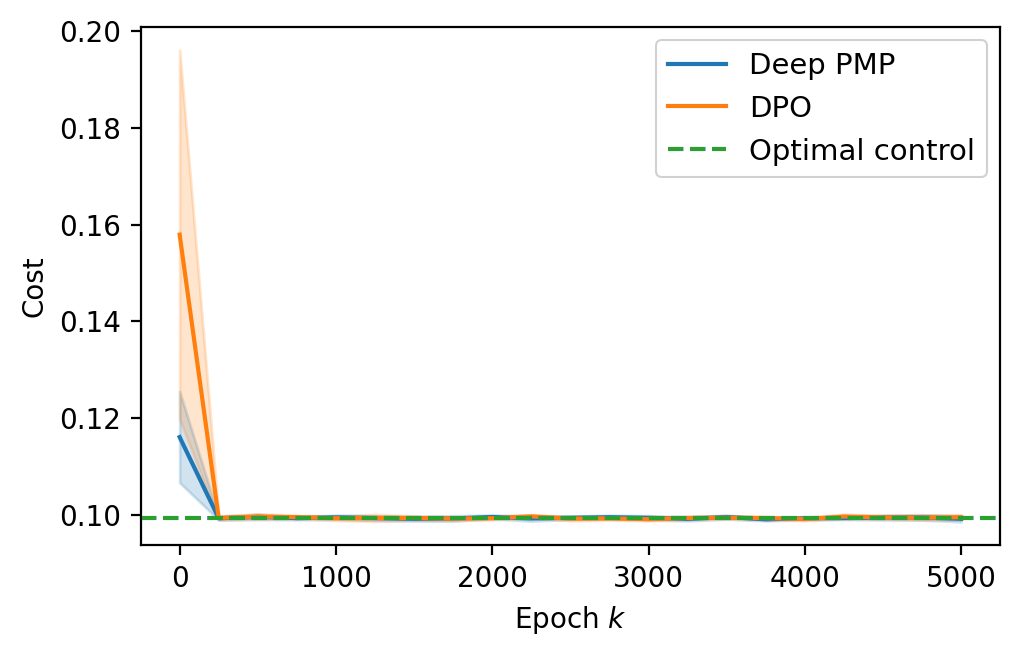}
			\label{ratejentzen}
		\end{subfigure}%
		\hspace{-1.2mm}	\begin{subfigure}{.5\textwidth}
			\includegraphics[width=0.99\linewidth]{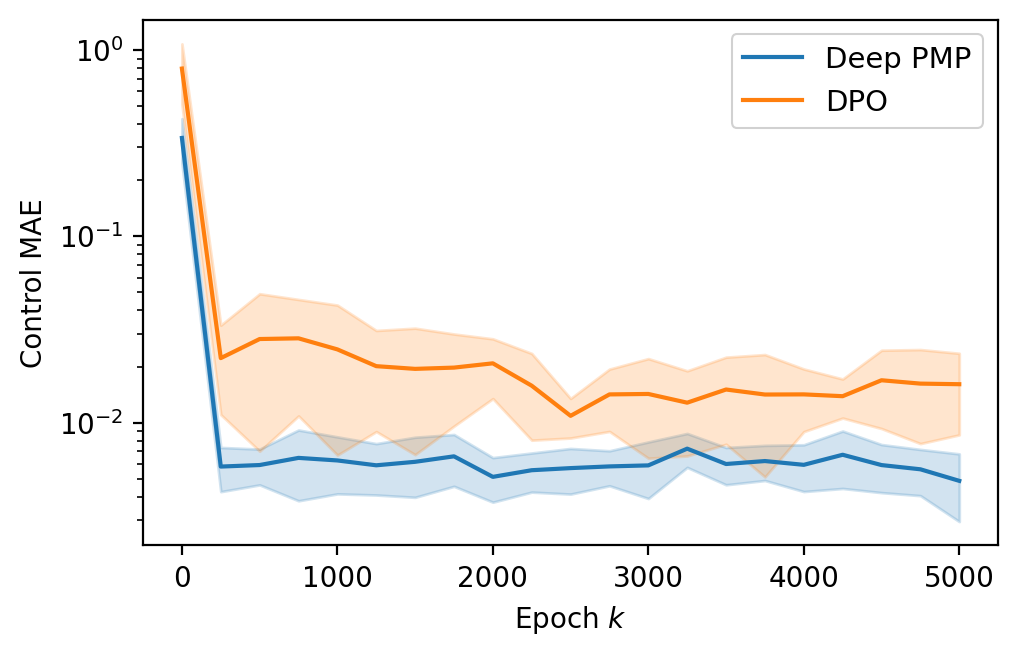}
			\label{invjentzen}
		\end{subfigure}
		\vspace{-3mm}
		\caption{Left: Cost functional $J(\alpha)$  as a function of the training epoch $k$, evaluated using the learned control from Algorithm~\ref{alg:bptt-pmp} (blue), the DPO method (orange), and the closed-form optimal control \eqref{opti_cont} (dashed line). Right: $\rm{MAE}_\alpha$ between the learned Deep PMP control (blue) and the learned DPO control (orange). Solid curves represent averages over 10 independent runs, and shaded regions indicate $\pm$ one standard deviation.}
		\label{figmaenvar}
	\end{figure}

	\vspace{-2mm}
	Every 50 epochs, we also evaluate the cost functional \(J(\alpha)\) in Figure~\ref{figmaenvar} (left) using the learned control from Algorithm~\ref{alg:bptt-pmp}, the one from DPO, and the closed-form optimal control \eqref{opti_cont}. At evaluation time, the initial condition is fixed at \(X_0=1\), and the time discretization uses \(N=1000\) steps. Figure~\ref{figmaenvar} (right)  then reports the control mean absolute error (MAE), computed along simulated validation trajectories initialized with
	\(X_0^{(m)} \sim \mathcal{U}([0.5,2.0])\), as
	\begin{equation} \label{mae_control}
		\mathrm{MAE}_{\alpha}
		=
		\frac{1}{M_1 N} 
		\sum_{m=1}^{M_1}
		\sum_{i=0}^{N -1}
		\left|
		\alpha_{\theta^k}(t_i,X_i^{(m)}) - \alpha^\ast(t_i,X_i^{(m)})
		\right|,
	\end{equation}
	where \(\alpha_{\theta^k}\) denotes the learned  control at epoch $k$ and \(\alpha^\ast\) the closed-form optimal control. The plotted curve shows the mean control MAE over 10 independent runs of each algorithm, while the shaded region represents $\pm$ one standard deviation.
	These figures again confirm the effectiveness of the Deep PMP method, which
	outperforms DPO even in this one-dimensional benchmark.

	\subsection{Optimal Consumption-Investment with Jumps}
	\label{optcv}
\begin{figure}[H]
		\hspace{-3mm}	\begin{subfigure}{.5\textwidth}
			\includegraphics[width=0.99\linewidth, height=4.6cm]{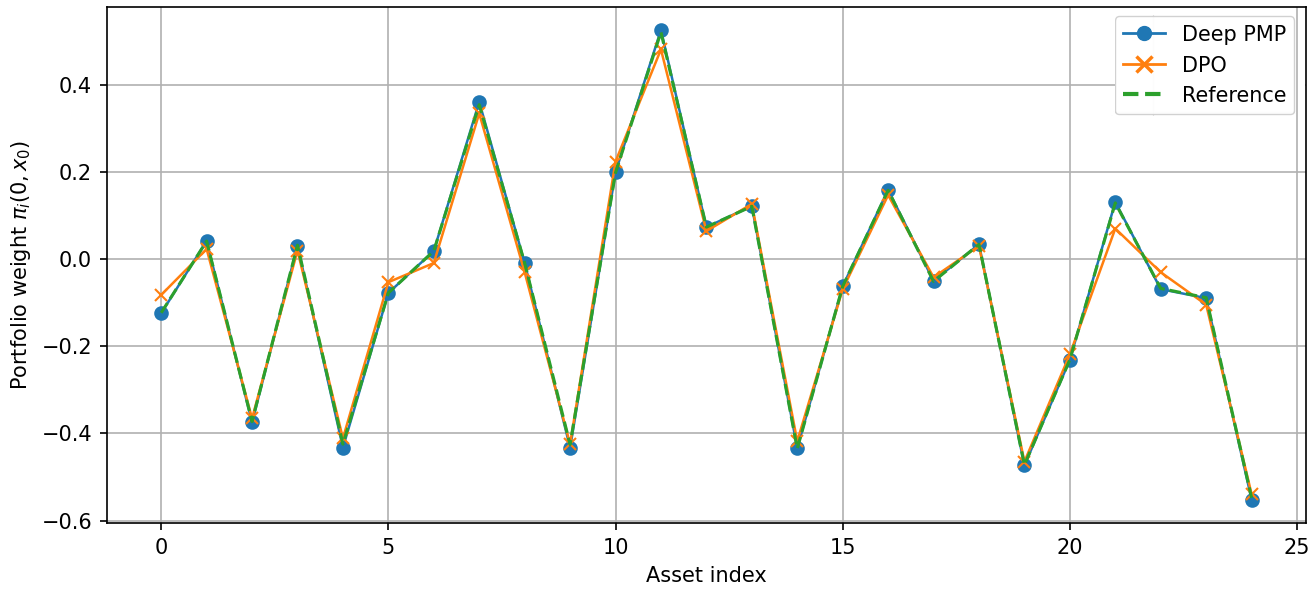}
		\end{subfigure}%
		\hspace{-1.2mm}	\begin{subfigure}{.5\textwidth}
			\includegraphics[width=0.99\linewidth, height=4.6cm]{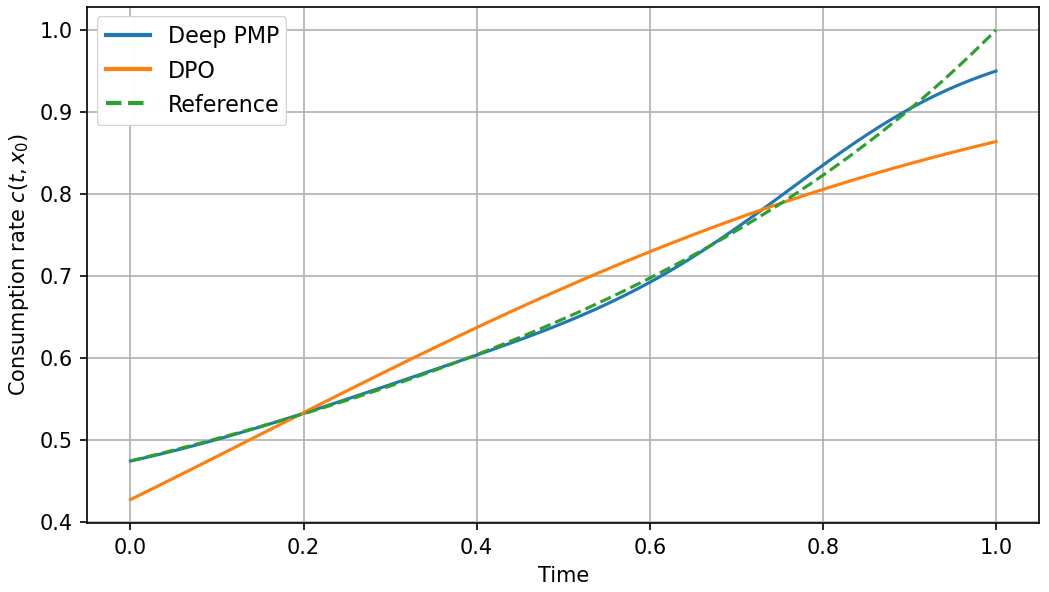}
		\end{subfigure}
		\vspace{-3mm}
		\caption{Left: Portfolio weights $\pi_i(t_0,x_0)$ for $t_0 =0, x_0 =1$ and the first 25 assets $i=1,\ldots, 25$. Right:  consumption rate $c(t,x_0)$ at $x_0 =1$ in function of time $t\in[0,1]$. Comparison of the deep PMP (blue), the DPO (orange) and the reference solution \eqref{pi-star}--\eqref{ustar} (dashed green).}
		\label{fig:optinv_weights}
	\end{figure}

\begin{figure}[H]
		\hspace{-3mm}	\begin{subfigure}{.5\textwidth}
			\includegraphics[width=0.99\linewidth]{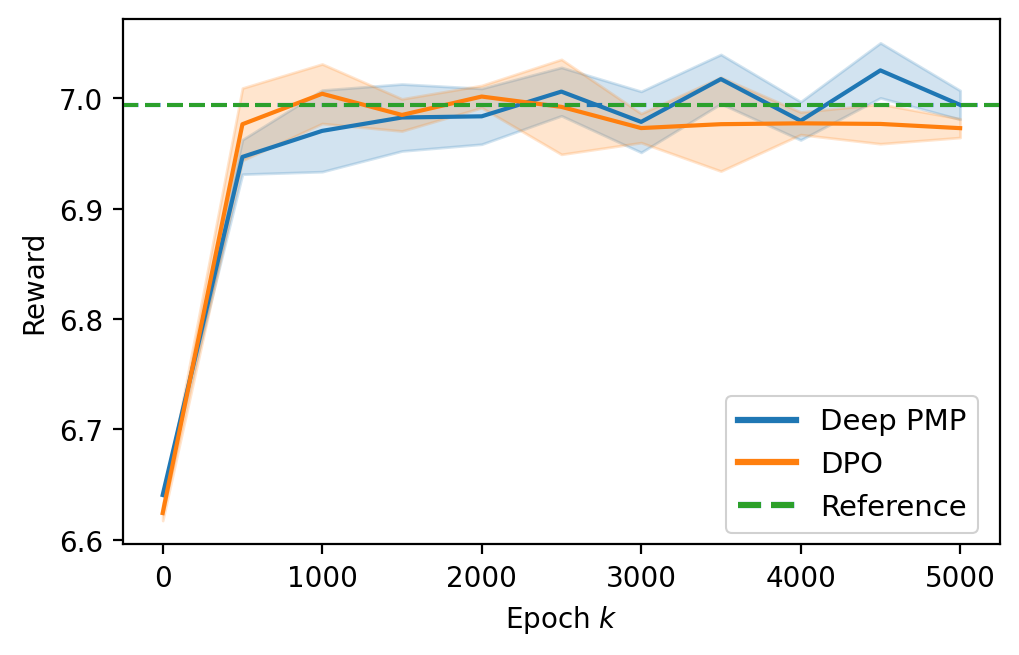}
		\end{subfigure}%
		\hspace{-1.2mm}	\begin{subfigure}{.5\textwidth}
			\includegraphics[width=0.99\linewidth]{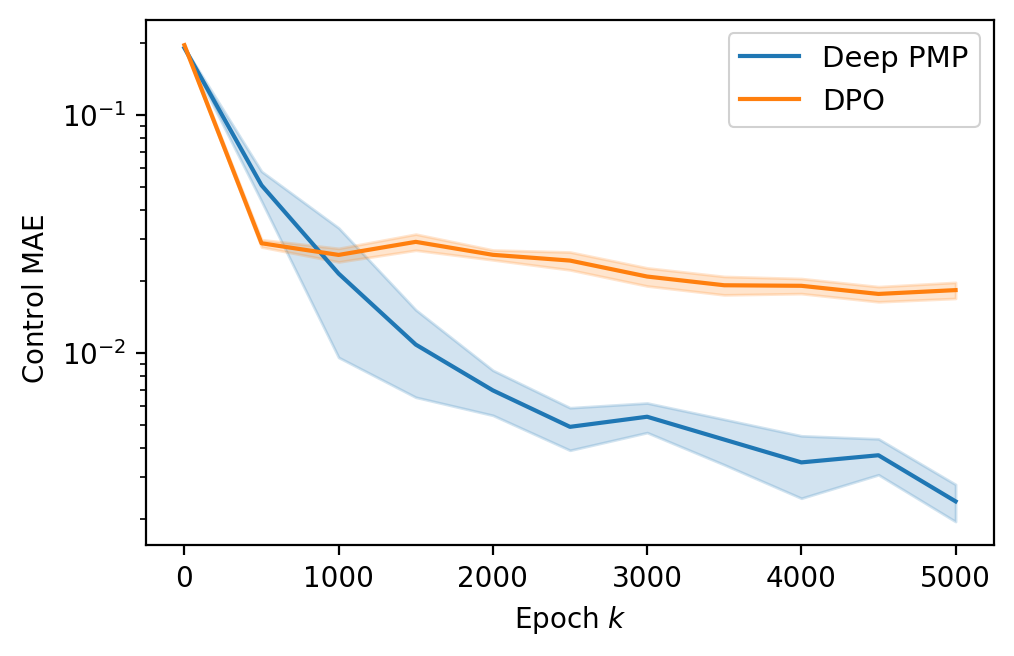}
		\end{subfigure}
		\vspace{-3mm}
		\caption{Left: Reward functional $J(\alpha)$ \eqref{optinvesst}  as a function of the training epoch $k$, evaluated using the learned control from Algorithm~\ref{alg:bptt-pmp} (blue), the DPO method (orange), and the optimal control \eqref{pi-star}--\eqref{ustar} (green dashed line). Right: $\rm{MAE}_\alpha$ between the learned Deep PMP control (blue) and the learned DPO control (orange). Solid curves represent averages over 10 independent runs, and shaded regions indicate $\pm$ one standard deviation}
		\label{fig:optinvcomp}
	\end{figure}

	We now study a multi-dimensional extension of the previous mean-variance portfolio example. We first consider the optimal consumption-investment problem in a market
	with constant coefficients, where the risk-free asset evolves according to
	\begin{equation*}
		d S^0_t = r_f S^0_t dt  \, , \label{r}
	\end{equation*}
	for a constant interest rate $r_f \in \R$, and where each stock $i=1,\ldots,n$ follows
	\begin{equation}
		\frac{dS^i_t}{S^i_{t-}} = \mu^i dt + \sigma^i \sum_{k=1}^n L_{ik}\,dB_t^k
		+ d \sum_{j=1}^{N^i_t} \brak{e^{Z^i_j} - 1} , \label{S}
	\end{equation}
	for constant drifts $\mu^i \in \R$ and volatilities $\sigma^i \in \R_+$,
	an $n$-dimensional standard Brownian motion $B$, a correlation matrix $\Sigma=LL^\top$, independent Poisson processes $N^i$ with constant
	intensities $\lambda^i \ge 0$ and i.i.d.\ normal random variables $Z^i_1, Z^i_2, ... \sim \nu^i = \mathcal{N}(\mu^i_Z, \sigma^i_Z)$ with $\mu^i_Z \in \mathbb{R}$ and $\sigma^i_Z \in \mathbb{R}_+$. The corresponding compensated Poisson measure is then written \(
	\tilde N^i(dt,dz):=N^i(dt,dz)-\lambda^i \nu^i(dz)\,dt
	\).  Suppose an investor starts 
	with an initial endowment of $X_0 > 0$, consumes at rate $c_t X^\alpha_{t-}$ and invests 
	$\pi^i_t X^\alpha_{t-}$ in each of the $n$ stocks. If the risk-free asset is used to 
	balance the transactions and writing $\mu = (\mu^1, \ldots, \mu^n)^\top$\hspace{-0.5mm}, $\sigma = \text{diag}(\sigma^1,\ldots, \sigma^n)$, the resulting wealth evolves according to 
	\begin{equation} \label{dynconst}
		\frac{d X^\alpha_t}{X^\alpha_{t-}} 
		=  \Bigl(r_f+\pi_t^\top(\mu-r_f\mathbf 1)-c_t\Bigr)\,dt
		+ \pi_t^\top \sigma L\,dB_t
		+ \sum_{i=1}^n \pi_t^i\,
		d\!\left(\sum_{j=1}^{N_t^i}(e^{Z_j^i}-1)\right) .
	\end{equation}
	We consider admissible controls \(\alpha_t=(c_t,\pi_t)\), where \(c_t\ge 0\), \(\pi_t\in\mathbb{R}^n\), and
	\[
	1+\pi_t^i(e^{Z_1^i}-1)>0
	\quad\text{a.s. for all }i=1,\dots,n,
	\]
	so that wealth remains strictly positive after jumps, and we assume the investor maximizes
	\begin{equation}	
		J(\alpha) = \mathbb{E}\left[\int_{0}^{T}e^{-\rho s}\frac{\left(c_{s}X^\alpha_{s}\right)^{\delta}}{\delta}ds+e^{-\rho T} \frac{\left(X^\alpha_{T}\right)^{\delta}}{\delta}  \right] , \label{optinvesst}
	\end{equation}
	for  a discount factor 
	$\rho \in \mathbb{R}$  and a constant relative risk aversion $\gamma >0$, $\gamma\neq 1$ with $\delta := 1-\gamma$. Since the driving noise 
	in \eqref{dynconst} has stationary and independent increments,
	it is enough to consider strategies of the form 
	$c_t = c(t, X^\alpha_{t-})$ and $\pi^i_t = \pi^i(t, X^\alpha_{t-})$ for functions $c, \pi^i \colon [0,T) \times \mathbb{R}_+ \to \R_+, \R$ and $i = 1, \dots, n$. 
	It is known from \cite{oksendal2013stochastic}  that the optimal fractions of wealth is a vector of constants $\pi^*_t= \pi^* :=(\pi^{1,*}, \ldots,\pi^{n,*})^\top$ satisfying
	\begin{equation}
		0
		=
		(\mu-r_f\mathbf 1)
		+(\delta-1)\sigma \Sigma\sigma \,\pi^*
		+
		\begin{pmatrix}
			\lambda^1 \mathbb E\!\left[\bigl(1+\pi^{1,*}(e^{Z_1^1}-1)\bigr)^{\delta-1}(e^{Z_1^1}-1)\right]\\
			\vdots\\
			\lambda^n \mathbb E\!\left[\bigl(1+\pi^{n,*}(e^{Z_1^n}-1)\bigr)^{\delta-1}(e^{Z_1^n}-1)\right]
		\end{pmatrix} ,
		\label{pi-star}
	\end{equation}
	while the value function $V(t,x) = \sup_{\alpha} J(t,x,\alpha)$ and the optimal consumption rate $c^*$ are  given, respectively, by
	\begin{equation} \label{ustar}
		V(t,x) = A(t)x^\delta/\delta \quad \text{and} \quad	c^*_t  = A(t)^{1/(\delta-1)} \, ,
	\end{equation}
	where $A : \R^+ \to \R^+$ satisfies the  ODE
	\begin{equation} \label{Aeq}
		\begin{aligned}
			\frac{	A'(t)}{A(t)} = \rho \hspace{0.5mm}- & \delta \left(r_f+(\mu-r_f \textbf{1})^{\top} \pi^* \right) - \frac{1}{2} \delta (\delta-1) \pi^{*\top}\sigma \Sigma \sigma^\top \pi^* 
			\\[-1mm]
			&\hspace{21mm}- \sum_{i=1}^{n} \lambda^i \mathbb{E} \left[ \left(1+\pi^{*,i} (e^{Z^i_1} - 1)\right)^{\delta} \hspace{-0.5mm}-1 \right] - (1-\delta)A(t)^{1/(\delta-1)} ,
		\end{aligned}
	\end{equation}
	with terminal condition $A(T) = 1$, which can be solved numerically using an ODE solver, e.g.\ Runge--Kutta method of order 5(4). 
	Finally, we derive the corresponding PMP. The Hamiltonian is
	\begin{equation}
		\begin{aligned}
			H(t,x,a,p,q,r)
			&=
			e^{-\rho t}\frac{(cx)^\delta}{\delta}
			+ x\Big(r_f+\pi^\top(\mu-r_f\mathbf 1)-c\Big)p
			+ x\,\pi^\top \sigma L\,q\\
			&\quad
			+ x\sum_{i=1}^n \lambda^i \pi^i \kappa^i \,p
			+ x\sum_{i=1}^n \lambda^i \pi^i \int_\R (e^{z}-1)\,r^i(z)\,\nu^i(dz),
		\end{aligned}
		\label{Ham-const}
	\end{equation}
	where \(p\in\mathbb{R}\), \(q=(q_1,\ldots,q_n) \in\mathbb{R}^n\), and \(r=(r^1,\dots,r^n)\) with \(r^i:\mathbb{R}\to\mathbb{R}\). The corresponding adjoint equation is the BSDE
	\begin{equation}
		\begin{aligned}
			dp_t
			&=
			-\nabla_x H\bigl(t,X_{t-}^{\alpha^*},\alpha_t^*,p_t,q_t,r_t\bigr)\,dt
			+ q_t^\top dB_t
			+ \sum_{i=1}^n \int_{\mathbb{R}} r_t^i(z)\,\tilde N^i(dt,dz),
		\end{aligned}
		\label{adjoint-const}
	\end{equation}
	with terminal condition $
	p_T=e^{-\rho T}(X_T^{\alpha^*})^{\delta-1}.$ 
	Finally, the stationary condition is given by
	\[
	\nabla_a H(t,x,a,p,q,r)
	=
	\begin{pmatrix}
		\partial_c H(t,x,a,p,q,r)\\[1mm]
		\nabla_\pi H(t,x,a,p,q,r)
	\end{pmatrix} =0,
	\]
	where
	\[
	\partial_c H(t,x,a,p,q,r)
	=
	e^{-\rho t}x^\delta c^{\delta-1}-xp,
	\]
	and $
	\nabla_\pi H(\cdot) = (\partial_{\pi^1} H(\cdot),\ldots, \partial_{\pi^n}H(\cdot) )^\top \, , $ with for $i=1,\ldots,n$,
	\[ \partial_{\pi^i}H(t,x,a,p,q,r)
	=
	x\left(
	(\mu^i-r_f)p
	+\sigma^i\sum_{k=1}^n L_{ik}q_k
	+\lambda^i\kappa^i\,p
	+\lambda^i\int_{\mathbb R}(e^z-1)r^i(z)\,\nu^i(dz)
	\right).
	\]
	For the numerical experiments, we set \(n=100\), \(T=1\), \(r_f=\rho=0.01\), \(\delta=0.25\) and randomly sampled $\bar\mu$, $\bar\sigma$, $\bar\lambda$, $\bar\mu_Z$, $\bar\sigma_Z, L$. 
	During training, we fix $N=50$ discretization steps and  the initial wealth used for the Deep PMP rollouts is sampled
	from
	\(
	X_0 \sim \mathcal U([0.5,1.5]).
	\)
	We then compare in Figure \ref{fig:optinv_weights} the learned feedback control $\alpha_{\theta^{k^*}}$ from the deep PMP, the DPO baseline and  the reference consumption-investment
	strategy \(\alpha^\ast=(c^\ast,\pi^\ast)\) in \eqref{pi-star}--\eqref{ustar}. Every \(500\) epochs, we also evaluate in Figure \ref{fig:optinvcomp} (left) the reward functional \(J(\alpha)\) on
	independent validation trajectories with fixed initial wealth 
	\(X_0=1\) and $N=1000$ steps.
	Figure~\ref{fig:optinvcomp} (right) then compares the
	control MAE \eqref{mae_control} for an initial wealth
	\(X_0\sim\mathcal U([0.5,1.5])\). These figures again confirm that the deep PMP approaches the reference value accurately
	and consistently outperforms DPO.
	
	\begin{figure}[H]
		\hspace{-3mm}	\begin{subfigure}{.5\textwidth}
			\includegraphics[width=0.99\linewidth, height=5cm]{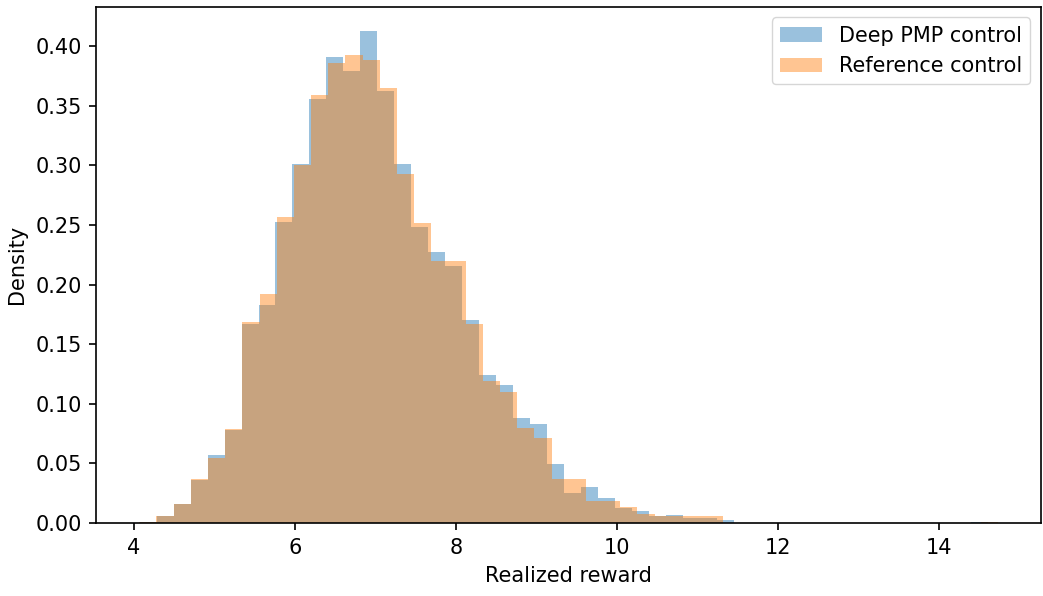}
		\end{subfigure}%
		\hspace{-1.2mm}	\begin{subfigure}{.5\textwidth}
			\includegraphics[width=0.99\linewidth, height=5.01cm]{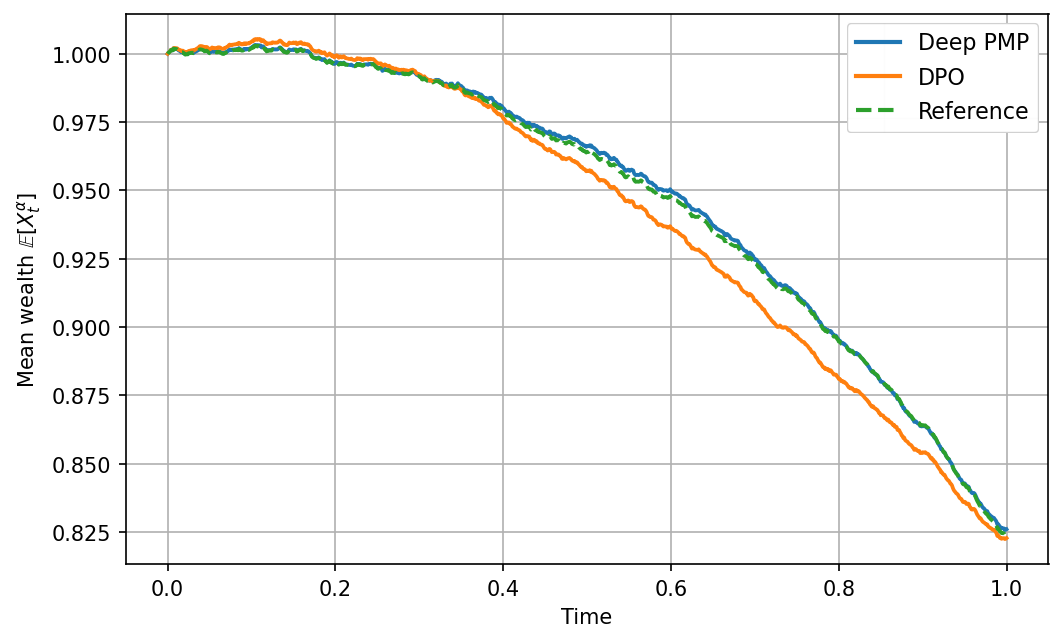}
		\end{subfigure}
		\vspace{-3mm}
		\caption{Left: Distribution of the reward in \eqref{optinvesst} under the learned deep PMP control  (blue) and under the reference control \eqref{pi-star}--\eqref{ustar} (orange). Right: Comparison of the mean state trajectory $(\mathbb{E}[X^\alpha_t])_{0\leq t\leq T}$ under the learned control via the deep PMP  (blue), the DPO method (orange) and under the reference control (dashed green). }
		\label{fig:optinv_reward}
	\end{figure}
	
	\subsection{LQR with marked Poisson controlled jumps}
	
	Consider the controlled jump-diffusion system
	\begin{equation}
		dX^\alpha_t
		=
		(AX^\alpha_t + B\alpha_t)\,dt
		+
		F\,dW_t
		+
		\int_E (CX^\alpha_{t^-} + D\alpha_t)zN(dt,dz),
	\end{equation}
	where \(X^\alpha_t \in   \mathbb{R}^n\), \(\alpha_t \in \mathbb{R}^m\), \(W_t\) is a $d$-dimensional standard Brownian motion, and \(N(dt,dz)\) is a Poisson random measure on \(E=\R^n\) with compensator $
	\lambda \nu(dz)\,dt$, $\lambda > 0$.
	We assume that the first two mark moments are finite and define
	\begin{equation}
		\mu_1
		:=
		\lambda  \int_E z\,\nu(dz),
		\qquad
		\mu_2
		:=
		\lambda \int_E z^2\,\nu(dz).
	\end{equation}
	The matrices \(A,B,C,D,F\) are assumed to have compatible dimensions. The goal is to minimize the finite-horizon quadratic cost 
	\begin{equation}
		J(\alpha)
		=
		\mathbb{E}
		\left[
		X_T^{\alpha \top} G X^\alpha_T
		+
		\int_0^T
		\left(
		X_t^{\alpha \top} QX^\alpha_t
		+
		\alpha_t^\top R\alpha_t
		\right)dt
		\right], \label{eq:LQR_cost}
	\end{equation}
	where \(Q\succeq 0\), \(G\succeq 0\), and \(R\succ 0\). Because of the additive Brownian noise, the value function is written as
	\begin{equation*}
		V(t,x)
		=
		x^\top P(t)x + \eta(t),
	\end{equation*}
	where \(P(t)\) is symmetric and \(\eta(t)\) is a scalar function. The optimal control is still a linear feedback law
	\begin{equation} \label{eq:Ricatti}
		\alpha_t^*
		=
		-K(t)X^\alpha_{t^-},
	\end{equation}
	where
	\begin{equation*}
		K(t)
		=
		\left(
		R+\mu_2 D^\top P(t)D
		\right)^{-1}
		\left(
		B^\top P(t)
		+
		\mu_1 D^\top P(t)
		+
		\mu_2 D^\top P(t)C
		\right).
	\end{equation*}
	
\begin{figure}[H]	
		\centering
		\includegraphics[scale=0.5]{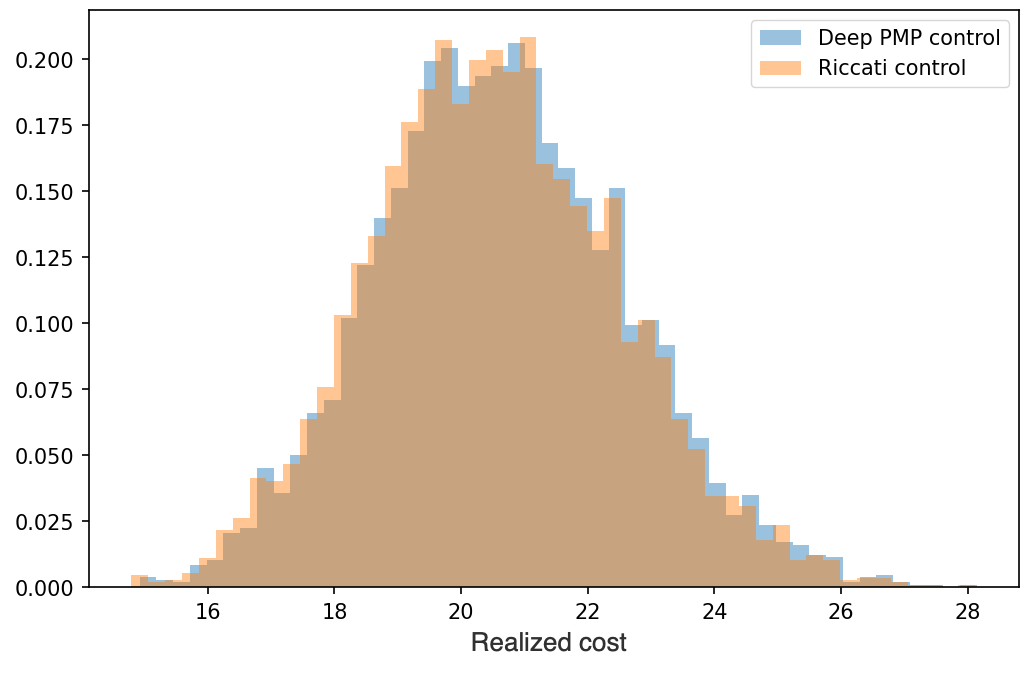}
		\label{lqrhist}
		\vspace{-2mm}
		\caption{Distribution of the quadratic cost in \eqref{eq:LQR_cost} under the learned control via the deep PMP algorithm \ref{alg:bptt-pmp} (blue) and under the Riccati control \eqref{eq:Ricatti} (orange).}
		\label{fig:hist_lqr} \vspace{-2mm}
	\end{figure}

\begin{figure}[H]
		\hspace{-3mm}	\begin{subfigure}{.5\textwidth}
			\includegraphics[width=0.999\linewidth, height=5.32cm]{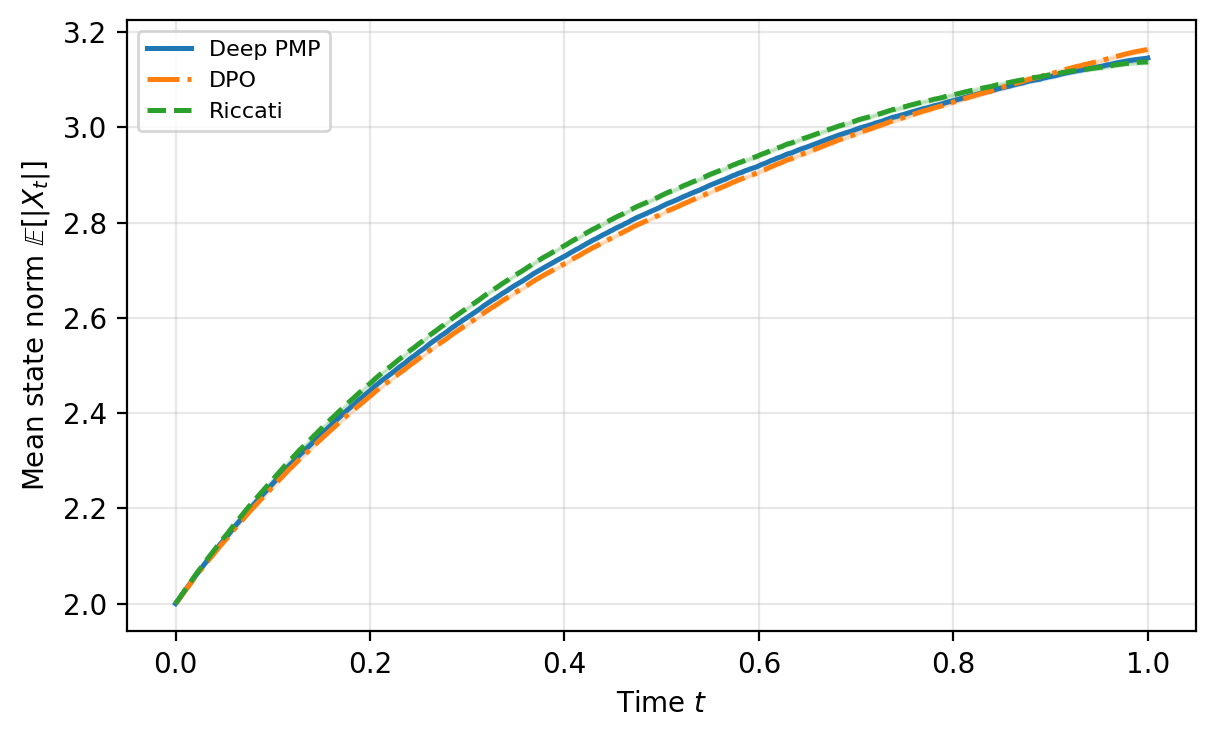}
		\end{subfigure}%
		\hspace{-1.2mm}	\begin{subfigure}{.5\textwidth}
			\includegraphics[width=0.99\linewidth, height=5.3cm]{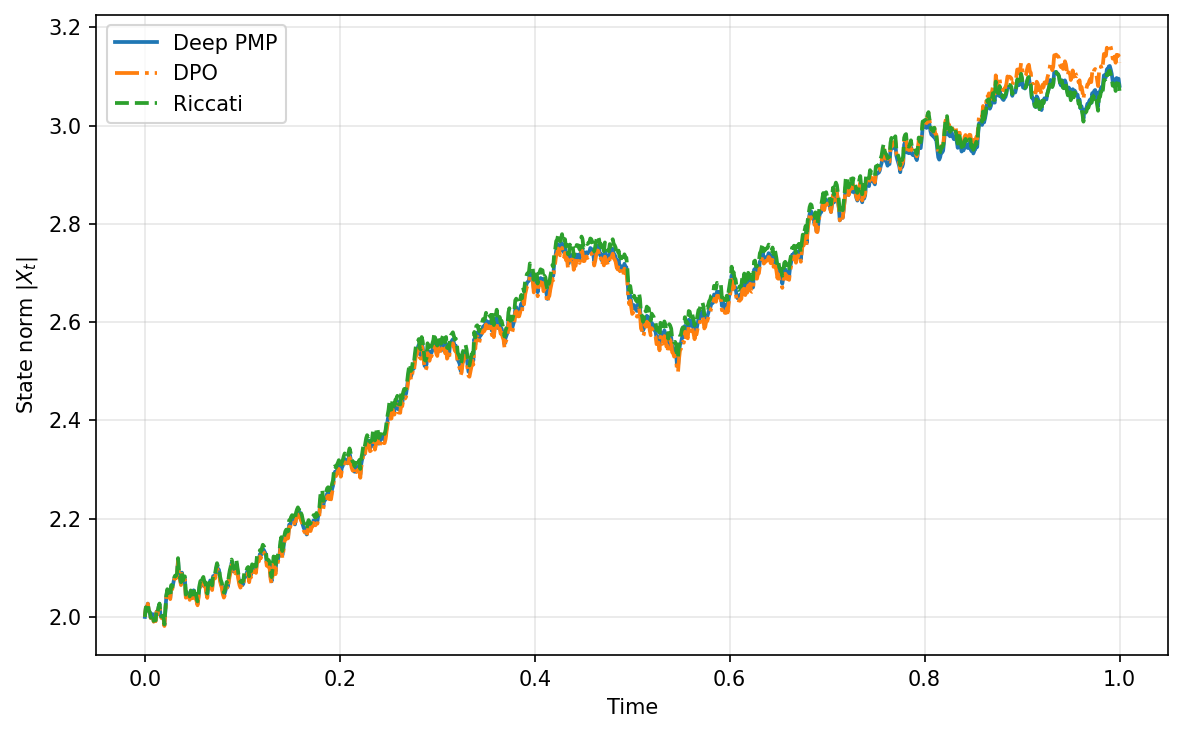}
		\end{subfigure}
		\vspace{-3mm}
		\caption{Comparison of the mean state norm trajectory $(\mathbb{E}[|X^\alpha_t|])_{0\leq t\leq T}$ (left) and of a  given state norm trajectory $(|X^\alpha_t|)_{0\leq t \leq T}$ (right) under the learned control via Algorithm \ref{alg:bptt-pmp} (blue), the DPO method (orange) and under the Riccati control \eqref{eq:Ricatti} (right).}
		\label{fig:paths_lqr}
	\end{figure}

\begin{figure}[H]
		\hspace{-3mm}	\begin{subfigure}{.5\textwidth}
			\includegraphics[width=0.99\linewidth]{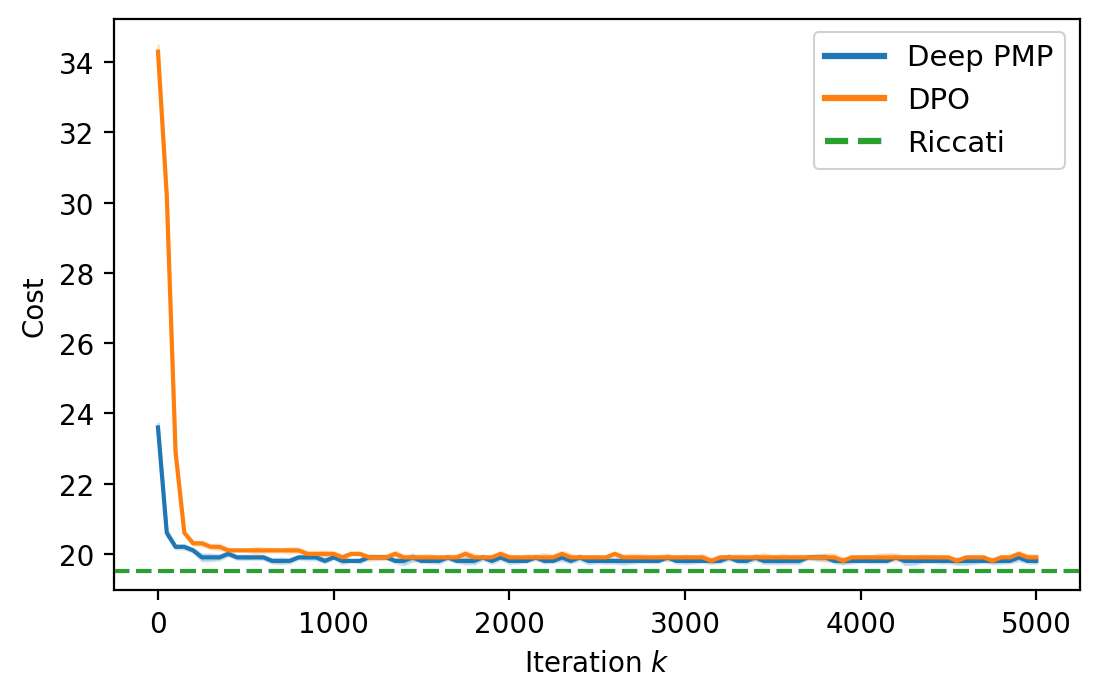}
			\label{ratejentzen}
		\end{subfigure}%
		\hspace{-1.2mm}	\begin{subfigure}{.5\textwidth}
			\includegraphics[width=0.99\linewidth]{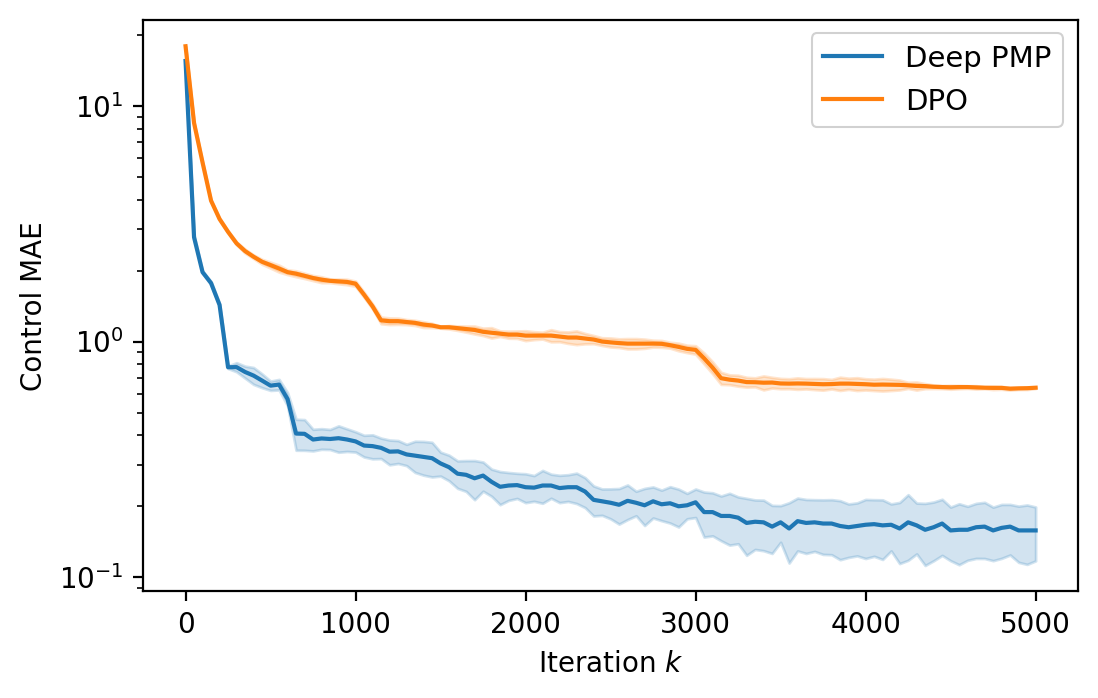}
			\label{invjentzen}
		\end{subfigure}
		\vspace{-3mm}
		\caption{Left: Cost functional $J(\alpha)$ \eqref{eq:LQR_cost}  as a function of the training epoch $k$, evaluated using the learned control from Algorithm~\ref{alg:bptt-pmp} (blue), the DPO method (orange), and the Riccati control \eqref{eq:Ricatti} (dashed line). Right: $\rm{MAE}_\alpha$ between the learned Deep PMP control (blue) and the learned DPO control (orange). Solid curves represent averages over 10 independent runs, and shaded regions indicate $\pm$ one standard deviation}
		\label{figLQRcomp}
	\end{figure}

	The Riccati differential equation for \(P(t)\) is
	{\small \begin{align*}
			-&\dot P(t)
			=
			A^\top P(t) + P(t)A + Q
			+
			\mu_1
			\left(
			C^\top P(t) + P(t)C
			\right)
			+
			\mu_2 C^\top P(t)C
			\notag \\
			&\quad
			-
			\left(
			P(t)B
			+
			\mu_1 P(t)D
			+
			\mu_2 C^\top P(t)D
			\right)
			\left(
			R+\mu_2 D^\top P(t)D
			\right)^{-1}
			\notag 
			\times
			\left(
			B^\top P(t)
			+
			\mu_1 D^\top P(t)
			+
			\mu_2 D^\top P(t)C
			\right),
	\end{align*}}with terminal condition $
	P(T)=G$, 
	while scalar term \(\eta(t)\) satisfies
	\begin{equation*}
		-\dot \eta(t)
		=
		\operatorname{tr}
		\left(
		F^\top P(t)F
		\right),
	\end{equation*}
	with terminal condition $\eta(T) = 0$.
	Since we consider a minimization problem, the Hamiltonian is given by
	\begin{align*}
		H(t,x,a,p,q,r)
		&=
		-x^\top Qx
		-
		a^\top Ra
		+
		p^\top
		\Big(
		(A+\mu_1 C)x+(B+\mu_1 D)a
		\Big)
		\notag \\
		&\quad
		+
		\operatorname{tr}(q^\top F)
		+
		\lambda  \int_E
		r(z)^\top (Cx+Da)z
		\,\nu(dz).
	\end{align*}
	The Pontryagin minimum conditions are as follows. The adjoint process \eqref{adjoint} satisfies the BSDE
	\begin{align*}
		dp_t
		&=
		-
		\nabla_x
		H
		(t, X_{t-}^{\alpha^*},\alpha_t^*,p_t,q_t,r_t(\cdot))
		\,dt
		+
		q_t\,dW_t
		+
		\int_E r_t(z)\,\widetilde N(dt,dz),
		\\
		&= 
		-
		\left[
		-2QX_t^{\alpha^*}
		+
		(A+\mu_1 C)^\top p_t
		+
		\lambda \int_E z C^\top r_t(z)\,\nu(dz)
		\right]dt
		+
		q_t\,dW_t
		+
		\int_E r_t(z)\,\widetilde N(dt,dz).
	\end{align*}
	with terminal condition $
	p_T = -2GX_T^{\alpha^*}.$ The optimal control then satisfies
	\begin{equation}
		\nabla_a  H(t,x,a,p,q,r)
		=
		-2Ra
		+
		(B+\mu_1 D)^\top p
		+
		\lambda \int_E z D^\top r(z)\,\nu(dz) = 0 \, .
	\end{equation}
	In the LQR experiment, we set the terminal time to \(T=1\) with state dimension \(n=150\) and control dimension \(m=150\). The model parameters are chosen as 
	\(
	Q=I_n,\ G=1.25 I_n, R=0.75 I_m, A=-0.35 I_n+ 0.08 \, \xi_A,
	B=0.70 \, \xi_B,
	C=0.08 \, \xi_C, 
	D= 0.10 \, \xi_D\) and \( F =0.35 I_N
	\)
	where the entries of \(\xi_A,\xi_B,\xi_C,\xi_D\) are independent standard Gaussian random variables. We take the  jump intensity to be \(\lambda=2\) and use Gaussian marks
	\(
	Z\sim \mathcal{N}(0.5,0.75^2)
	\)
	such that  $\mu_1 =1, \mu_2 =1.625$. 
	\\
	\\
	During training, the initial condition is sampled componentwise from
	\(
	X_0\sim \mathcal{U}([-4,4]^n).
	\)
	For the final cost-distribution Figure \ref{fig:hist_lqr}, we instead fix \(X_0=e_1 = (1,0,\ldots, 0)^\top\) and compute the realized costs over Monte Carlo trajectories for both the learned control $\alpha_{\theta^{k^*}}$ from Algorithm \ref{alg:bptt-pmp} and the Riccati feedback control \eqref{eq:Ricatti}. The histogram compares the empirical distributions of these realized costs and indicates that the learned policy recovers not only the Riccati value on average, but also the pathwise cost distribution induced by the stochastic jump-diffusion dynamics. This is further supported by Figure \ref{fig:paths_lqr}, which shows that the  state-norm trajectories and their mean are nearly indistinguishable under the learned and Riccati controls  \footnote{Individual jumps are not necessarily visible after taking the norm over the 150 state dimensions.}.
	Every 50 epochs, we evaluate the cost functional \(J(\alpha)\) given by Eq.\ \eqref{eq:LQR_cost} and shown in Figure~\ref{figLQRcomp} (left), using the learned PMP control from Algorithm~\ref{alg:bptt-pmp}, the DPO method, and the Riccati control \eqref{eq:Ricatti}. In the validation plots used for the cost plot, the initial condition is fixed at \(x_0=e_1\), and the time discretization uses \(N=1000\) steps. Figure~\ref{figLQRcomp} (right) reports the control MAE given by \eqref{mae_control}, computed along simulated validation trajectories initialized componentwise with
	\(
	X_0^{(m)} \sim \mathcal{U}([-4,4]^n) .
	\)
	Again, the deep PMP algorithm gives very accurate results and outperforms the DPO method in terms of accuracy.
	
	\subsection{Allen--Cahn-type stochastic control problem with jumps}
	
	Let \(X^\alpha_t \in \mathbb{R}^n\) denote the state process and let
	\(\alpha_t \in \mathbb{R}^n\) denote the control. We consider the
	controlled jump-diffusion
	\begin{equation}
		dX^\alpha_t
		=
		\bigl(h(X^\alpha_t)+\alpha_t\bigr)\,dt
		+ \sigma\,dW_t
		+ \int_E \gamma z\,\widetilde N(dt,dz),
		\qquad X_0=x_0,
	\end{equation}
	where \(W_t\) is an \(n\)-dimensional Brownian motion and
	\(\widetilde N(dt,dz)=N(dt,dz)-\lambda\nu(dz)dt\) is a compensated
	Poisson random measure on \(E=\mathbb{R}^n\), with \(\lambda>0\). The jump marks are assumed to be
	Gaussian
	\(
	\nu = \mathcal N(0,I_n).
	\)
	The nonlinear  drift \(h:\mathbb{R}^n\to\mathbb{R}^n\) is
	defined componentwise by
	\begin{equation}
		h_i(x)=x_i-x_i^3,
		\qquad i=1,\ldots,n.
	\end{equation}
	
\begin{figure}[H]	
		\centering
		\includegraphics[scale=0.55]{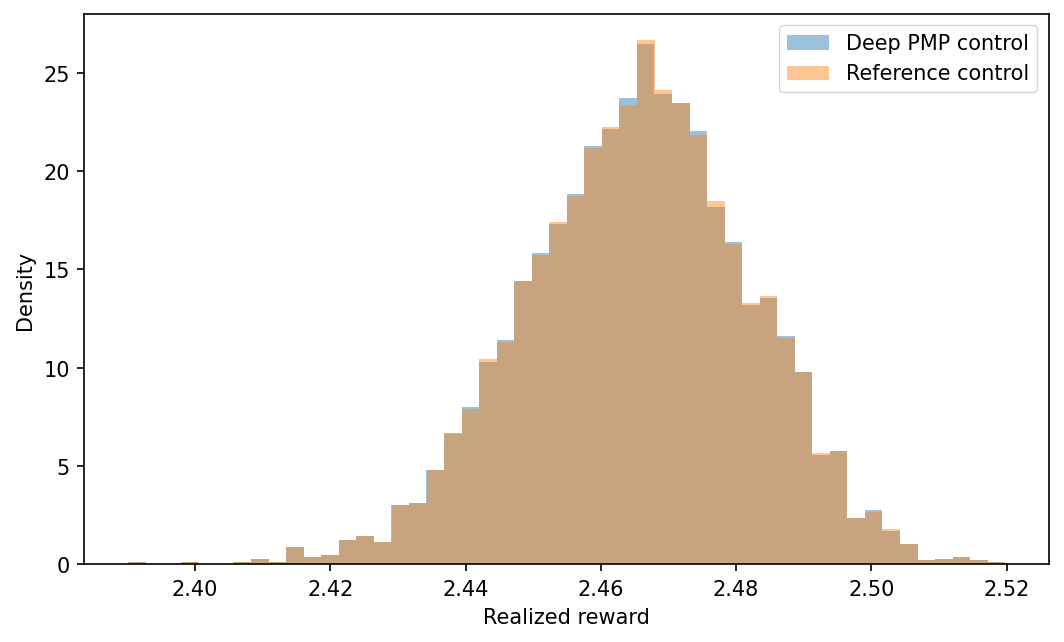}
		\vspace{-2mm}
		\caption{Distribution of the objective reward in \eqref{allen_cost} under the learned PMP control via Algorithm \ref{alg:bptt-pmp} (blue) and under the optimal control \eqref{allen:optimal} (orange).}
		\label{fig:allen_hist} \vspace{-2mm}
	\end{figure}

\begin{figure}[H]
		\hspace{-3mm}	\begin{subfigure}{.5\textwidth}
			\includegraphics[width=0.99\linewidth]{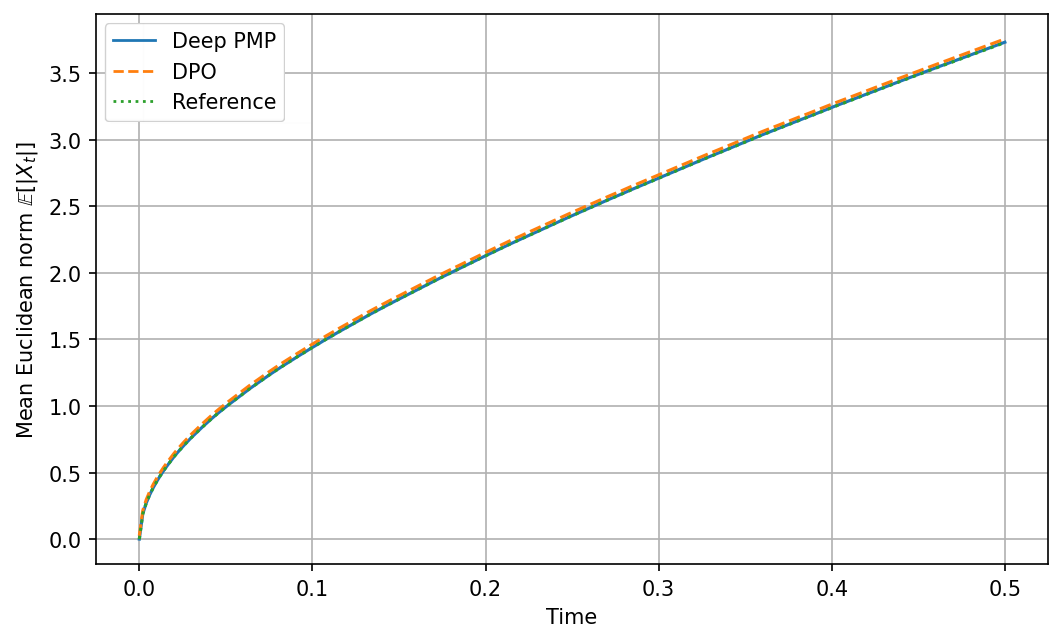}
		\end{subfigure}%
		\hspace{-1.2mm}	\begin{subfigure}{.5\textwidth}
			\includegraphics[width=0.99\linewidth]{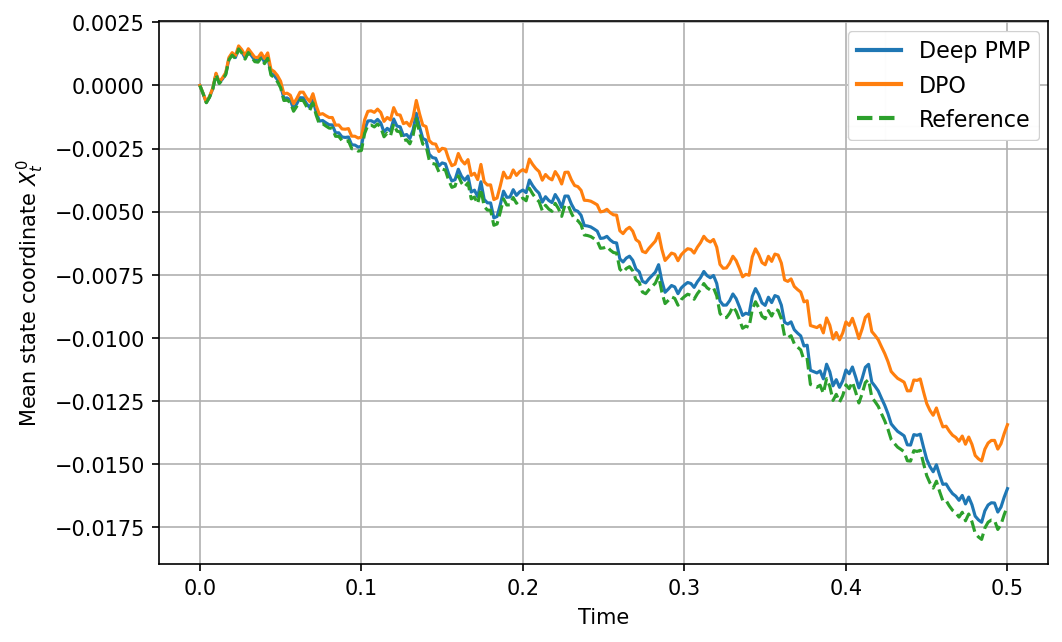}
		\end{subfigure}
		\vspace{-3mm}
		\caption{Comparison of the mean state norm trajectory $(\mathbb{E}[|X^\alpha_t|])_{0\leq t \leq T}$ (left plot) and the mean first state coordinate trajectory $(\mathbb{E}[ X^{\alpha,0}_{t}])_{0\leq t \leq T}$ under the learned control via Algorithm \ref{alg:bptt-pmp} (blue), the DPO method (orange) and under the optimal control \eqref{allen:optimal} (green), with $X_0^{(m)} = 0$.}
		\label{fig:AllenPaht}
	\end{figure}

\begin{figure}[H]
		\hspace{-3mm}	\begin{subfigure}{.5\textwidth}
			\includegraphics[width=0.99\linewidth]{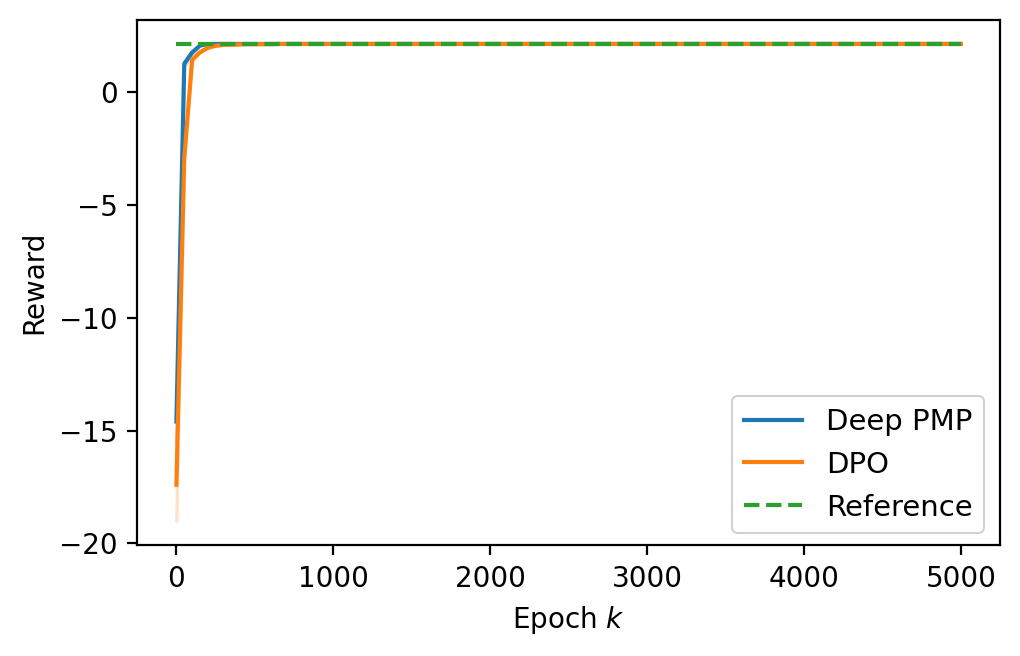}
		\end{subfigure}%
		\hspace{-1.2mm}	\begin{subfigure}{.5\textwidth}
			\includegraphics[width=0.99\linewidth]{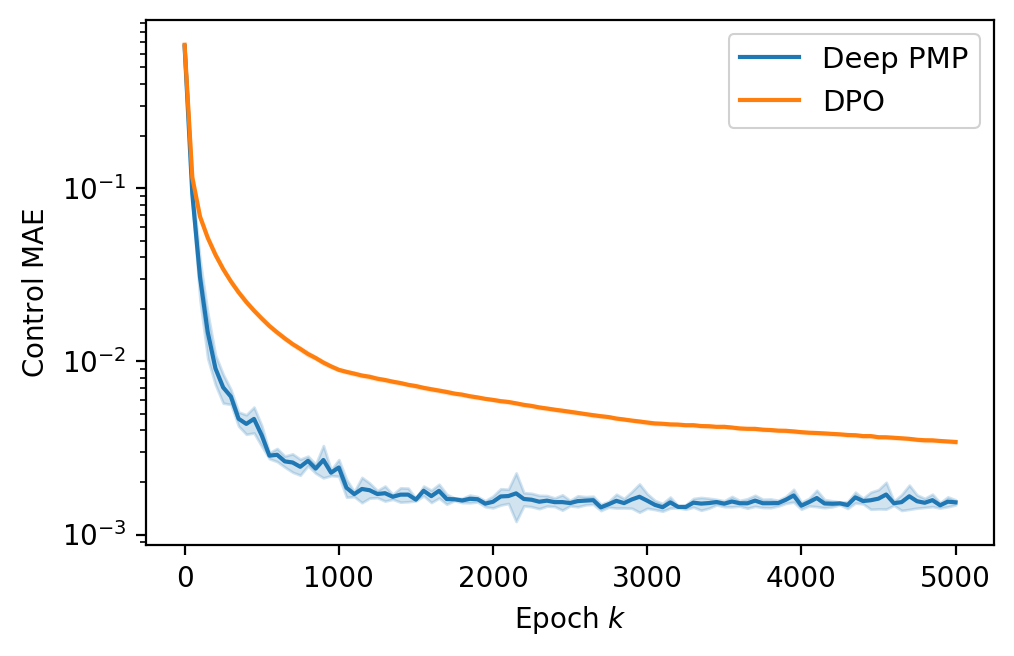}
		\end{subfigure}
		\vspace{-3mm}
		\caption{Left: Reward functional $J(\alpha)$ \eqref{allen_cost}  as a function of the training epoch $k$, evaluated using the learned control from Algorithm~\ref{alg:bptt-pmp} (blue), the DPO method (orange), and the optimal control \eqref{allen:optimal} (green dashed line). Right: $\rm{MAE}_\alpha$ between the learned Deep PMP control (blue) and the learned DPO control (orange). Solid curves represent averages over 10 independent runs, and shaded regions indicate $\pm$ one standard deviation}
		\label{fig:Allencomp}
	\end{figure}

	The objective is to maximize
	\begin{equation} \label{allen_cost}
		J(\alpha) = 
		\mathbb{E}
		\left[
		G(X^\alpha_T)
		+
		\int_0^T
		\left(
		R(t,X^\alpha_t)
		-
		\frac{\eta}{2}\|\alpha_t\|^2
		\right)dt
		\right],
	\end{equation}
	where \(\eta>0\) is the control-cost parameter.  We use here a manufactured solution to design a test problem whose exact solution is known by construction.  We first prescribe the exact reference value function 
	\begin{equation}
		V^\star(t,x)
		=
		e^{-\beta(T-t)}
		\left(
		c_0
		+
		\frac{1}{n}\sum_{i=1}^n \cos(x_i)
		\right),
	\end{equation}
	where \(\beta>0\) and \(c_0>0\) are fixed constants and then build the rewards $R,G$ that match this solution. The  terminal reward is then directly given by $G(x) = V^*(T,x)$. For the running reward $R$, we first derive
	the associated HJB-PIDE 
	{\small \begin{equation}
			\begin{aligned}
				0
				=
				\partial_t V(t,x)
				&+
				h(x)\cdot \nabla_x V(t,x)
				+
				\frac{\sigma^2}{2}\Delta V(t,x)
				+
				\mathcal I[V](t,x)
				+
				R(t,x) +
				\sup_{\alpha\in\mathbb{R}^n}
				\left\{
				\alpha\cdot \nabla_x V(t,x)
				-
				\frac{\eta}{2}\|\alpha\|^2
				\right\} ,
			\end{aligned}
	\end{equation}}
	with nonlocal jump operator 
	\begin{equation}
		\mathcal I[V](t,x)
		=
		\lambda
		\int_E
		\left[
		V(t,x+\gamma z)
		-
		V(t,x)
		-
		\gamma z\cdot \nabla_x V(t,x)
		\right]\nu(dz) ,
	\end{equation}
	and $\Delta V = \operatorname{Tr}[\nabla_x^2 V(t,x)]$.  
	Since \(Z\sim\mathcal N(0,I_n)\),
	we obtain
	\begin{equation}
		\mathcal I[V^\star](t,x)
		=
		\lambda
		e^{-\beta(T-t)}
		\frac{e^{-\gamma^2/2}-1}{n}
		\sum_{i=1}^n \cos(x_i).
	\end{equation}The maximization over \(\alpha\) is explicit and hence the optimal feedback control is
	\begin{align} \label{allen:optimal}
		\alpha^\star(t,x)
		=
		\frac{1}{\eta}\nabla_x V^\star(t,x)
		= 
		-
		\frac{e^{-\beta(T-t)}}{\eta n}
		\sin(x) \, .
	\end{align}
	The running reward \(R(t,x)\) is finally defined by plugging \(V^\star\) into the HJB-PIDE
	\begin{equation}
		R(t,x)
		=
		-
		\left[
		\partial_t V^\star(t,x)
		+
		h(x)\cdot \nabla_x V^\star(t,x)
		+
		\frac{\sigma^2}{2}\Delta V^\star(t,x)
		+
		\mathcal I[V^\star](t,x)
		+
		\frac{1}{2\eta}\|\nabla_x V^\star(t,x)\|^2
		\right] ,
	\end{equation}
	such that  \(V^\star\) is the exact value
	function of the control problem, and \(\alpha^\star\) is the exact optimal
	feedback control.
	The Pontryagin Hamiltonian associated with the control-dependent part of
	the problem is
	\begin{equation}
		\mathcal H(t,x,a,p,q,r)
		=
		R(t,x)
		-
		\frac{\eta}{2}\|a\|^2
		+
		p\cdot \bigl(h(x)+a\bigr),
	\end{equation}
	where \(p\in\mathbb{R}^d\) denotes the adjoint variable. The stationarity
	condition is
	\begin{equation}
		\nabla_a  H(t,x,a,p,q,r)
		=
		p-\eta a
		=
		0.
	\end{equation}
	In the numerical experiment, we set $n=100$, $T=0.5$, $\sigma=0.4$, $\gamma=0.25$, $\lambda=0.5$, $\eta=0.5$, $\beta=0.3$, and $c_0=2$. The training initial condition $X_0^{(m)}$ is sampled componentwise from $\mathcal U([-1,1]^n)$, while fixed-initial-condition evaluations use the deterministic  $X_0=0$. The time grid uses $50$ time steps for training and $250$ time steps at evaluation time. Figure~\ref{fig:allen_hist} compares the distribution of the terminal reward under the learned Deep PMP control and the exact optimal feedback, showing that the learned policy reproduces the reward distribution induced by $\alpha^\star$. Figure~\ref{fig:AllenPaht} confirms this pathwise agreement by comparing both the empirical mean of the state-norm trajectories $(\mathbb{E}[|X^\alpha_t|])_{0\leq t \leq T}$ and of the first state coordinate $(\mathbb{E}[X^{\alpha,0}_t|])_{0\leq t \leq T}$ under Deep PMP, DPO, and the analytic control. Finally, Figure~\ref{fig:Allencomp} reports the evolution of the estimated reward and the control error \eqref{mae_control} during training; the Deep PMP curve approaches the analytic benchmark and the corresponding control MAE decreases, indicating that the stationarity-based update recovers the manufactured optimal feedback more accurately than the DPO.
	
	
	\subsection{Mean-field Allen--Cahn control problem with nonlinear actuation}
	We finally consider a more involved extension  of the Allen--Cahn control problem with mean-field interactions, state-dependent
	control and a nonquadratic control cost.  Let
	\[
	\overline{x}=\frac1n\sum_{j=1}^n x_j,
	\qquad
	g_i(x)=1+\beta\cos(x_i),
	\]
for $i=1,\ldots,n$,	and define the controlled dynamics componentwise by
	\begin{equation}
		\begin{aligned}
			dX_t^{\alpha, i}
			&=
			\left[
			X_t^{\alpha, i}-(X_t^{\alpha, i})^3
			+\kappa(\bar{X}^\alpha_t-X_t^{\alpha, i})
			+g_i(X^\alpha_t)\alpha_t^i
			\right]dt
			+\sigma\,dW_t^i
			+\int_{\mathbb R}\gamma z \widetilde N_i(dt,dz),
		\end{aligned}
	\end{equation}
	where the compensated Poisson random measures $\widetilde{N}_i$ are independent, with
	intensity \(\lambda\) and Gaussian marks \(Z\sim\mathcal N(0,1)\). The objective is to maximize
	\begin{equation} \label{cost_adv_allen}
		J(\alpha)
		=
		\mathbb E\left[
		G(X^\alpha_T)
		+
		\int_0^T
		\left(
		R(t,X^\alpha_t)
		-
		\sum_{i=1}^n
		\left[
		\frac{\eta}{2}(\alpha_t^i)^2
		+
		\frac{\zeta}{4}(\alpha_t^i)^4
		\right]
		\right)dt
		\right].
	\end{equation}
	
	We prescribe the manufactured value function
	\begin{equation}
		\label{eq:advanced_allen_value}
		V^\star(t,x)
		=
		e^{-\beta(T-t)}
		\left[
		nc_0
		+
		\sum_{i=1}^n\cos(x_i)
		+
		\frac{\rho}{n}
		\left(\sum_{i=1}^n\sin(x_i)\right)^2
		\right].
	\end{equation}
	Its gradient is
	\begin{equation}
		\partial_{x_i}V^\star(t,x)
		=
		e^{-\beta(T-t)}
		\left[
		-\sin(x_i)
		+
		2\rho \frac{\cos(x_i)}{n}\sum_{j=1}^n\sin(x_j)
		\right].
	\end{equation}
	The HJB-PIDE contains the control maximization
	\begin{equation}
		\sup_{\alpha\in\mathbb R^n}
		\sum_{i=1}^n
		\left\{
		g_i(x)\alpha_i\partial_{x_i}V
		-
		\frac{\eta}{2}\alpha_i^2
		-
		\frac{\zeta}{4}\alpha_i^4
		\right\}.
	\end{equation}
	Defining
	\[
	r_i(t,x)=g_i(x)\partial_{x_i}V^\star(t,x),
	\]
	and since \(\eta,\zeta>0\), this equation has a unique real solution:
	\begin{equation}
		\label{eq:advanced_allen_control}
		\alpha_i^\star(t,x)
		=
		2\sqrt{\frac{\eta}{3\zeta}}\,
		\sinh\left[
		\frac13
		\operatorname{arsinh}
		\left(
		\frac{3r_i(t,x)}{2\eta}
		\sqrt{\frac{3\zeta}{\eta}}
		\right)
		\right].
	\end{equation}
	
	The running reward is manufactured by substituting \(V^\star\) into the
	HJB-PIDE:
	\begin{equation}
		\begin{aligned}
			R(t,x)
			=
			-\Bigg[
			&V_t^\star
			+
			\ell(x)\cdot\nabla_x V^\star
			+
			\frac{\sigma^2}{2}\Delta V^\star
			+
			\mathcal I[V^\star]
			+
			\sum_{i=1}^n
			\left(
			r_i\alpha_i^\star
			-
			\frac{\eta}{2}(\alpha_i^\star)^2
			-
			\frac{\zeta}{4}(\alpha_i^\star)^4
			\right)
			\Bigg],
		\end{aligned}
	\end{equation}
	where
	\(
	\ell_i(x)=x_i-x_i^3+\kappa(\overline{x}-x_i),
	\
	G(x)=V^\star(T,x).
	\)
	With this choice, \(V^\star\) is the exact value function and
	\(\alpha^\star\) is the exact optimal feedback. The Pontryagin stationarity condition is
	\begin{equation}
		\nabla_a H(t,x,a,p,q,r) = g(x)\odot p
		-
		\eta a
		-
		\zeta a^{\odot 3}
		=
		0.
	\end{equation}
	Thus, unlike the simpler Allen--Cahn benchmark, the optimal control is a nonlinear
	function of the adjoint variable and is coupled across coordinates
	through the manufactured value function.

	\begin{figure}[H]	
		\centering
		\includegraphics[scale=0.62]{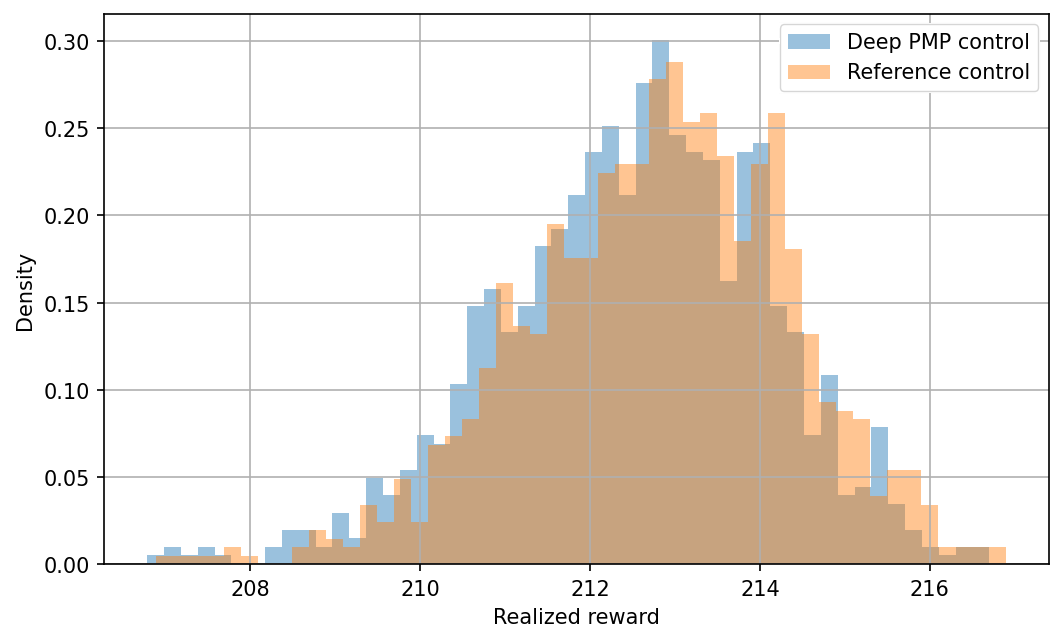}
		\vspace{-2mm}
		\caption{Distribution of the terminal reward in \eqref{cost_adv_allen} under the learned control via Algorithm \ref{alg:bptt-pmp} (blue) and under the optimal control \eqref{eq:advanced_allen_control} (orange).}
		\label{fig:allen_adv_hist} \vspace{-2mm}
	\end{figure}
	\begin{figure}[H]
		\hspace{-3mm}	\begin{subfigure}{.5\textwidth}
			\includegraphics[width=0.99\linewidth]{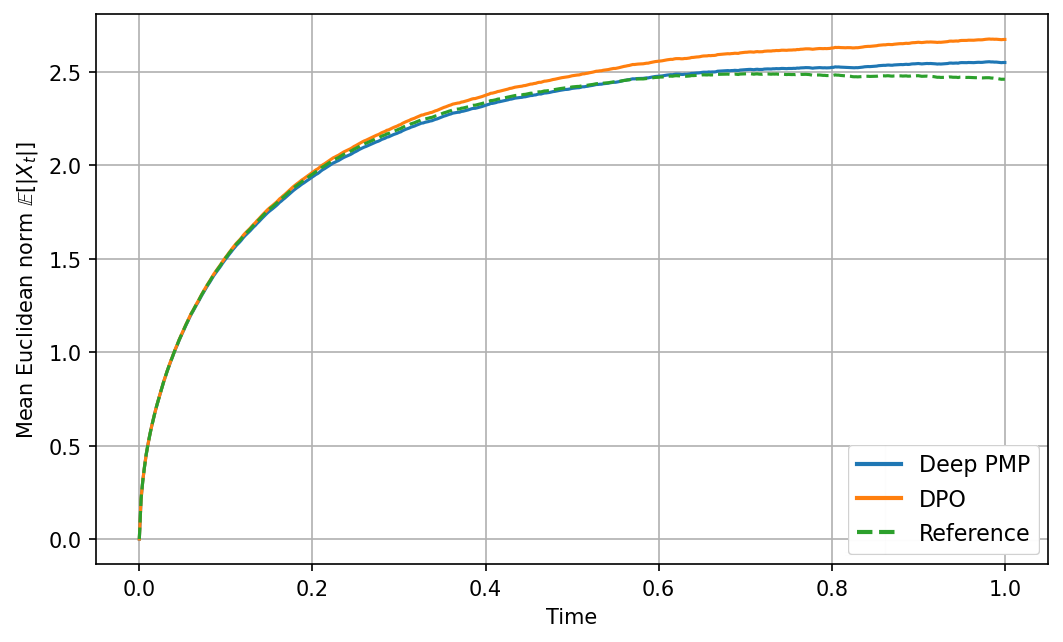}
		\end{subfigure}%
		\hspace{-1.2mm}	\begin{subfigure}{.5\textwidth}
			\includegraphics[width=0.99\linewidth]{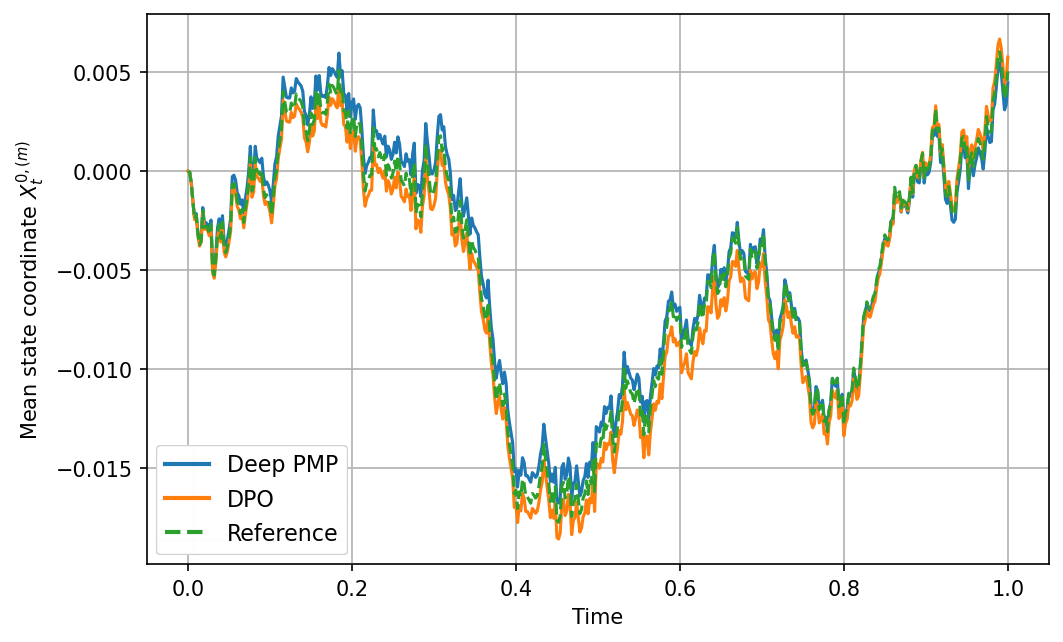}
		\end{subfigure}
		\vspace{-3mm}
		\caption{Comparison of the mean state norm trajectory $(\mathbb{E}[|X^\alpha_t|])_{0\leq t \leq T}$ (left plot) and the mean first state coordinate  trajectory $(\mathbb{E}[X^{\alpha,0}_t|])_{0\leq t \leq T}$ under the learned control via Algorithm \ref{alg:bptt-pmp} (blue), the DPO method (orange) and under the optimal control \eqref{eq:advanced_allen_control} (green), with $X_0^{(m)} = 0$.}
		\label{fig:AllenPath2}
	\end{figure}
	\begin{figure}[H]
		\hspace{-3mm}	\begin{subfigure}{.5\textwidth}
			\includegraphics[width=0.99\linewidth]{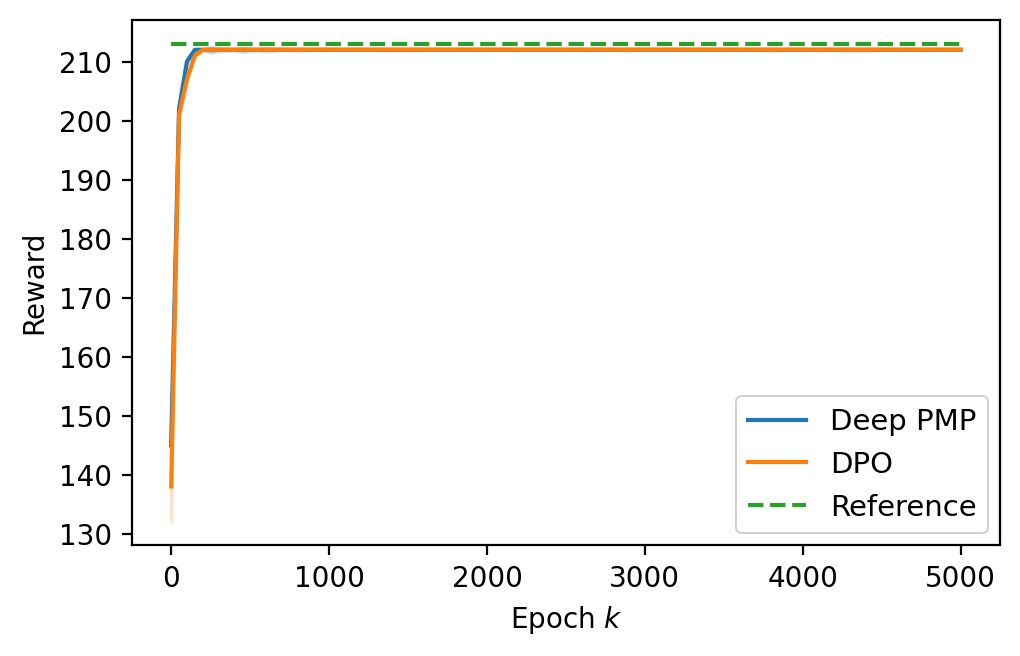}
		\end{subfigure}%
		\hspace{-1.2mm}	\begin{subfigure}{.5\textwidth}
			\includegraphics[width=0.99\linewidth]{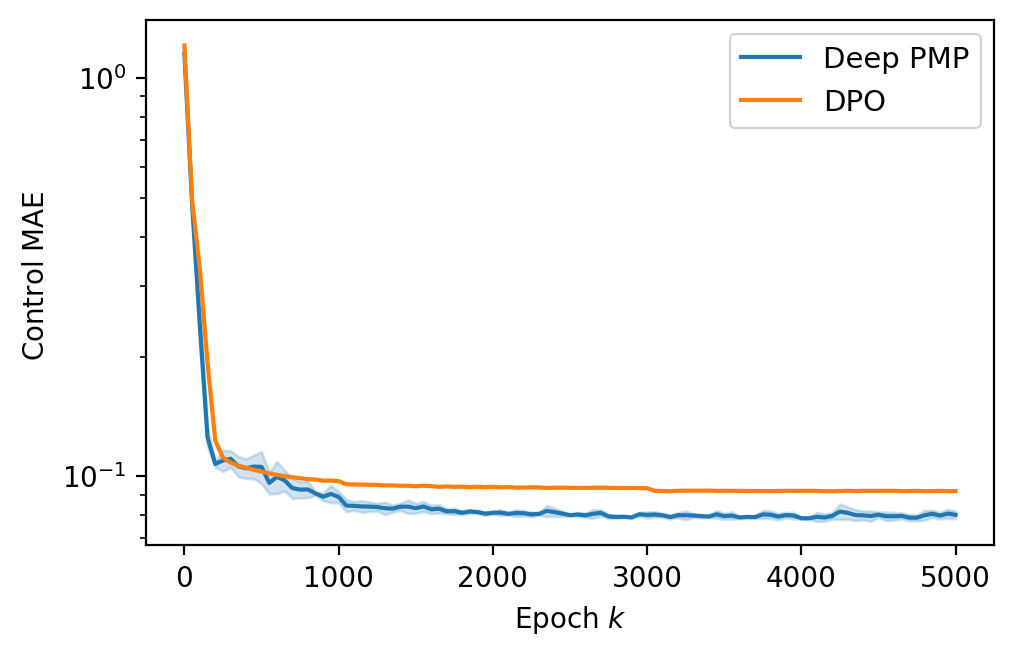}
		\end{subfigure}
		\vspace{-3mm}
		\caption{Left: Reward functional $J(\alpha)$ \eqref{cost_adv_allen}  as a function of the training epoch $k$, evaluated using the learned control from Algorithm~\ref{alg:bptt-pmp} (blue), the DPO method (orange), and the optimal control \eqref{eq:advanced_allen_control} (green dashed line). Right: $\rm{MAE}_\alpha$ between the learned Deep PMP control (blue) and the learned DPO control (orange). Solid curves represent averages over 10 independent runs, and shaded regions indicate $\pm$ one standard deviation}
		\label{fig:Allencomp_advanced}
	\end{figure}
	\section{Conclusion}
 This paper introduces a Deep Pontryagin learning framework for continuous-time stochastic control problems with jump-diffusion dynamics. The method exploits the adjoint structure of the Pontryagin stochastic maximum principle while avoiding nested Monte Carlo estimation of value-function gradients. More precisely, the deep PMP algorithm is based on a two-step policy-improvement scheme. First, for a fixed feedback control, pathwise BPTT derivatives provide noisy adjoint labels, which are then projected onto the Markov state by supervised regression. The learned adjoint field is then  used, together with the Markov representation \eqref{eq:algo1}--\eqref{eq:algo2},  to update the control through a pointwise Hamiltonian stationarity loss.
	\\[3mm]
This separation between adjoint estimation and control improvement reduces the learning problem to supervised regression and pointwise Hamiltonian maximization. It therefore avoids repeated end-to-end backpropagation through the full time-discretized objective, which is a key limitation of DPO  methods. On the theoretical side, we showed that the projected BPTT adjoint satisfies a discrete-time Pontryagin principle and converges, as the time step tends to zero, to the continuous-time adjoint BSDE.
	\\[3mm]
	The numerical experiments demonstrate that the proposed method accurately recovers reference controls and values across several jump-diffusion benchmarks, including high-dimensional and nonlinear examples. They also indicate improved stability and accuracy compared with direct policy optimization. Future work includes extensions to path-dependent problems, state- or control-dependent jump intensities, and a more detailed analysis of the neural approximation error.

	\section{Appendix}
	\subsection{Proof of Theorem \ref{th:PMP}} \label{proof:theorem}
	In order to extend the proof of Theorem~2.1 in \citet{framstad2004sufficient} to the case where the compensator of $N$ is the time-inhomogeneous measure $\lambda(t) \nu(dz) dt$, we need the following integration-by-parts formula, compare with Lemma~2.1 in \citet{framstad2004sufficient}.
	\begin{Lemma} \label{lem:integration by parts}
		Let \(b,\widetilde b\) be progressively measurable \(\mathbb R^n\)-valued processes, let \(\sigma,\widetilde\sigma\) be predictable \(\mathbb R^{n\times d}\)-valued processes, and let \(\gamma,\widetilde\gamma\) be jointly predictable maps from \([0,T]\times E\times\Omega\) to \(\mathbb R^n\). Assume that, almost surely,
		\begin{equation} \label{eq:integration by parts integrability}
			\begin{aligned}
				\int_0^T |b_t| dt + \int_0^T |\sigma_t|^2 dt + \int_0^T \int_{E} |\gamma_t(z) |^2 \lambda(t) \nu(dz) dt &< \infty, \\
				\int_0^T |\tilde{b}_t | dt + \int_0^T |\tilde{\sigma}_t |^2 dt + \int_0^T \int_{E} |\tilde{\gamma}_t(z)|^2 \lambda(t) \nu(dz) dt &< \infty.
			\end{aligned}
		\end{equation}
		
		Then for any $y_0, \tilde{y}_0 \in \R^n$, the processes
		\begin{equation*}
			\begin{aligned}
				Y_t &= y_0 + \int_0^t b_s ds + \int_0^t \sigma_s dW_s  + \int_0^t \int_{E} \gamma_s(z) \widetilde{N}(ds, dz) \\
				\tilde{Y}_t &= \tilde{y}_0 + \int_0^t \tilde{b}_s ds + \int_0^t \tilde{\sigma}_s dW_s  + \int_0^t \int_{E} \tilde{\gamma}_s(z) \widetilde{N}(ds, dz)
			\end{aligned}
		\end{equation*}
		satisfies
		\begin{equation} \label{eq:integration by parts}
			\begin{aligned}
				Y_T^\top \tilde{Y}_T = y_0^\top \tilde{y}_0 
				&+ \int_0^T Y_{t-}^\top  d \tilde{Y}_t 
				+ \int_0^T \tilde{Y}_{t-}^\top d Y_t 
				+ \int_0^T \operatorname{Tr} \big[ \sigma_t^\top \tilde{\sigma}_t \big] dt +\int_0^T \int_E \gamma_t^\top(z)\tilde{\gamma}_t(z) N(dt,dz) \, .
			\end{aligned}
		\end{equation}
	\end{Lemma}
	\begin{proof}
		Applying Itô's formula in Section~A.3 of \citet{bouchard2007introduction} with the choice
		\[
		X = \begin{bmatrix}
			Y \\
			\tilde{Y}
		\end{bmatrix}
		\quad \text{and} \quad
		f: [0, T] \times \R^{2n} \to \R;\; \left( t, \begin{bmatrix}
			y \\
			\tilde{y}
		\end{bmatrix}
		\right) \mapsto y^\top \tilde{y},
		\]
		we obtain
		\begin{align*}
			Y_T^\top \tilde{Y}_T &= y_0^\top \tilde{y}_0 
			+ \int_0^T \big(Y_t^\top \tilde{b}_t + \tilde{Y}_t^\top b_t \big) dt 
			+ \int_0^T \operatorname{Tr} \big[ \sigma_t^\top \tilde{\sigma}_t \big] dt
			+ \int_0^T \big(Y_{t-}^\top \tilde{\sigma}_t + \tilde{Y}_{t-}^\top \sigma_t \big) dW_t
			\\
			& \phantom{{}= y_0^\top \tilde{y}_0}
			+ \int_0^T \int_{E} \Big( \big( Y_{t-} + \gamma_t(z) \big)^\top \big( \tilde{Y}_{t-} + \tilde{\gamma}_t(z) \big) - Y_{t-}^\top \tilde{Y}^{\vphantom{\top}}_{t-} \Big) N(dz, dt)
			\\
			& \phantom{{}= y_0^\top \tilde{y}_0}
			- \int_0^T \int_{E} \big( Y_t^\top \tilde{\gamma}_t(z) + \tilde{Y}_t^\top \gamma_t(z) \big) \lambda(t) \nu(dz) dt
			\\
			&= y_0^\top \tilde{y}_0 
			+ \int_0^T Y_t^\top \tilde{b}_t dt
			+ \int_0^T \tilde{Y}_t^\top b_t dt
			+ \int_0^T \operatorname{Tr} \big[ \sigma_t^\top \tilde{\sigma}_t \big] dt
			\\
			& \phantom{{}= y_0^\top \tilde{y}_0}
			+ \int_0^T Y_{t-}^\top \tilde{\sigma}_t dW_t
			+ \int_0^T \tilde{Y}_{t-}^\top \sigma_t dW_t
			+\int_0^T \int_E \gamma_t^\top(z)\tilde{\gamma}_t(z) N(dt,dz) 
			\\
			& \phantom{{}= y_0^\top \tilde{y}_0}
			+ \int_0^T \int_{E} Y_{t-}^\top \tilde{\gamma}_t(z) \tilde{N}(dz, dt)
			+ \int_0^T \int_{E} \tilde{Y}_{t-}^\top \gamma_t(z) \tilde{N}(dz, dt)
			\\
			& = y_0^\top \tilde{y}_0 + \int_0^T Y^\top_{t-} d \tilde{Y}_t  + \int_0^T \tilde{Y}^\top_{t-} d Y_t
			+ \int_0^T \operatorname{Tr} \big[ \sigma_t^\top \tilde{\sigma}_t \big] dt +\int_0^T \int_E \gamma_t^\top(z)\tilde{\gamma}_t(z) N(dt,dz) 
			\qedhere
		\end{align*}
	\end{proof}
	\begin{proof}[Proof of Theorem~\ref{th:PMP}]
		Let $(\alpha, X)$ be an arbitrary admissible pair to the primal equation \eqref{sde} with the same initial value $x_0 \in \R^n$. Our goal is to show that
		\begin{equation} \label{eq:goal of proof}
			J(\hat{\alpha}) - J(\alpha) = \E \bigg[ \int_0^T \big( f(t, \hat{X}_t, \hat{\alpha}_t) - f(t, X_t, \alpha_t) \big) dt + F(\hat{X}_T) - F(X_T) \bigg] \geqslant 0.
		\end{equation}
		Since $\hat{X} - X$ is given by
		\begin{equation*}
			\begin{aligned}
				d(\hat{X} - X)_t &= \big( b(t, \hat{X}_t, \hat{\alpha}_t) - b(t, X_t, \alpha_t) \big) dt + \big( \sigma(t, \hat{X}_t, \hat{\alpha}_t) - \sigma(t, X_t, \alpha_t) \big) dW_t
				\\& \phantom{{}= \big( b(t, \hat{X}_t, \hat{\alpha}_t)}
				+ \int_{E} \big( \gamma(t, \hat{X}_{t-}, z,\hat{\alpha}_t) - \gamma(t, X_{t-}, z, \alpha_t) \big) \Tilde{N}(dt, dz),
			\end{aligned}
		\end{equation*}
		with $\hat{X}_0 - X_0 = 0$, and $(\widehat{p}, \hat{q}, \hat{r})$ is a solution tuple to the adjoint BSDE
		\begin{equation} \tag{\ref{adjoint}}
			\begin{cases}
				d\widehat{p}_t &= -\nabla_x H(t, \hat{X}_{t-}, \hat{\alpha}_t, \widehat{p}_t, \hat{q}_t, \hat{r}_t(\cdot))dt + \hat{q}_t dW_t + \int_{E} \hat{r}_{t}(z) \Tilde{N}(dt, dz)\\
				\widehat{p}_T &= \nabla_x F(\hat{X}_T)
			\end{cases}
		\end{equation}
		we deduce that
		\begin{equation*}
			\begin{aligned}
				\E \bigg[ F(\hat{X}_T) - F(X_T) \bigg] 
				& \geqslant \E \bigg[ (\hat{X}_T - X_T)^\top \nabla_x F(\hat{X}_T) \bigg]
				\\
				&= \E \bigg[ (\hat{X}_T - X_T)^\top \widehat{p}_T \bigg]
				\\
				&= \E \bigg[ \int_0^T (\hat{X}_{t-} - X_{t-})^\top  d\widehat{p}_t + \int_0^T \widehat{p}^\top_{t-} d(\hat{X} - X)_t 
				\\
				& \phantom{{}= \E \bigg[}
				+ \int_0^T \operatorname{Tr} \big[ \big(\sigma(t, \hat{X}_{t-}, \hat{\alpha}_t) - \sigma(t, X_{t-}, \alpha_t)\big)^\top \hat{q}_t \big] dt
				\\
				& \phantom{{}= \E \bigg[}
				+ \int_0^T \int_{E} \big( \gamma(t, \hat{X}_{t-}, z,\hat{\alpha}_t) - \gamma(t, X_{t-}, z, \alpha_t) \big)^\top \hat{r}_{t}(z) \lambda(t) \nu(dz) dt \bigg]
			\end{aligned}
		\end{equation*}
		where we have used the concavity of $F$ in the first line, terminal condition of $\widehat{p}$ in the second line and  Lemma~\ref{lem:integration by parts} in the third line with absolute integrability of the jump term from Cauchy-Schwarz.
		Next, by the integrability condition \ref{prop:contrac:item1}, the the following stochastic integrals are square-integrable martingales 
		\begin{equation*}
			\begin{aligned}
				&\int_0^T \big( \hat{X}_{t-} - X_{t-} \big)^\top \hat{q}_t dW_t
				\quad \text{and} \quad
				\int_0^T \int_{E} \big( \hat{X}_{t-} - X_{t-} \big)^\top \hat{r}_{t}(z) \Tilde{N}(dt, dz)
				\\
				&\int_0^T \widehat{p}_{t-}^\top \left( \sigma(t, \hat{X}_{t-}, \hat{\alpha}_t) 
				-\sigma(t, X_{t-}, \alpha_t) \right) dW_t
				\\
				&\int_0^T \int_{E} \widehat{p}_{t-}^\top  \left( \gamma(t, \hat{X}_{t-}, z,\hat{\alpha}_t) 
				- \gamma(t, X_{t-}, z, \alpha_t)\right)  \Tilde{N}(dz, dt) 
			\end{aligned}
		\end{equation*}
		and therefore have zero expectation. This implies that
		\begin{align*}
			\E \bigg[ \int_0^T \big( \hat{X}_{t-} - X_{t-} \big)^\top d\widehat{p}_t \bigg]
			&= \E \bigg[ \int_0^T \big( \hat{X}_{t-} - X_{t-} \big)^\top \bigg( \!\! - \nabla_x H(t, \hat{X}_{t-}, \hat{\alpha}_t, \widehat{p}_t, \hat{q}_t, \hat{r}_t(\cdot)) \bigg) dt \bigg]
			\\
			\E \bigg[ \int_0^T \widehat{p}^\top_{t-} d(\hat{X} - X)_t \bigg]
			&= \E \bigg[ \int_0^T \widehat{p}_t^\top \big( b(t, \hat{X}_{t-}, \hat{\alpha}_t) - b(t, X_{t-}, \alpha_t) \big) dt \bigg]
		\end{align*}
		and therefore
		\begin{equation*} 
			\begin{aligned}
				\E \bigg[ F(\hat{X}_T) - F(X_T) \bigg] 
				&\geqslant \E \bigg[ \int_0^T \big( \hat{X}_{t-} - X_{t-} \big)^\top \bigg( \!\! - \nabla_x H(t, \hat{X}_{t-}, \hat{\alpha}_t, \widehat{p}_t, \hat{q}_t, \hat{r}_t(\cdot)) \bigg) dt
				\\
				& \phantom{{}= \E \bigg[}
				+ \int_0^T \widehat{p}_t^\top \big( b(t, \hat{X}_t, \hat{\alpha}_t) - b(t, X_t, \alpha_t) \big) dt
				\\
				& \phantom{{}= \E \bigg[}
				+ \int_0^T \operatorname{Tr} \big[ \big(\sigma(t, \hat{X}_{t-}, \hat{\alpha}_t) - \sigma(t, X_{t-}, \alpha_t)\big)^\top \hat{q}_t \big] dt
				\\
				& \phantom{{}= \E \bigg[}
				+ \int_0^T \int_{E} \big( \gamma(t, \hat{X}_{t-}, z,\hat{\alpha}_t) - \gamma(t, X_{t-}, z, \alpha_t) \big)^\top \hat{r}_t(z) \lambda(t) \nu(dz) dt \bigg].
				\\
			\end{aligned}
		\end{equation*}
		On the other hand, by the definition of the Hamiltonian $H$, we have
		\begin{equation*} 
			\begin{aligned}
				\E \bigg[ \int_0^T \big( f(t, \hat{X}_t, & \hat{\alpha}_t) - f(t, X_t, \alpha_t) \big) dt \bigg]
				\\
				&= \E \bigg[ \int_0^T \big( H(t, \hat{X}_{t-}, \hat{\alpha}_t, \widehat{p}_t, \hat{q}_t, \hat{r}_t(\cdot)) - H(t, X_{t-}, \alpha_t, \widehat{p}_t, \hat{q}_t, \hat{r}_t(\cdot)) \big) dt
				\\
				& \phantom{{}= \E \bigg[}
				- \int_0^T \big( b(t, \hat{X}_t, \hat{\alpha}_t) - b(t, X_t, \alpha_t) \big)^\top \widehat{p}_t dt
				\\
				& \phantom{{}= \E \bigg[}
				- \int_0^T \Big( \operatorname{Tr} \big[ \sigma(t, \hat{X}_{t-}, \hat{\alpha}_t)^\top \hat{q}_t \big] - \operatorname{Tr} \big[ \sigma(t, X_{t-}, \alpha_t)^\top \hat{q}_t \big] \Big) dt
				\\
				& \phantom{{}= \E \bigg[}
				- \int_0^T \int_{E} \big( \gamma(t, \hat{X}_{t-}, z,\hat{\alpha}_t) - \gamma(t, X_{t-}, z, \alpha_t) \big)^\top \hat{r}_t(z) \lambda(t) \nu(dz) dt \bigg].
			\end{aligned}
		\end{equation*}
		Adding the above two (in)equalities together, we obtain
		\begin{equation*}
			\begin{aligned}
				J(\hat{\alpha}) - J(\alpha)
				&= \E \bigg[ \int_0^T \big( f(t, \hat{X}_t, \hat{\alpha}_t) - f(t, X_t, \alpha_t) \big) dt + F(\hat{X}_T) - F(X_T) \bigg]
				\\
				&\geqslant \E \Bigg[ \int_0^T \bigg( \big( \hat{X}_{t-} - X_{t-} \big)^\top \bigg( \!\! - \nabla_x H(t, \hat{X}_{t-}, \hat{\alpha}_t, \widehat{p}_t, \hat{q}_t, \hat{r}_t(\cdot)) \bigg) 
				\\
				& \phantom{{}= \E \Bigg[ \int_0^T \bigg(}
				+ H(t, \hat{X}_{t-}, \hat{\alpha}_t, \widehat{p}_t, \hat{q}_t, \hat{r}_t(\cdot)) - H(t, X_{t-}, \alpha_t, \widehat{p}_t, \hat{q}_t, \hat{r}_t(\cdot)) \bigg) dt \Bigg]
			\end{aligned}
		\end{equation*}
		The rest of the proof follows exactly the same steps as in the proof of Theorem~2.1 in \citet{framstad2004sufficient}: using the arrow condition \eqref{max2} and maximum principle \eqref{max1}, we can show that the integrand
		\[
		H(t, \hat{X}_{t-}, \hat{\alpha}_t, \widehat{p}_t, \hat{q}_t, \hat{r}_t(\cdot)) - H(t, X_{t-}, \alpha_t, \widehat{p}_t, \hat{q}_t, \hat{r}_t(\cdot)) - \big( \hat{X}_{t-} - X_{t-} \big)^\top \bigg( \nabla_x H(t, \hat{X}_{t-}, \hat{\alpha}_t, \widehat{p}_t, \hat{q}_t, \hat{r}_t(\cdot))\bigg)
		\]
		is $\mathbb{P}$-a.s.\ non-negative for all $t \in [0, T]$, and thereby deduce \eqref{eq:goal of proof}.
	\end{proof}

	\subsection{Proof of Proposition \ref{prop:noisy}} \label{proof:noisy}
	\begin{proof} 
		Since \(\widehat p_i\in L^2(\Omega,\mathcal F_T;\mathbb R^n)\), the
		conditional expectation
		\[
		p_i
		:=
		\mathbb E\!\left[
		\widehat p_i
		\,\middle|\,
		\sigma(X_i)
		\right]
		\]
		is well-defined and belongs to
		\(L^2(\Omega,\sigma(X_i);\mathbb R^n)\). By the Doob--Dynkin lemma,
		there exists a Borel function \(p^\Delta_i:\R^n\to\mathbb R^n\) such that
		\[
		p_i = p^\Delta_i(X_i)
		\qquad \mathbb{P}\text{-a.s.}
		\]
		The function $p_i^\Delta$ is unique \(\mathbb P_{X_i}\)-almost
		everywhere. Since \(X_i\) is \(\mathcal F_{t_i}\)-measurable, $\sigma(X_i) \subseteq \mathcal{F}_{t_i}$ and therefore  $p_i$ is  \(\mathcal F_{t_i}\)-measurable.
		\\[2mm]
		Moreover, since \(\mathbb R^n\) is a standard Borel space, a regular conditional
		distribution given \(X_i\) exists. Hence, \(p_i^\Delta\) can be chosen
		such that
		\[
		p_i^\Delta(x)
		=
		\mathbb E\left[
		\widehat p_i
		\,\middle|\,
		X_i=x
		\right]
		\]
		for \(\mathbb P_{X_i}\)-almost every \(x\in\mathbb R^n\).
		Evaluating this identity at \(X_i\) yields directly
		\[
		p_i
		=
		\mathbb E\!\left[
		\widehat p_i
		\,\middle|\,
		X_i
		\right].
		\]
		
		Finally, the last statement follows from the last identity and the standard \(L^2\)-projection property of conditional expectation,  see e.g.\ Theorem 4.1.15 of \cite{durrett}.
	\end{proof}
	\subsection{Proof of Proposition \ref{prop:step-averaged-martingale-coefficients}}
	\label{proof:coeff}
	\begin{proof} Let us first show (i).
		Set
		\[
		\xi_{i+1}:=p_{i+1}-\mathbb E_i[p_{i+1}].
		\]
		Then $\xi_{i+1}$ is square-integrable and satisfies
		$\mathbb E_i[\xi_{i+1}]=0$. By the predictable representation property on
		$(t_i,t_{i+1}]$, there exist predictable processes
		$q^{i+1}$ and $r^{i+1}$ such that
		\[
		\xi_{i+1}
		=
		\int_{t_i}^{t_{i+1}} q_s^{i+1}\,dW_s
		+
		\int_{t_i}^{t_{i+1}}\int_E
		r_s^{i+1}(z)\,\widetilde N(ds,dz).
		\]
		This gives \eqref{eq:pred_repr_markov}.
		\\[2mm]
		We then show (ii) and identify $\bar q_{i+1}$. Since
		$\mathbb E_i[\Delta W_i]=0$, we have
		\[
		\mathbb E_i[p_{i+1}\Delta W_i^\top]
		=
		\mathbb E_i[\xi_{i+1}\Delta W_i^\top].
		\]
		Using the representation of $\xi_{i+1}$, the orthogonality between the Brownian
		and jump martingales, and Itô's isometry,
		\[
		\begin{aligned}
			\mathbb E_i[p_{i+1}\Delta W_i^\top]
			&=
			\mathbb E_i\!\left[
			\left(\int_{t_i}^{t_{i+1}} q_s^{i+1}\,dW_s\right)
			\Delta W_i^\top
			\right]  =
			\mathbb E_i\!\left[
			\int_{t_i}^{t_{i+1}} q_s^{i+1}\,ds
			\right].
		\end{aligned}
		\]
		Dividing by $\Delta t$ gives \eqref{eq:qbar-time-average}. Moreover, by conditional Jensen's inequality and the
		Cauchy--Schwarz inequality,
		\[
		\mathbb E\left[
		|\bar q_{i+1}|^2
		\right]
		\le
		\frac{1}{\Delta t}
		\mathbb E\left[
		\int_{t_i}^{t_{i+1}}
		|q_s^{i+1}|^2\,ds
		\right],
		\]
		so \(\bar q_{i+1}\) is square-integrable.
		\\[2mm]
		We now demonstrate (iii) and  identify $\bar r_{i+1}$. Let
		\(
		h
		\in
		L^2\left(
		\Omega,\mathcal F_{t_i},\mathbb P;
		L^2(E,\nu;\mathbb R^n)
		\right).
		\)
		Since \(h\) is \(\mathcal F_{t_i}\)-measurable,
		\[
		\mathbb E_i\left[
		\int_E h(z)\,\widetilde N_i(dz)
		\right]
		=0.
		\]
		It follows that
		\[
		\begin{aligned}
			\mathbb E_i\left[
			\left\langle
			p_{i+1},
			\int_E h(z)\,\widetilde N_i(dz)
			\right\rangle
			\right]
			=
			\mathbb E_i\left[
			\left\langle
			\xi_{i+1},
			\int_E h(z)\,\widetilde N_i(dz)
			\right\rangle
			\right].
		\end{aligned}
		\]
		By orthogonality between the Brownian martingale and the compensated jump martingale and by the conditional isometry for compensated Poisson
		integrals,
		\[
		\begin{aligned}
			\mathbb E_i\left[
			\left\langle
			\xi_{i+1},
			\int_E h(z)\,\widetilde N_i(dz)
			\right\rangle
			\right]
			=
			\mathbb E_i\left[
			\int_{t_i}^{t_{i+1}}\int_E
			\left\langle
			r_s^{i+1}(z),h(z)
			\right\rangle
			\lambda(s)\nu(dz)\,ds
			\right].
		\end{aligned}
		\]
		Since \(h\) is \(\mathcal F_{t_i}\)-measurable, conditional Fubini's
		theorem yields
		\[
		\begin{aligned}
			\mathbb E_i\left[
			\int_{t_i}^{t_{i+1}}\int_E
			\left\langle
			r_s^{i+1}(z),h(z)
			\right\rangle
			\lambda(s)\nu(dz)\,ds
			\right]
			=
			\int_E
			\left\langle
			\mathbb E_i\left[
			\int_{t_i}^{t_{i+1}}
			\lambda(s)r_s^{i+1}(z)\,ds
			\right],
			h(z)
			\right\rangle
			\nu(dz).
		\end{aligned}
		\]
		Therefore,
		\[
		\begin{aligned}
			\mathbb E_i\left[
			\left\langle
			p_{i+1},
			\int_E h(z)\,\widetilde N_i(dz)
			\right\rangle
			\right]
			=
			\Lambda_i
			\int_E
			\left\langle
			\frac{1}{\Lambda_i}
			\mathbb E_i\left[
			\int_{t_i}^{t_{i+1}}
			\lambda(s)r_s^{i+1}(z)\,ds
			\right],
			h(z)
			\right\rangle
			\nu(dz).
		\end{aligned}
		\]
		Comparing this identity with \eqref{eq:rbar-def-markov} and using the
		uniqueness of the representing element in
		\(L^2(E,\nu;\mathbb R^n)\) from the Riesz representation theorem, we obtain
		\[
		\bar r_{i+1}
		=
		\frac{1}{\Lambda_i}
		\mathbb E_i\left[
		\int_{t_i}^{t_{i+1}}
		\lambda(s)r_s^{i+1}(\cdot)\,ds
		\right].
		\]
		This proves \eqref{eq:rbar-time-average}.
	\end{proof}

	\subsection{Proof of Proposition \ref{prop:jump-envelope-markov}} 
	\label{proof:envelope}
	\begin{proof} For notational convenience, we set in this proof 
		\[
		f_i := f(t_i, X_i, a^{\operatorname{sg}}_i), \qquad 
		b_i:=b(t_i,X_i,a^{\operatorname{sg}}_i),\qquad
		\sigma_i:=\sigma(t_i,X_i,a^{\operatorname{sg}}_i), \qquad
		\gamma_i(z):=\gamma(t_i,X_i,z,a^{\operatorname{sg}}_i).
		\]
		By the pathwise chain rule \eqref{eq:bptt-recursion},
		\[
		\widehat p_i
		=
		\nabla_x f_i \, \Delta t
		+
		\left(\nabla_xX^{i,X_i}_{i+1}\right)^\top
		\widehat p_{i+1},
		\qquad i=N-1,\dots,0,
		\] with $X^{i,X_i}_i = X_i$.
		Since the frozen-control construction  \eqref{eq:jump-scheme} is Markovian and  future noise increments are independent of \(\mathcal F_{t_i}\), we have
		\[
		\mathbb{E}_i[\widehat{p}_i] = 
		\mathbb E[\widehat p_i\mid \sigma(X_i)] = p_i,
		\qquad
		\mathbb E_{i+1}[\widehat{p}_{i+1}] = 
		\mathbb E[\widehat p_{i+1}\mid \sigma(X_{i+1})] =  p_{i+1}
		.
		\]
		Moreover, \(\nabla_x X_{i+1}^{i,X_i}\) is \(\mathcal F_{t_{i+1}}\)-measurable. Therefore, by the tower property,
		\begin{equation}
			\label{eq:lambda-cond-markov}
			p_i
			=
			\nabla_x f_i\,\Delta t
			+
			\mathbb E_i\!\left[
			\left(\nabla_x X^{i, X_i}_{i+1}\right)^\top p_{i+1}
			\right].
		\end{equation}
		
		Differentiating the one-step scheme \eqref{eq:jump-scheme} with respect to
		$x$ and evaluating at $X_i$ yields
		\begin{equation}
			\label{eq:dX-markov}
			\nabla_x X^{i, X_i}_{i+1}
			=
			I
			+
			\nabla_x b_i\,\Delta t
			+
			\nabla_x\sigma_i\,\Delta W_i
			+
			\int_E \nabla_x\gamma_i(z)\,\widetilde N_i(dz) \, .
		\end{equation}
		Substituting \eqref{eq:dX-markov} into \eqref{eq:lambda-cond-markov}, we obtain
		\begin{align}
			p_i
			&=
			\nabla_x f_i\,\Delta t
			+
			\mathbb E_i[p_{i+1}]
			+
			\Delta t\,(\nabla_xb_i)^\top\mathbb E_i[p_{i+1}]
			\nonumber 
			+
			\mathbb E_i\!\left[
			\left\langle p_{i+1},\nabla_x\sigma_i\,\Delta W_i\right\rangle
			\right]
			\nonumber \\
			&\qquad \qquad
			+
			\mathbb E_i\!\left[
			\left\langle
			p_{i+1},
			\int_E \nabla_x\gamma_i(z)\,\widetilde N_i(dz)
			\right\rangle
			\right] .
			\label{eq:expand-markov}
		\end{align}
		Since $\nabla_x\sigma_i$ is $\mathcal F_{t_i}$-measurable and from \eqref{eq:qbar-def-markov},
		\[
		\begin{aligned}
			\mathbb E_i\!\left[
			\left\langle p_{i+1},\nabla_x\sigma_i\,\Delta W_i\right\rangle
			\right] = 
			\mathrm{Tr}\!\left[
			(\nabla_x\sigma_i)^\top
			\mathbb E_i[p_{i+1}\Delta W_i^\top]
			\right]  =
			\Delta t\,
			\mathrm{Tr}\!\left[
			(\nabla_x\sigma_i)^\top \bar q_{i+1}
			\right].
		\end{aligned} \]
		Likewise, by \eqref{eq:rbar-def-markov} with $h=\nabla_x\gamma_i$,
		\[
		\begin{aligned}
			&\mathbb E_i\!\left[
			\left\langle
			p_{i+1},
			\int_E \nabla_x\gamma_i(z)\,\widetilde N_i(dz)
			\right\rangle
			\right]  =
			\Delta t\, \frac{\Lambda_i}{\Delta t}
			\int_E
			(\nabla_x\gamma_i(z))^\top
			\bar r_{i+1}(z)\,\nu(dz).
		\end{aligned}
		\]
		Combining the preceding identities, and since
		\begin{align*}
			\nabla_x &H_i^\Delta
			\bigl(X_i,a^{\operatorname{sg}}_i,\mathbb E_i[p_{i+1}],\bar q_{i+1},\bar r_{i+1}\bigr)
			=
			\\
			&\hspace{4mm}\nabla_x f_i
			+
			(\nabla_x b_i)^\top \mathbb E_i[p_{i+1}]
			+
			\mathrm{Tr}\!\left[(\nabla_x\sigma_i)^\top \bar q_{i+1}\right]
			+
			\frac{\Lambda_i}{\Delta t}
			\int_E
			(\nabla_x\gamma_i(z))^\top \bar r_{i+1}(z)\,\nu(dz),
		\end{align*}
		we find
		\[
		p_i
		=
		\E_i[p_{i+1}]
		+
		\nabla_x H_i^\Delta
		\bigl(X_i,a^{\operatorname{sg}}_i,\mathbb E_i[p_{i+1}],\bar q_{i+1},\bar r_{i+1}\bigr) \, \Delta t \, ,
		\]
		proving \eqref{eq:lambda-fin1-markov}. Finally, we subtract $\E_i[p_{i+1}]$ from \eqref{eq:pred_repr_markov} and insert \eqref{eq:lambda-fin1-markov} to obtain \eqref{eq:lambda-fin2-markov}.
	\end{proof}
	\subsection{Proof of Proposition \ref{prop:bptt-pmp-convergence-jumps}} \label{proof: convergence}
	\begin{proof} 	The proof follows the standard stability argument of \cite{bouchard2008discrete} for Euler schemes of BSDEs with jumps, but one has to
		distinguish carefully between the martingale-representation coefficients
		$(q^{i+1},r^{i+1})$ and their step-averages
		$(\bar q_{i+1},\bar r_{i+1})$. Moreover, we set for $t\in[t_i,t_{i+1})$,
		\[
		\widetilde{p}_t^\Delta:=\mathbb E_i[p_{i+1}],
		\qquad
		\tau_\Delta(t):=t_i .
		\]Notice that \(\widetilde p_t^\Delta\) differs from the piecewise-constant interpolation \(
		\bar p_t^\Delta:=p_i, \)
		although both processes have the same weak limit as \(\Delta t \to0\), as shown in Step~4. We also recall that
		\[
		\widehat q_t^\Delta=q_t^{i+1},
		\qquad
		\widehat r_t^\Delta=r_t^{i+1},
		\]
		whereas
		\[
		\bar q_t^\Delta=\bar q_{i+1},
		\qquad
		\bar r_t^\Delta=\bar r_{i+1}.
		\]
		The former appear in the stochastic integrals \eqref{eq:marting_coeff}, while the latter appear in
		the Hamiltonian \eqref{eq:qbar-time-average}--\eqref{eq:rbar-time-average}.
		 We now split the proof into five steps. \vspace{2mm}
		
		\textbf{Step 1: Discrete adjoint BSDE representation.}
 	By Proposition~\ref{prop:jump-envelope-markov}, the discrete Markov adjoint
		satisfies
		{\small \begin{equation}
				\label{eq:one-step-euler-jump-adjoint-corrected}
				\begin{aligned}
					p_i-p_{i+1}
					&=
					\nabla_xH^\Delta_i
					\bigl(
					X_i,a^{\operatorname{sg}}_i,
					\mathbb E_i[p_{i+1}],
					\bar q_{i+1},
					\bar r_{i+1}
					\bigr)\Delta t 
					-
					\int_{t_i}^{t_{i+1}}q_s^{i+1}\,dW_s
					-
					\int_{t_i}^{t_{i+1}}\int_E
					r_s^{i+1}(z)\,\widetilde N(ds,dz)
					.
				\end{aligned}
		\end{equation}}
		Let 
		$$\lambda^\Delta(t):=\frac{\Lambda_i}{\Delta t}, 
		\qquad t\in[t_i,t_{i+1}),$$ and define the piece-wise constant interpolation of the discrete Hamiltonian $H^\Delta_i$ in \eqref{eq:lambda-fin1-markov} by
		\[
		\begin{aligned}
			H^\Delta(t,x,a,p,q,r)
			&:=
			f(t,x,a)
			+
			b(t,x,a)^\top p
			+
			\operatorname{Tr}\!\left[\sigma(t,x,a)^\top q\right]  
			+
			\lambda^\Delta(t)
			\int_E
			\gamma(t,x,z,a)^\top r(z)\,\nu(dz).
		\end{aligned}
		\]
		Then, for $t\in[t_i,t_{i+1})$,
		\[
		\nabla_xH_i^\Delta
		\bigl(
		X_i,
		a_i^{\operatorname{sg}},
		\mathbb E_i[p_{i+1}],
		\bar q_{i+1},
		\bar r_{i+1}
		\bigr)
		=
		\nabla_xH^\Delta
		\bigl(
		\tau_\Delta(t),
		\bar X_t^\Delta,
		\bar a_t^\Delta,
			\widetilde{p}_t^\Delta,
		\bar q_t^\Delta,
		\bar r_t^\Delta
		\bigr).
		\]
		Thus \eqref{eq:one-step-euler-jump-adjoint-corrected} becomes
		{\small \begin{equation}
				\label{eq:one-step-euler-jump-adjoint-interpolated}
				\begin{aligned}
					p_i-p_{i+1}
					&=
					\int_{t_i}^{t_{i+1}}
					\nabla_x H^\Delta
					\bigl(
					\tau_\Delta(s),
					\bar X_s^\Delta,
					\bar a_s^\Delta,
					\widetilde{p}_s^\Delta,
					\bar q_s^\Delta,
					\bar r_s^\Delta
					\bigr)\,ds 
					-
					\int_{t_i}^{t_{i+1}}\widehat q_s^\Delta\,dW_s
					-
					\int_{t_i}^{t_{i+1}}\int_E
					\widehat r_s^\Delta(z)\,\widetilde N(ds,dz)
					.
				\end{aligned}
		\end{equation}}
		
		Summing backward from $i$ to $N-1$ gives, at grid points,
		{\small \begin{equation}
				\label{eq:grid-summed-euler-jump}
				\begin{aligned}
					p_i
					&=
					p_N
					+
					\sum_{k=i}^{N-1}
					\int_{t_k}^{t_{k+1}}
					\nabla_x H^\Delta
					\bigl(
					\tau_\Delta(s),
					\bar X_s^\Delta,
					\bar a_s^\Delta,
					\widetilde{p}_s^\Delta,
					\bar q_s^\Delta,
					\bar r_s^\Delta
					\bigr)\,ds 
					-
					\int_{t_i}^{T}\widehat q_s^\Delta\,dW_s
					-
					\int_{t_i}^{T}\int_E
					\widehat r_s^\Delta(z)\,\widetilde N(ds,dz)
					.
				\end{aligned}
		\end{equation}}\textbf{Step 2: Uniform a priori estimates.}
We first bound the averaged coefficients. By Jensen's inequality and the tower
		property,
		\[
		\begin{aligned}
			\mathbb E\int_{t_i}^{t_{i+1}}|\bar q_{i+1}|^2\,dt
			&=
			\Delta t\,
			\mathbb E\left[
			\left|
			\frac1{\Delta t}
			\mathbb E_i
			\left[
			\int_{t_i}^{t_{i+1}}q_s^{i+1}\,ds
			\right]
			\right|^2
			\right]  \leq
			\mathbb E\int_{t_i}^{t_{i+1}}|q_s^{i+1}|^2\,ds .
		\end{aligned}
		\]
		Summing over $i$ gives
		\begin{equation} \label{eq:contract1}
			\mathbb E\int_0^T|\bar q_t^\Delta|^2\,dt
			\leq
			\mathbb E\int_0^T|\widehat q_t^\Delta|^2\,dt .
		\end{equation}
		Similarly,
		\begin{equation*} 
			\begin{aligned}
				\mathbb E\int_{t_i}^{t_{i+1}}\int_E
				|\bar r_{i+1}(z)|^2\lambda(t)\nu(dz)\,dt &=
				\Lambda_i
				\mathbb E\int_E|\bar r_{i+1}(z)|^2\nu(dz) \leq
				\mathbb E\int_{t_i}^{t_{i+1}}\int_E
				|r_s^{i+1}(z)|^2\lambda(s)\nu(dz)\,ds .
			\end{aligned}
		\end{equation*}
		Hence
		\begin{equation} \label{eq:contract2}
			\mathbb E\int_0^T\int_E
			|\bar r_t^\Delta(z)|^2\lambda(t)\nu(dz)\,dt
			\leq
			\mathbb E\int_0^T\int_E
			|\widehat r_t^\Delta(z)|^2\lambda(t)\nu(dz)\,dt .
		\end{equation} 
		From the uniform bounds \ref{Ass1}(i) on $\nabla_xb$, $\nabla_x\sigma$, and
		$\nabla_x\gamma$, the driver
		\(
		(p,q,r)\mapsto
		\nabla_xH^\Delta(t,x,a,p,q,r)
		\)
		is uniformly Lipschitz in $(p,q,r)$.
		Indeed, for any $(p,q,r)$ and $(p',q',r')$,
		\[
		\begin{aligned}
			&
			\left|
			\nabla_xH^\Delta(t,x,a,p,q,r)
			-
			\nabla_xH^\Delta(t,x,a,p',q',r')
			\right|
			\le
			L_x|p-p'|
			+
			L_x|q-q'|
			+
			\lambda^\Delta(t)L_x
			\|r-r'\|_{L^2(E,\nu)}.
		\end{aligned}
		\]
		Since
		\[
		\|r-r'\|_{L^2(E,\nu)}
		=
		\frac{1}{\sqrt{\lambda(t)}}
		\|r-r'\|_{L^2(E,\lambda(t)\nu)},
		\]
		we obtain
		\[
		\begin{aligned}
			&
			\left|
			\nabla_xH^\Delta(t,x,a,p,q,r)
			-
			\nabla_xH^\Delta(t,x,a,p',q',r')
			\right|
			\le
			C\left(
			|p-p'|
			+
			|q-q'|
			+
			\|r-r'\|_{L^2(E,\lambda(t)\nu)}
			\right),
		\end{aligned}
		\]
		where
		\(
		C
		=
		L_x (
		2+ \sup_{\Delta,t}\lambda^\Delta(t) / 
		\sqrt{\underline\lambda}
		).
		\)
		Because $\lambda$ is continuous and strictly positive on $[0,T]$,
		\[
		0<\underline\lambda
		\le\lambda(t)\le\overline\lambda<\infty,
		\qquad 
		\text{and}
		\qquad
		\sup_{\Delta,t}\lambda^\Delta(t)\le\overline\lambda \, .
		\]
		Hence the Lipschitz constant is independent of $\Delta$.
		\\[3mm]
		Applying the \(L^2\) discrete a priori estimate for Lipschitz BSDEs with jumps, see Lemma A.2(i) and Proposition A.1 of \cite{bouchard2008discrete}, to the piecewise-interpolated equation \eqref{eq:grid-summed-euler-jump}, we obtain
		\[
		\begin{aligned}
			&\mathbb E\left[
			\max_{0\leq i\leq N}|p_i|^2
			+
			\int_0^T|\widehat q_t^\Delta|^2\,dt
			+
			\int_0^T\int_E
			|\widehat r_t^\Delta(z)|^2
			\lambda(t)\nu(dz)\,dt
			\right] \\
			&\qquad\leq
			C\,
			\mathbb E\left[
			|p_N|^2
			+
			\left(
			\int_0^T
			\left|
			\nabla_xH^\Delta
			\bigl(
			\tau_\Delta(t),
			\bar X_t^\Delta,
			\bar a_t^\Delta,
			0,0,0
			\bigr)
			\right|\,dt
			\right)^2
			\right],
		\end{aligned}
		\]
		where $C$ is independent of $\Delta$.  By Assumption~\ref{Ass1}(ii),
		\[
		|\nabla_xf(t,x,a)|^2
		\le
		C\bigl(1+|x|^2+|a|^2\bigr),
		\qquad
		|\nabla_xF(x)|^2
		\le
		C\bigl(1+|x|^2\bigr).
		\]
		Since
		\(
		\bar X^\Delta\to X_-
	\) and \(
		\bar a^\Delta\to a
		\)
		in their respective $L^2$ spaces, both sequences are uniformly
		bounded in $L^2$. Therefore, using \( \nabla_xH^\Delta(t,x,a,0,0,0)=\nabla_x f(t,x,a) \), we obtain 
		\begin{align*}
			\sup_\Delta
			\mathbb E\left[
			\left(
			\int_0^T
			\left|
			\nabla_xH^\Delta
			\bigl(
			\tau_\Delta(t),
			\bar X_t^\Delta,
			\bar a_t^\Delta,
			0,0,0
			\bigr)
			\right|dt
			\right)^2
			\right]
			&=
			\sup_\Delta
			\mathbb E\left[
			\left(
			\int_0^T
			\left|
			\nabla_xf
			\bigl(
			\tau_\Delta(t),
			\bar X_t^\Delta,
			\bar a_t^\Delta
			\bigr)
			\right|dt
			\right)^2
			\right] \nonumber
			\\
			& \le
			CT\sup_\Delta
			\mathbb E\int_0^T
			\left(
			1+|\bar X_t^\Delta|^2
			+|\bar a_t^\Delta|^2
			\right)dt
			<\infty.
		\end{align*}
		Moreover, since $p_N=\nabla_xF(X_N)$ and
		$X_N\to X_T$ in $L^2$,
		\[
		\sup_\Delta\mathbb E|p_N|^2
		\le
		C\left(1+\sup_\Delta\mathbb E|X_N|^2\right)
		<\infty.
		\]
		The discrete a priori estimate therefore yields
	\begin{equation}  \label{firs_bound}
		\sup_\Delta
		\mathbb E\left[
		\max_{0\le i\le N}|p_i|^2
		+
		\int_0^T|\widehat q_t^\Delta|^2dt
		+
		\int_0^T\int_E
		|\widehat r_t^\Delta(z)|^2
		\lambda(t)\nu(dz)dt
		\right]
		<\infty.
		\end{equation}

		This proves the claimed stability estimates for both the martingale
		coefficients and their step averages.  Moreover,  again 
		by Assumption~\ref{Ass1}(ii) and the Cauchy--Schwarz inequality, we have
		\begin{equation} \label{second_bound}
		\sup_\Delta
		\mathbb E\left[
		\left(
		\int_0^T
		\left|
		\nabla_xf\bigl(
		\tau_\Delta(t),
		\bar X_t^\Delta,
		\bar a_t^\Delta
		\bigr)
		\right|dt
		\right)^2
		\right]
		\le
		T
		\sup_\Delta
		\mathbb E\int_0^T
		\left|
		\nabla_xf\bigl(
		\tau_\Delta(t),
		\bar X_t^\Delta,
		\bar a_t^\Delta
		\bigr)
		\right|^2dt
		<\infty.
		\end{equation}\vspace{2mm}
		
		\textbf{Step 3: Weak compactness.}
	By weak compactness in the corresponding Hilbert spaces, there exist a
		subsequence, not relabeled, and processes \(p,q,r\) such that
		\[
		\bar p^\Delta\rightharpoonup p
		\quad\text{weakly in }
		L^2(\Omega\times[0,T];\mathbb R^n),
		\]
		\[
		\widehat q^\Delta\rightharpoonup q
		\quad\text{weakly in }
		L^2_{\mathbb F}(0,T;\mathbb R^{n\times d}),
		\]
		\[
		\widehat r^\Delta\rightharpoonup r
		\quad\text{weakly in }
		L^2_{\nu,\mathbb F}(0,T;\mathbb R^n).
		\]
		
		We next show that the averaged coefficients have the same weak limits. Define
		the Brownian averaging operator
		\begin{equation} \label{op1}
			(A_\Delta u)_t
			:=
			\frac1{\Delta t}
			\mathbb E_i
			\left[
			\int_{t_i}^{t_{i+1}}u_s\,ds
			\right],
			\qquad
			t\in[t_i,t_{i+1}).
		\end{equation}
		Then
		\(\
		\bar q^\Delta=A_\Delta\widehat q^\Delta .
		\)
		Reasoning as above, it is clear that the operators $A_\Delta$ are contractions on $L^2_{\mathbb{F}}(0,T;\R^{n \times d})$, are self-adjoint with respect to the $L^2$ inner product, and satisfy
		\[
		A_\Delta\varphi\to\varphi
		\quad\text{strongly in }L^2_{\mathbb{F}}(0,T;\R^{n \times d}) .
		\]
		for every predictable test process $\varphi\in L^2_{\mathbb{F}}(0,T;\R^{n\times d})$. Therefore,
		\[
		\begin{aligned}
			\langle \bar q^\Delta,\varphi\rangle
			&=
			\langle A_\Delta\widehat q^\Delta,\varphi\rangle =
			\langle \widehat q^\Delta,A_\Delta\varphi\rangle
			\to
			\langle q,\varphi\rangle ,
		\end{aligned}
		\]
		and hence
		\[
		\bar q^\Delta\rightharpoonup q
		\quad\text{weakly in } L^2_{\mathbb{F}}(0,T;\R^{n\times d}).
		\]
		
		The same argument applies to the jump coefficient. Define
		\begin{equation} \label{op2}
			(B_\Delta u)_t(z)
			:=
			\frac1{\Lambda_i}
			\mathbb E_i
			\left[
			\int_{t_i}^{t_{i+1}}\lambda(s)u_s(z)\,ds
			\right],
			\qquad
			t\in[t_i,t_{i+1}).
		\end{equation}
		Since $\lambda$ is bounded away from zero, this is well-defined. Then
		\[
		\bar r^\Delta=B_\Delta\widehat r^\Delta .
		\]
		The operators $B_\Delta$ are contractions on
		$
		L^2_{\nu, \mathbb{F}}(0,T;\R^n)
		$, are self-adjoint
		and satisfy $
		B_\Delta\psi\to\psi$ strongly for every predictable test function $\psi$. Consequently,
		\[
		\bar r^\Delta\rightharpoonup r
		\quad\text{weakly in }
		L^2_{\nu, \mathbb{F}}(0,T;\R^n).
		\]
		
		\textbf{Step 4: Identification of the limit.}
		We now identify the weak limit. Since $X_N\to X_T$ in
		$L^2(\Omega,\mathcal F_T;\mathbb R^n)$, we have
		$X_N\to X_T$ in probability. By continuity of $\nabla_xF$,
		\[
		\nabla_xF(X_N)\to\nabla_xF(X_T)
		\quad\text{in probability}.
		\]
		Moreover, the linear-growth bound from Assumption \ref{Ass1} implies 
		\[
		|\nabla_xF(X_N)-\nabla_xF(X_T)|^2
		\le
		C\bigl(1+|X_N|^2+|X_T|^2\bigr).
		\] 
		The $L^2$-convergence of $X_N$ implies that
		$\{|X_N|^2\}_\Delta$ is uniformly integrable. Hence 
		the family
		\(
		\bigl\{|\nabla_xF(X_N) - \nabla_x F(X_T)|^2\bigr\}_{\Delta}
		\)
		is uniformly integrable. Hence, by Vitali's convergence theorem,
		\[
		p_N=\nabla_xF(X_N)
		\longrightarrow
		\nabla_xF(X_T)
		\quad\text{in }
		L^2(\Omega,\mathcal F_T;\mathbb R^n).
		\]
		Moreover, from \eqref{eq:lambda-fin1-markov}, we find for $t\in[t_i,t_{i+1})$,
		\[
		\bar p_t^\Delta-	\widetilde{p}_t^\Delta
		=
		\nabla_xH^\Delta
		\bigl(
		\tau_\Delta(t),
		\bar X_t^\Delta,
		\bar a_t^\Delta,
			\widetilde{p}_t^\Delta,
		\bar q_t^\Delta,
		\bar r_t^\Delta
		\bigr)\Delta t
		.
		\]
		
		Hence, we obtain
		\[
		\begin{aligned}
			\mathbb E\int_0^T|\bar p_t^\Delta-	\widetilde{p}_t^\Delta|^2\,dt
			&=
			(\Delta t)^2
			\mathbb E\int_0^T
			\left|
			\nabla_xH^\Delta
			\bigl(
			\tau_\Delta(t),
			\bar X_t^\Delta,
			\bar a_t^\Delta,
				\widetilde{p}_t^\Delta,
			\bar q_t^\Delta,
			\bar r_t^\Delta
			\bigr)
			\right|^2\,dt  \,. 
		\end{aligned}
		\]
		Moreover, since the driver $\nabla_x H^\Delta$ is uniformly Lipschitz in \((p,q,r)\), we have
		\[
		\begin{aligned}
			&
			\left|
			\nabla_xH^\Delta
			\bigl(
			\tau_\Delta(t),
			\bar X_t^\Delta,
			\bar a_t^\Delta,
				\widetilde{p}_t^\Delta,
			\bar q_t^\Delta,
			\bar r_t^\Delta
			\bigr)
			\right|^2 
			\\
			&\hspace{5mm}
			\le
			C\left(
			\left|
			\nabla_xH^\Delta
			\bigl(
			\tau_\Delta(t),
			\bar X_t^\Delta,
			\bar a_t^\Delta,
			0,0,0
			\bigr)
			\right|^2
			+
			|\widetilde{p}_t^\Delta|^2
			+
			|\bar q_t^\Delta|^2
			+
			\int_E |\bar r_t^\Delta(z)|^2 \lambda(t)\nu(dz) 
			\right).
		\end{aligned}
		\]
		The first term on the right-hand side is uniformly bounded, see \eqref{second_bound}. The terms involving \(\bar q^\Delta\) and \(\bar r^\Delta\) are uniformly bounded by the estimates in \eqref{firs_bound}. Finally, \(	\widetilde{p}^\Delta\) is uniformly bounded in \(L^2(\Omega\times[0,T];\R^n)\), since 
		\(
			\widetilde{p}_t^\Delta=\mathbb E_i[p_{i+1}]\) for \(t\in[t_i,t_{i+1})\),
		and therefore, by Jensen's inequality,
	\[
		\begin{aligned}
			\mathbb E\int_0^T |	\widetilde{p}_t^\Delta|^2 dt
			=
			\sum_{i=0}^{N-1}
			\mathbb E\int_{t_i}^{t_{i+1}}
			\left|\mathbb E_i[p_{i+1}]\right|^2 dt 
			\le
			\sum_{i=0}^{N-1}
			\Delta t \,
			\mathbb E\left[|p_{i+1}|^2\right] 
			\le
			T \, \mathbb E\left[
			\max_{0\le i\le N}|p_i|^2
			\right],
		\end{aligned}
		\]
		which is uniformly bounded by the a priori estimate \eqref{firs_bound}. Hence
		\[
		\sup_\Delta
		\mathbb E\int_0^T
		\left|
		\nabla_xH^\Delta
		\bigl(
		\tau_\Delta(t),
		\bar X_t^\Delta,
		\bar a_t^\Delta,
		\widetilde{p}_t^\Delta,
		\bar q_t^\Delta,
		\bar r_t^\Delta
		\bigr)
		\right|^2dt
		<\infty.
		\]
		Consequently,
		\[
		\begin{aligned}
			\mathbb E\int_0^T|\bar p_t^\Delta-	\widetilde{p}_t^\Delta|^2 \, dt
			=
			(\Delta t)^2
			\mathbb E\int_0^T
			\left|
			\nabla_xH^\Delta
			\bigl(
			\tau_\Delta(t),
			\bar X_t^\Delta,
			\bar a_t^\Delta,
				\widetilde{p}_t^\Delta,
			\bar q_t^\Delta,
			\bar r_t^\Delta
			\bigr)
			\right|^2dt \le C(\Delta t)^2
			\to0.
		\end{aligned}
		\]
		Therefore,
		\(
		\bar p^\Delta-	\widetilde{p}^\Delta
		\to 0 \) in $L^2(\Omega\times[0,T];\R^n)$, and hence, $	\widetilde{p}^\Delta$ and $\bar p^\Delta$ have the same weak limit which we denote $p$. \\[3.5mm]
		Next, $\lambda^\Delta\to\lambda$ uniformly on $[0,T]$, because $\lambda$ is
		continuous on $[0,T]$. Also, by Assumption \ref{itemize:2},
		\[
		\bar X^\Delta\to X_-,
		\qquad
		\bar a^\Delta\to a
		\quad\text{in }L^2(\Omega\times[0,T];\R^n).
		\]
		Together with the continuity and growth assumptions in  \ref{itemize:1} on the spatial
		derivatives of \(b,\sigma,\gamma,f\), this implies that the coefficients in
		the driver converge strongly in the corresponding \(L^2\)-spaces. Lemma \ref{lem:weak-driver-convergence} then shows that the strong convergence of these coefficients together with the weak convergences 
		\(\widetilde{p}^\Delta\rightharpoonup p\), \(\bar q^\Delta\rightharpoonup q\), and
		\(\bar r^\Delta\rightharpoonup r\), implies weak convergence of the full driver
		\[
		\begin{aligned}
			&\nabla_xH^\Delta
			\bigl(
			\tau_\Delta(t),
			\bar X_t^\Delta,
			\bar a_t^\Delta,
			\widetilde{p}_t^\Delta,
			\bar q_t^\Delta,
			\bar r_t^\Delta
			\bigr)  \rightharpoonup
			\nabla_xH(t,X_{t-},a_t,p_t,q_t,r_t)
			\quad\text{weakly in }L^2(\Omega\times[0,T]; \R^n).
		\end{aligned}
		\]
		
		Passing to the limit in the grid identity
		\eqref{eq:grid-summed-euler-jump}, in the standard weak formulation of
		BSDE Euler schemes, see \cite{bouchard2008discrete, alasseur2024deep}, gives
		\[
		p_t
		=
		\nabla_xF(X_T)
		+
		\int_t^T
		\nabla_xH(s,X_{s-},a_s,p_s,q_s,r_s)\,ds
		-
		\int_t^T q_s\,dW_s
		-
		\int_t^T\int_E r_s(z)\,\widetilde N(ds,dz),
		\]
		for the càdlàg adapted modification of $p$. Equivalently,
		\[
		dp_t
		=
		-\nabla_xH(t,X_{t-},a_t,p_t,q_t,r_t)\,dt
		+
		q_t\,dW_t
		+
		\int_E r_t(z)\,\widetilde N(dt,dz),
		\qquad
		p_T=\nabla_xF(X_T).
		\]
		By the well-posedness result established after
		\eqref{adjoint}, the limiting Lipschitz BSDE admits a unique
		square-integrable adapted solution. Therefore, every weakly
		convergent subsequence has the same limit, and the entire family
		converges weakly to the PMP adjoint triplet. \\[3mm]
		\textbf{Step 5: Identification of the limiting Markov form.}
		It remains to identify the Markov form of the limiting martingale
		coefficients. Suppose that the limiting adjoint admits a smooth deterministic
		decoupling field
		\[
		p\in C^{1,2}([0,T]\times \R^n),
		\qquad
		p_t=p(t,X_t).
		\]
		Applying It\^o--L\'evy's formula to $p(t,X_t)$ gives
		\[
		\begin{aligned}
			dp(t,X_t)
			&=
			\text{(finite variation term)}\,dt
			+
			\nabla_xp(t,X_{t^-})
			\sigma(t,X_{t^-},a_t)\,dW_t \\
			&\quad
			+
			\int_E
			\left[
			p\!\left(
			t,
			X_{t-}+\gamma(t,X_{t-},z,a_t)
			\right)
			-
			p(t,X_{t-})
			\right]\widetilde N(dt,dz).
		\end{aligned}
		\]
		Comparing this semimartingale decomposition with
		\[
		dp_t
		=
		-\nabla_xH(t,X_{t-},a_t,p_t,q_t,r_t)\,dt
		+
		q_t\,dW_t
		+
		\int_E r_t(z)\,\widetilde N(dt,dz),
		\]
		and using uniqueness of the Brownian and compensated-jump martingale
		representation, we obtain
		\[
		q_t
		=
		\nabla_xp(t,X_{t^-})
		\sigma(t,X_{t^-},a_t),
		\]
		for $dt\otimes d\mathbb P$-a.e.\ $(t,\omega)$,  and
		\[
		r_t(z)
		=
		p\!\left(
		t,
		X_{t-}+\gamma(t,X_{t-},z,a_t)
		\right)
		-
		p(t,X_{t-}) ,
		\]
		for
		$dt\otimes d\mathbb P\otimes\nu(dz)$-a.e.\
		$(t,\omega,z)$.
	\end{proof}
\begin{Lemma}[Weak convergence of the adjoint drivers]
	\label{lem:weak-driver-convergence}
	Under the assumptions of
	Proposition~\ref{prop:bptt-pmp-convergence-jumps}, let
	$\Delta\to0$ along a subsequence such that
	\[
	\widetilde p^\Delta\rightharpoonup p
	\quad\text{weakly in }
	L^2(\Omega\times[0,T];\mathbb R^n),
	\]
	\[
	\bar q^\Delta\rightharpoonup q
	\quad\text{weakly in }
	L^2_{\mathbb F}(0,T;\mathbb R^{n\times d}),
	\]
	\[
	\bar r^\Delta\rightharpoonup r
	\quad\text{weakly in }
	L^2_{\nu,\mathbb F}(0,T;\mathbb R^n).
	\]
	Then
	\[
	\begin{aligned}
		&\nabla_xH^\Delta
		\bigl(
		\tau_\Delta(\cdot),
		\bar X^\Delta,
		\bar a^\Delta,
		\widetilde p^\Delta,
		\bar q^\Delta,
		\bar r^\Delta
		\bigr)
	\rightharpoonup
		\nabla_xH
		\bigl(
		\cdot,X_-,a,p,q,r
		\bigr) 	\text{ \  weakly in  \ }
		L^2(\Omega\times[0,T];\mathbb R^n).
	\end{aligned}
	\]
\end{Lemma}

\begin{proof}
	Set
	\[
	\begin{aligned}
	\ell_t^\Delta
		&:=
		\nabla_xf\bigl(
		\tau_\Delta(t),
		\bar X_t^\Delta,
		\bar a_t^\Delta
		\bigr),\\
		\beta_t^\Delta
		&:=
		\nabla_xb\bigl(
		\tau_\Delta(t),
		\bar X_t^\Delta,
		\bar a_t^\Delta
		\bigr),\\
		\Sigma^\Delta_{t}
		&:=
		\nabla_x\sigma\bigl(
		\tau_\Delta(t),
		\bar X_t^\Delta,
		\bar a_t^\Delta
		\bigr),\\
		\Gamma_{t}^\Delta(z)
		&:=
		\nabla_x\gamma\bigl(
		\tau_\Delta(t),
		\bar X_t^\Delta,
		z,
		\bar a_t^\Delta
		\bigr),
	\end{aligned}
	\]
	and define $\ell_{t}$, $\beta_{t}$, $\Sigma_{t}$, and $\Gamma_{t}(z)$ analogously by replacing
	\(
	\bigl(
	\tau_\Delta(t),\bar X_t^\Delta,\bar a_t^\Delta
	\bigr)\) with \(
	(t,X_{t-},a_t).
	\)
	
	By the convergence assumptions of
	Proposition~\ref{prop:bptt-pmp-convergence-jumps} and the continuity
	of the derivatives, the composed coefficients converge in
	probability. The uniform boundedness of
	$\beta^\Delta$, $\Sigma^\Delta$, and $\Gamma^\Delta$ implies
	\[
	\beta^\Delta\to \beta,
	\qquad
	\Sigma^\Delta\to \Sigma
	\]
	strongly in the corresponding $L^2$ spaces, and
	\[
	\Gamma^\Delta\to\Gamma
	\quad\text{strongly in }
	L^2\bigl(
	\Omega\times[0,T]\times E,
	\mathbb P\otimes dt\otimes\nu
	\bigr).
	\]
	Moreover, the linear-growth bound on $\nabla_xf$ and the
	$L^2$-convergence of $(\bar X^\Delta,\bar a^\Delta)$ yield, by
	Vitali's theorem,
	\[
	\ell^\Delta\to \ell
	\quad\text{strongly in }
	L^2(\Omega\times[0,T];\mathbb R^n).
	\]
	
	Let
	\(
	\varphi\in
	L^2_\mathbb{F}(0,T;\mathbb R^n)
	\)
	be an arbitrary test process. For the drift term,
	\[
	\mathbb E\int_0^T
	\left\langle
	(\beta_t^\Delta)^\top\widetilde p_t^\Delta,\varphi_t
	\right\rangle dt
	=
	\mathbb E\int_0^T
	\left\langle
	\widetilde p_t^\Delta,\beta_t^\Delta\varphi_t
	\right\rangle dt.
	\]
	Since $\beta^\Delta$ is uniformly bounded and converges in probability
	to $\beta$,
	\[
	\beta^\Delta\varphi\to \beta\varphi
	\quad\text{strongly in }
	L^2(\Omega\times[0,T]).
	\]
	The weak convergence of $\widetilde p^\Delta$ therefore gives
	\[
	(\beta^\Delta)^\top\widetilde p^\Delta
	\rightharpoonup
	\beta^\top p
	\quad\text{weakly in }
	L^2(\Omega\times[0,T];\mathbb R^n).
	\]
	Similarly, define the linear map
	\[
	\mathcal S_t^\Delta:
	\mathbb R^{n\times d}\to\mathbb R^n,
	\qquad
	\mathcal S_t^\Delta(Q)
	:=
	\nabla_x\operatorname{Tr}\!\left[
	\sigma\bigl(
	\tau_\Delta(t),
	\bar X_t^\Delta,
	\bar a_t^\Delta
	\bigr)^\top Q
	\right] ,
	\]
	and define $\mathcal S_t$ analogously. Then
	\[
	\begin{aligned}
		\mathbb E\int_0^T
		\left\langle
		\mathcal S_t^\Delta(\bar q_t^\Delta),
		\varphi_t
		\right\rangle dt
		&=
		\mathbb E\int_0^T
		\left\langle
		\bar q_t^\Delta,
		(\mathcal S_t^\Delta)^*\varphi_t
		\right\rangle dt.
	\end{aligned}
	\]
	Since
	\[
	(\mathcal S^\Delta)^*\varphi
	\to
	\mathcal S^*\varphi
	\quad\text{strongly in }
	L^2_{\mathbb F}(0,T;\mathbb R^{n\times d}),
	\]
	the weak convergence of $\bar q^\Delta$ yields
	\[
	\mathcal S^\Delta(\bar q^\Delta)
	\rightharpoonup
	\mathcal S(q)
	\quad\text{weakly in }
	L^2(\Omega\times[0,T];\mathbb R^n).
	\]
	For the jump term,
	\[
	\begin{aligned}
		&\mathbb E\int_0^T
		\left\langle
		\lambda^\Delta(t)
		\int_E
		\Gamma_t^\Delta(z)^\top\bar r_t^\Delta(z)\,\nu(dz),
		\varphi_t
		\right\rangle dt
		\\
		&\qquad=
		\mathbb E\int_0^T\int_E
		\left\langle
		\bar r_t^\Delta(z),
		\frac{\lambda^\Delta(t)}{\lambda(t)}
		\Gamma_t^\Delta(z)\varphi_t
		\right\rangle
		\lambda(t)\nu(dz)\,dt.
	\end{aligned}
	\]
	The uniform convergence $\lambda^\Delta\to\lambda$, the lower bound
	on $\lambda$, and the convergence of $\Gamma^\Delta$ imply
	\[
	\frac{\lambda^\Delta(t)}{\lambda(t)}
	\Gamma_t^\Delta(\cdot)\varphi_t
	\longrightarrow
	\Gamma_t(\cdot)\varphi_t
	\]
	strongly in
	$L^2_{\nu,\mathbb F}(0,T;\mathbb R^n)$.
	The weak convergence of $\bar r^\Delta$ therefore yields
	\[
	\lambda^\Delta(t)
	\int_E
	\Gamma_t^\Delta(z)^\top\bar r_t^\Delta(z)\,\nu(dz)
	\rightharpoonup
	\lambda(t)
	\int_E
	\Gamma_t(z)^\top r_t(z)\,\nu(dz)
	\]
	weakly in
	$L^2(\Omega\times[0,T];\mathbb R^n)$.
	
	Finally,
	\[
	\begin{aligned}
		&\nabla_xH^\Delta
		\bigl(
		\tau_\Delta(t),
		\bar X_t^\Delta,
		\bar a_t^\Delta,
		\widetilde p_t^\Delta,
		\bar q_t^\Delta,
		\bar r_t^\Delta
		\bigr)
		=
		\ell_t^\Delta
		+
		(\beta_t^\Delta)^\top\widetilde p_t^\Delta
		+
		\mathcal S_t^\Delta(\bar q_t^\Delta)
		+
		\lambda^\Delta(t)
		\int_E
		\Gamma_t^\Delta(z)^\top
		\bar r_t^\Delta(z)\,\nu(dz).
	\end{aligned}
	\]
	Combining the preceding convergences proves the result.
\end{proof}
	{\footnotesize
		\bibliographystyle{apa}  
		\bibliography{bibPMP}   
		
	}

\end{document}